\documentclass[conference,twocolumn]{style/IEEEtran}
\IEEEoverridecommandlockouts
\usepackage[utf8]{inputenc}
\usepackage[T1]{fontenc}
\usepackage[english]{babel}
\usepackage{amsmath,amssymb,amsthm,bbm}
\usepackage{graphicx}
\usepackage{float}
\usepackage{cuted}
\usepackage{placeins}
\usepackage{afterpage}
\usepackage{xfrac}
\usepackage{tikz}
\usepackage[dvipsnames]{xcolor}
\usepackage[sort]{natbib}
\ifdefined\TexpadTeXMode\else
  \usepackage{microtype}
\fi
\usepackage{dblfloatfix}
\usepackage{balance}
\usepackage{caption}
\definecolor{mydarkblue}{rgb}{0,0.08,0.65}
\usepackage[colorlinks=true,linkcolor=mydarkblue,citecolor=mydarkblue,filecolor=mydarkblue,urlcolor=mydarkblue]{hyperref}
\usepackage[nameinlink,noabbrev]{cleveref}
\usepackage{style/math}
\usepackage{titletoc}
\ifdefined\TexpadTeXMode
  \makeatletter
  \def\ttl@gobblecontents#1#2#3#4{\ignorespaces}
  \makeatother
\fi

\setcitestyle{numbers,square,comma}
\makeatletter
\def\@IEEEsectpunct{\enspace}
\renewcommand{\paragraph}{\@startsection{paragraph}{4}{\z@}%
  {1.5ex plus 0.5ex minus 0.2ex}{-1em}{\normalfont\normalsize\itshape}}
\makeatother

\newcommand{\zyphraaffiliation}{\textsuperscript{1}\,Zyphra Research}
\newcommand{\equalcontribution}{{\mdseries\footnotesize *\,Equal contribution}}

\newsavebox{\mainpointbox}

\crefname{figure}{Figure}{Figures}
\crefname{table}{Table}{Tables}
\crefname{section}{Section}{Sections}

\crefname{appendix}{Appendix}{Appendices}
\Crefname{appendix}{Appendix}{Appendices}
\makeatletter
\newcommand{\legacylabelalias}[2]{%
  \@ifundefined{r@#2}{}{%
    \expandafter\let\expandafter\legacylabeltarget\csname r@#2\endcsname
    \expandafter\let\csname r@#1\endcsname\legacylabeltarget
  }%
  \@ifundefined{r@#2@cref}{}{%
    \expandafter\let\expandafter\legacylabeltarget\csname r@#2@cref\endcsname
    \expandafter\let\csname r@#1@cref\endcsname\legacylabeltarget
  }%
}
\AtBeginDocument{%
  \legacylabelalias{app:dense-top-p-heads}{app:top-p-dense-heads}%
  \legacylabelalias{app:eval-protocols}{app:evaluation-protocols}%
  \legacylabelalias{app:eval-thresholds}{app:training-evaluation}%
  \legacylabelalias{app:hatt-sparsity-scaling}{app:sparsity-scaling}%
  \legacylabelalias{app:nemotron-evals}{app:evaluation-nemotron}%
  \legacylabelalias{app:rollout-linear-preference}{app:top-p-rollout-fits}%
  \legacylabelalias{app:ruler-64k-128k}{app:evaluation-ruler-contexts}%
  \legacylabelalias{app:ruler-geometry-tasks}{app:geometry}%
  \legacylabelalias{app:training-setup}{app:training-compute}%
  \legacylabelalias{eq:dense-split}{eq:roofline-dense-split-latency}%
  \legacylabelalias{eq:dense-unsplit}{eq:roofline-dense-unsplit-latency}%
  \legacylabelalias{eq:hatt-exponent-compounding}{eq:sparsity-exponent-compounding}%
  \legacylabelalias{eq:hatt-pruning-correction}{eq:sparsity-pruning-correction}%
  \legacylabelalias{eq:hatt-selection-power-law}{eq:sparsity-selection-power-law}%
  \legacylabelalias{eq:hatt-top-exponent-coefficients}{eq:sparsity-top-exponent-coefficients}%
  \legacylabelalias{eq:hatt-top-exponent-depth}{eq:sparsity-top-exponent-depth}%
  \legacylabelalias{eq:hatt-top-exponent-per-threshold}{eq:sparsity-top-exponent-thresholds}%
  \legacylabelalias{eq:hatt_roofline}{eq:roofline-zha-latency}%
  \legacylabelalias{eq:ruler-geometry-model}{eq:geometry-accuracy-model}%
  \legacylabelalias{eq:topp_denoiser}{eq:motivation-top-p-decomposition}%
  \legacylabelalias{fig:app-nemotron-dense-top-p}{fig:dense-top-p-nemotron-heads}%
  \legacylabelalias{fig:arch-schema}{fig:architecture-overview}%
  \legacylabelalias{fig:dense-bf16-relative}{fig:roofline-dense-bf16-ratio}%
  \legacylabelalias{fig:dense-bf16-times}{fig:roofline-dense-bf16-latency}%
  \legacylabelalias{fig:dense-fp16-relative}{fig:roofline-dense-fp16-ratio}%
  \legacylabelalias{fig:dense-fp16-times}{fig:roofline-dense-fp16-latency}%
  \legacylabelalias{fig:geometry-schema}{fig:architecture-geometry}%
  \legacylabelalias{fig:hatt-depth-latency}{fig:roofline-depth-latency}%
  \legacylabelalias{fig:hatt-latency}{fig:speedup-latency}%
  \legacylabelalias{fig:hatt-latency-thresholds}{fig:speedup-latency-thresholds}%
  \legacylabelalias{fig:hatt-roofline-tree}{fig:roofline-traversal}%
  \legacylabelalias{fig:hatt-sparsity-scaling}{fig:sparsity-scaling}%
  \legacylabelalias{fig:hatt-sparsity-scaling-thresholds}{fig:sparsity-thresholds}%
  \legacylabelalias{fig:hatt-top-exponent-depth}{fig:sparsity-top-exponent-depth}%
  \legacylabelalias{fig:latency-elasticity}{fig:roofline-latency-elasticity}%
  \legacylabelalias{fig:nemotron-lightning-evals}{fig:evaluation-nemotron}%
  \legacylabelalias{fig:paper-dense-vs-hatt-allp}{fig:evaluation-qwen-thresholds}%
  \legacylabelalias{fig:paper-dense-vs-hatt-nemotron-allp}{fig:evaluation-nemotron-thresholds}%
  \legacylabelalias{fig:paper-dense-vs-hatt-r1c8}{fig:evaluation-qwen-small-router}%
  \legacylabelalias{fig:paper-dense-vs-hatt-wide-fp4}{fig:evaluation-qwen-fp4}%
  \legacylabelalias{fig:paper-keys-msa-hatt-allp}{fig:evaluation-qwen-key-share}%
  \legacylabelalias{fig:qwen-evals}{fig:evaluation-qwen}%
  \legacylabelalias{fig:ruler-256k-pareto}{fig:geometry-ruler-pareto}%
  \legacylabelalias{fig:ruler-geometry-256k}{fig:geometry-ruler-model}%
  \legacylabelalias{fig:ruler-geometry-coverage}{fig:geometry-coverage}%
  \legacylabelalias{fig:ruler-geometry-tasks-256k}{fig:geometry-ruler-tasks}%
  \legacylabelalias{sec:arch}{sec:architecture}%
  \legacylabelalias{sec:evals}{sec:evaluation}%
  \legacylabelalias{sec:evals:geom}{sec:evaluation-geometry}%
  \legacylabelalias{sec:intro}{sec:introduction}%
  \legacylabelalias{sec:motives}{sec:motivation}%
  \legacylabelalias{sec:msa-roofline}{app:roofline-msa}%
  \legacylabelalias{subsec:natural_sparsity}{sec:motivation-natural-sparsity}%
  \legacylabelalias{tab:dense_variables}{tab:roofline-dense-variables}%
  \legacylabelalias{tab:hatt_tree_variables}{tab:roofline-tree-variables}%
  \legacylabelalias{tab:hatt_variables}{tab:roofline-zha-variables}%
  \legacylabelalias{tab:hatt_worker_variables}{tab:roofline-worker-variables}%
  \legacylabelalias{tab:msa_variables}{tab:roofline-msa-variables}%
}
\makeatother

\usepackage{booktabs,subcaption}

\title{Subquadratic Attention with a Hierarchical Tree-Structured Indexer}%
\author{
Leon Lufkin*, Cutter Dawes*, Alexander Speicher, Kamesh Krishnamurthy, Tom\'as Figliolia, Beren Millidge%
\\[0.35em]
\zyphraaffiliation \quad \equalcontribution \quad %
\\[0.35em]
{\small Correspondence: \texttt{\{leon,kamesh,tom\}@zyphra.com}}
}

\begin{document}
\maketitle
\setcounter{page}{1}
\begin{abstract}%
The quadratic scaling of attention is a fundamental bottleneck limiting the long-context capabilities of large language models (LLMs). A variety of approaches have attempted to circumvent this, but they either fail to match attention's expressiveness (e.g., sliding-window or linear attention variants with fixed-size state) or only achieve constant-factor speedups (e.g., sparse attention methods whose indexers themselves scale quadratically). In this paper, we derive an asymptotically subquadratic sparse attention method based on a hierarchical, tree-structured indexer, which we call Zyphra Hierarchical Attention (ZHA). ZHA constructs a tree of recursively compressed key summaries over the KV cache, which future queries traverse from the top down to retrieve a sparse subset of keys. Crucially, by iteratively growing the tree's depth, the total cost including routing and final attention grows sublinearly in sequence length, resulting in overall subquadratic (but superlinear) scaling for attention. We demonstrate that ZHA can be cheaply applied to existing pretrained models with only a short KL distillation training stage of a billion tokens, and maintains performance on long-context retrieval and challenging reasoning, mathematical, and agentic coding evaluation tasks. Moreover, we show that our method is highly robust to tree geometry including very deep trees of up to 6 levels. Finally, we develop a detailed performance model and show that our approach scales subquadratically, dramatically outperforms competing methods beyond 1 million tokens of context, and ultimately has the potential to extend effective context windows to the hundreds of millions or billions of tokens.
\end{abstract}

\section{Introduction} \label{sec:introduction}

Attention is the fundamental sequence-mixing primitive in modern transformer architectures~\citep{vaswani2017attention}, and it is the basis for the highly flexible and expressive sequence modeling that underpins the recent rapid and dramatic advances in AI. Attention expresses an all-to-all mapping between sequence elements, where each token in the sequence is compared to every other token to identify those which are relevant. It is highly expressive and unstructured, yet parallelizable and efficiently implementable on GPUs. These properties have made it the predominant sequence-mixing architecture and enabled massive scaling of model size and sequence lengths.

However, the all-to-all nature of attention leads to fundamentally unfavorable scaling in computational and memory costs. In prefill, its quadratic computational cost dominates other operations, while during decode, re-reading the whole key-value (KV) cache per new token imposes memory bandwidth and storage costs which linearly increase in the sequence length. This has prohibited extremely long LLM contexts, despite continuing gains from longer reasoning and input lengths across a wide range of increasingly challenging tasks~\citep{deepseekr1,longbenchv2}.

This has motivated a variety of approaches that reduce the computational cost of attention~\citep{pisa}. State-space models and sliding-window attention \cite{dao2024transformersssmsgeneralizedmodels, gu2024mambalineartimesequencemodeling, lahoti2026mamba3improvedsequencemodeling, fu2025slidingwindowattentiontraining} replace the full KV cache with a fixed-size state that is updated along the sequence. While these eliminate the asymptotic cost of attention with linear prefill and constant decode cost, their fixed-size state saturates at long contexts, leading to worse performance~\citep{jelassi2024repeat}. Practical ``hybrid'' methods \citep{glorioso2024zamba,lieber2024jamba,glorioso2024zamba2,team2025kimi} thus often interleave a fixed-state mixer with full attention, reaping the computational benefits in most layers while restricting the attention costs to only a few necessary layers. However, at sufficiently long contexts, the few layers using full attention still dominate.

Other approaches aim to compress the sequence directly by online-compressing multiple tokens into a single token during prefill or decode (e.g., CSA~\citep{deepseekai2026deepseekv4highlyefficientmilliontoken}, NSA~\citep{NSA}, HNet~\citep{HNet}), or by compressing the KV cache by reducing the number of independent heads \citep{GQA,liu2024deepseek,figliolia2025compressed}. A related set of methods focuses on offline KV cache eviction, where tokens deemed to be of little value are evicted periodically from the KV cache, reducing the cost of all future queries \citep{kim2026fastkvzipefficientaccurate}. However, although head-compression methods such as GQA and MLA have become ubiquitous, they are fundamentally limited in possible compression rate by the number of heads. Meanwhile, sequence-compression and KV-eviction methods theoretically enable large compression rates but have fared less well empirically. Many of these methods obtain only a constant-factor speedup and are usually brittle and scale poorly in practice. A core reason for this brittleness is that it is practically impossible to know ahead of time, or on average, whether a token is useful or not.

This leads directly to a family of \emph{fine-grained sparse attention} methods that enable each query to retrieve its own sparse subset of keys and values from the context (e.g., DSA~\citep{DSV3.2}, MSA~\citep{MSA}). In these methods, a lightweight indexer attention scores every KV pair's relevance to each query, and full attention is performed only on the top-$K$ most relevant pairs. Since $K$ is fixed, the full attention has a constant cost. Crucially, this approach still permits each query to select an independent subset of keys most important to it, which enables higher precision for retrieval than compressing the same set of keys for all queries. However, this flexibility comes with two costs. First, the full KV cache must be stored, even if not computed over every token. Second, and most importantly, it relocates the cost of attention into the indexer. In practice, one can obtain significant speedups by making the indexer cheap with fewer heads or lower precision, but at contexts near 1M tokens, 60-90\% of time in these layers is still spent on indexing \citep{IndexCache,LongCat}. A subfamily of \emph{chunk-sparse} attention methods \citep{NSA,hils,dash} attempts to address this by compressing the indexer itself, and sparsely selecting chunks from the KV cache over which to perform attention. Of interest are the method of \citet{dash}, which uses an adaptive sparsity function $\alpha$-entmax, similar to our top-$p$ routing in spirit, and that of \citet{hils}, which constructs chunk summaries as an attention-weighted sum, similar to our own key construction method. These chunk-sparse methods perform chunking over a single level, yielding only constant-factor speedups. Multi-level methods such as PISA~\citep{pisa} and HiP~\citep{lee2025hip} achieve log-linear complexity, but their evaluations do not establish performance on the challenging mathematical reasoning and agentic coding tasks considered here, and show substantial retrieval degradation at longer contexts.

This leads to the conclusion that to construct an effective attention replacement yielding better than constant-factor speedups, two properties must be satisfied. The first property, derived from sparse attention, is \emph{query-dependent sparsity}, since different queries can require substantially different information from the KV cache. %
The second property is a \emph{growing but sublinear active state size}.
New sequence elements add predictive information indefinitely, so a fixed-size state will inevitably reach capacity and degrade in performance. By contrast, full attention can be viewed as a non-parametric state that grows linearly with sequence length. This linear growth is maximally expressive but highly redundant \citep{H2O,Scissorhands,PyramidKV,AdaKV}, since the predictive information of each new token diminishes: the millionth token contains information, but likely less than the first. The challenge is to design a state that can grow dynamically, but sublinearly, at the rate that predictive information accumulates in the sequence.

Motivated by these considerations,  we introduce a novel hierarchical sparse attention method, which we call Zyphra Hierarchical Attention (ZHA), that can achieve better than constant-factor speedups over dense attention.
By analyzing the structure of attention's representations, we find that the size of its effective state is sublinear and that it is amenable to a hierarchical indexer. Moreover, we show that we can adapt pretrained models to use our hierarchical indexer with distillation training of only a billion tokens and very few additional parameters, without changing any of the base model's weights. We convert models up to 35B parameters and observe little to no degradation on challenging evaluations covering mathematical reasoning, agentic coding, and long-context retrieval (though, for models without positional encoding, we do see drops on retrieval in line with prior work \citep{hils}). We also perform a detailed study of various indexer geometries and demonstrate that ZHA is highly robust to such variations and, crucially, can maintain performance at large depths.

Finally, we develop a detailed theoretical performance model of our method as well as comprehensive empirical sparsity results for a variety of indexer geometries. Our results demonstrate that ZHA can remain asymptotically subquadratic, subject to careful top-$p$ scaling, when extrapolated to extremely long sequences of tens of millions to billions of tokens, and can thus theoretically strongly outperform sparse attention methods at these context lengths in our performance model. We model ZHA's speedups against dense attention and MSA with a roofline model~\citep{williams2009roofline}, which predicts a crossover point against the latter around 0.5M tokens for a performant geometry with $p=0.75$.
\section{Motivation} \label{sec:motivation}

\subsection{The Natural Sparsity of Attention}
\label{sec:motivation-natural-sparsity}

Our work starts from the observation that attention's representations on language data already admit a sublinear state. Importantly, sublinear compressibility is not fundamental to the attention architecture but rather determined by the data and task distribution. Adversarial sequences consisting of random tokens from which arbitrary tokens must be retrieved uniformly throughout the sequence would require full quadratic attention to process. However, we observe that natural language sequences appear, on average, to follow a sublinear growth of predictive information with sequence length. 

We operationalize this notion by studying the size of the top-$p$ KV subset of global attention layers in a number of language models on a dataset of long-context fiction books. The top-$p$ set is the smallest subset of keys whose attention probabilities sum to at least $p$. Top-$p$ can be thought of as a denoiser for identifying attention's ``true'' receptive field. We can write the output of attention as
\begin{equation}
\begin{alignedat}{2}
\operatorname{Attention}(q,K,V)
=&\, p \cdot\operatorname{Attention}(q,K_p,V_p) \\
&+(1-p) \cdot \operatorname{Attention}
    (q,K_{\bar p},V_{\bar p}),
\end{alignedat}
\label{eq:motivation-top-p-decomposition}
\end{equation}
where $(K_p,V_p)$ denotes the top-$p$ subset for query $q$, and $(K_{\bar p},V_{\bar p})$ contains the remaining KV pairs. When performing top-$p$ attention, we keep only the first term in \cref{eq:motivation-top-p-decomposition}, without the prefactor $p$. If it is the case that only a subset of tokens are relevant for a query, it is still difficult for attention to set the other tokens' probabilities precisely to zero, but they should still have far smaller scores than the relevant subset. So, we can interpret the first term in \cref{eq:motivation-top-p-decomposition} as the true ``signal'' and the second term as noise. By choosing the appropriate $p<1$, we can extract the desired signal.

\begin{figure}[!t]
    \centering
    \includegraphics[width=\linewidth]{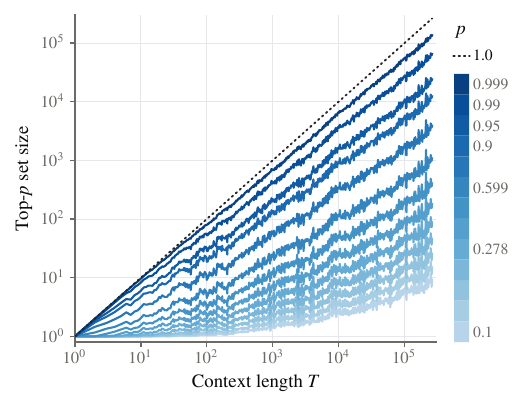}
    \caption{Dense top-$p$ set size versus causal context length in Qwen3.6-35B-A3B on six fiction books (averaged across books, layers, and heads). We observe that at all $p < 1$ (shades of blue), the top-$p$ set size increases sublinearly as a power law, indicating that longer sequences become increasingly compressible.}
    \label{fig:dense-top-p-growth}
\end{figure}

We find the size of the top-$p$ subset in each attention layer grows as a sublinear power law in the context length, with exponent depending on $p$ (see \cref{fig:dense-top-p-growth}). Appendix~\ref{app:top-p-dense-heads} shows top-$p$ curves broken down by layer and head.\footnote{Assessing whether an entire layer or model admits a power law is more conservative than studying a single attention head. Here, we wish to see that all the layers with significant attention exhibit sublinear growth in their top-$p$ subset sizes, establishing that compressibility extends across the model rather than being confined to isolated heads.} \Cref{tab:rollout-linear-preference} provides a complementary comparison on ZHA rollout selection counts, where power laws are preferred in many configurations but not uniformly across checkpoint geometries and query-tile widths. These fits describe router-selected leaf unions, rather than dense top-$p$ sets.

That this sublinear growth is a power law already suggests hierarchical structure, although it does not prove this structure is efficiently searchable in practice. Power laws are typically associated with scale invariance, that is, an underlying representation that remains the same at every level of analysis. Here, this would likely correspond to the statistics of some summary of tokens remaining the same as we summarize groups of tokens of geometrically increasing size. If we could learn such statistics and use them for routing, then perhaps we could route recursively through these geometric groups at increasing levels of precision according to a simple decision rule.

We note that this power-law structure does not always hold, and it seems to depend on positional encoding. Qwen3.6-35B-A3B uses partial RoPE, whereas Nemotron-3.5-Lightning-30B-A3B, which we study in \cref{fig:dense-top-p-nemotron-heads}, uses NoPE and does not show as robust power-law top-$p$ scaling (and shows larger retrieval losses after conversion to ZHA; see Appendix~\ref{app:evaluation-nemotron}).

\subsection{Sublinear Retrieval through Hierarchical Search}

That attention for an individual query is sparse does not imply that most KV entries can be permanently discarded.  The relevant subset can change with the query such that keys receiving negligible attention for a given query may nevertheless become important for a later query. Sparse individual attention distributions can therefore coexist with broad coverage of the cache across the union of future queries.

To assess whether sparse attention supports permanent KV eviction, we measure the union of the top-$p$ subsets over increasing numbers of future queries, repeating this analysis for several values of $p$. More precisely, among sequences from long fiction books of length $T=261{,}548$ tokens, we consider the window of the $W$ most recent queries ($1\le W<T$) and study 
\[
U_p(W)=\frac{1}{T-W}\left|\bigcup_{q=T-W}^{T-1}S_p(q)\right|,
\]
where $S_p(q)$ is the top-$p$ subset of indices (using full causal attention) for the query at position $q$, but this subset of indices is restricted to the prefix $\{0,\ldots,T-W-1\}$ after selection, without renormalization. Thus, $U$ is the fraction of the prefix covered by the union of all the top-$p$ sets across all $W$ most recent queries.

\begin{figure}[!b]
    \centering
    \includegraphics[width=\linewidth]{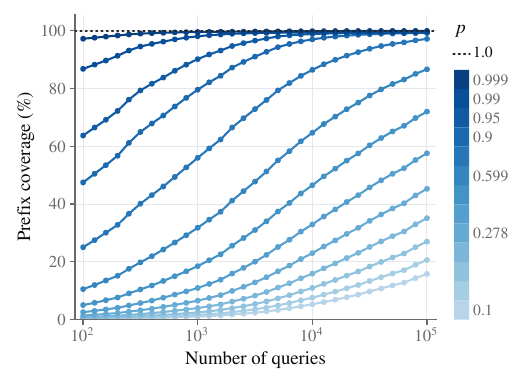}
    \caption{Fraction of KV entries of a 256K sequence covered by the union of top-$p$ attention subsets as the number of future queries increases, for several values of $p$. We observe that for the higher $p$ values we use in practice, a substantial fraction of the total KV cache is included in the union of the queries. }
    \label{fig:top-p-query-union}
\end{figure}

If $U_p(W)$ is near 1, this suggests that nearly all keys prior to time $T-W$ are in the top-$p$ set for some query at or after time $T-W$. We see in \cref{fig:top-p-query-union} that this is inevitably the case as $W$ grows large. That is, the union expands to cover the entire cache, indicating that entries receiving negligible attention from some queries may still be relevant to others.  These results motivate retaining the full KV cache while retrieving a query-dependent subset for each attention computation.

If we are to identify the top-$p$ subset asymptotically faster than quadratic attention, then the work needed to identify relevant entries must grow sublinearly with context length, i.e., we cannot compare every query directly to every prior key. A hierarchical, geometric tree-structured indexer offers a natural mechanism for addressing this search problem. Coarse summaries guide the query toward promising groups of keys.
If a group of keys is deemed relevant via its summary, then those keys are scored again at a higher resolution, while if they are not deemed relevant, then they can all be discarded together without needing to examine them individually. This can be an effective way to reduce the number of query-key comparisons, provided that the summaries can sufficiently capture relevance.

It is not obvious a priori that a tree is expressive enough to identify the top-$p$ subset with sufficient accuracy, since exact computation of this subset requires scoring the entire KV cache. However, we saw in \cref{sec:motivation-natural-sparsity} that the top-$p$ subset specifically follows a power law, which suggests an underlying tree-like structure that we aim to learn.

\section{Architecture} \label{sec:architecture}

ZHA is designed to approximate quadratic attention despite subquadratic computational complexity. Our approach involves first constructing a tree of recursively compressed keys, each of which summarizes a chunk of keys in the level below it (which may themselves be compressed). Compression is performed using chunk-wise attention scores from trained reference queries. Then, once the tree is constructed, we route through the tree with a given query by selecting the top-$p$ nodes at a given level to expand. 

\begin{figure*}[t]
    \centering
    \includegraphics[width=0.9\textwidth]{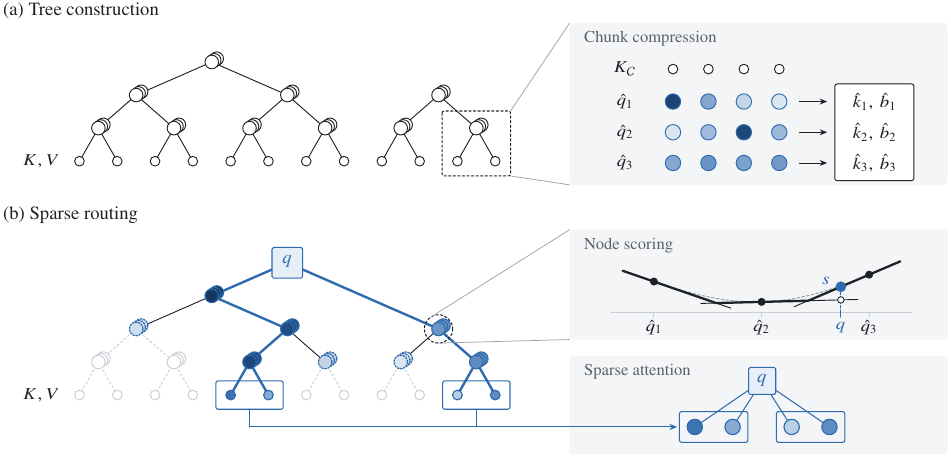}
    \caption{Overview of the ZHA architecture: (a) tree construction and (b) sparse routing. During tree construction, for each chunk of contiguous key nodes, we construct a compressed key node at the higher level by an attention-weighted sum of the keys, where the attention is computed with respect to the learned reference queries. In practice, we use multiple reference queries and store the key nodes for each in parallel. This process continues recursively with higher-level key nodes being constructed from the summaries of lower levels. During routing, we start at the top of the tree and compute dot-product similarity scores between our query and the key nodes of that level. We select the set of nodes to expand using top-$p$, and then recursively descend the tree until we reach the leaf nodes corresponding to raw KV cache entries. The union of all the leaves we reach defines the sparse KV cache subset that we dispatch to the main attention.}
    \label{fig:architecture-overview}
\end{figure*}

We train just the hierarchical indexer, whose only learnable components are a small set of reference queries. \emph{ZHA may therefore be applied to existing pretrained models with only a small distillation phase}; in fact, we find that a short KL warmup of only a billion tokens is sufficient to match the base model's performance. ZHA does not require finetuning the full model, and can support a wide range of attention configurations such as different positional encodings \citep{su2024roformer}, gating \citep{qiu2026gated}, and head-structures such as GQA, MLA, or CCA \citep{GQA,liu2024deepseek,figliolia2025compressed}. An overview of the ZHA architecture is shown in \cref{fig:architecture-overview}.

\subsection{Tree Construction}

To construct the tree, ZHA recursively creates key summaries that compress contiguous chunks of keys in the level below it (which themselves are either key summaries or, at the bottom of the tree, the original keys). During decode, we construct the key summaries immediately upon generation of a complete chunk of keys, maintaining a buffer until sufficient keys have accumulated to construct that chunk's summary. Supposing the chunk size is $C$, this implies that we add a compressed key to the first level of the tree every $C$ decode steps, to the second level every $C^2$ steps, and so on. Our approach is intentionally lightweight, so that tree construction adds negligible latency amortized across the sequence (around 1\% or less in our roofline model).

To perform well, it is vital that the compressed keys are a faithful summarization of the underlying chunk of keys. Thus, we aim to directly approximate a chunk's total attention mass over a wide distribution of queries. To perform this approximation, we use the observation that the log unnormalized attention mass of a chunk is a convex function of the query that is naturally approximated by a piecewise-affine tiling of tangent planes. In particular, we take the maximum over a set of tangents centered at some point in the query distribution (these become our reference queries), each of which is cheaply computable by appropriately attention-weighting the keys and measuring the entropy of the attention weights at that point.

\begin{figure}[H]
    \centering
    \includegraphics[width=\columnwidth]{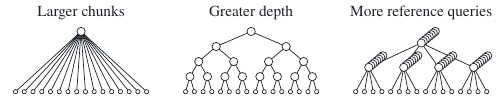}
    \caption{Illustrative ZHA tree geometries of the three core tree parameters. We can trade off depth and width for a given level of compression in the tree, since compression scales as $C^L$. Secondly, at the cost of additional compute, we can increase the number of reference queries. }
    \label{fig:architecture-geometry}
\end{figure}

To make this concrete, denote a contiguous chunk of keys by $K_C = [k_0, \dots, k_{C-1}]$ with chunk size $C$, and let $R$ be the number of tangents and $d$ the head dimension. Its log unnormalized attention mass for a given query is the log-sum-exp (LSE) of its scaled attention scores:
\[
    F_C(q) := \operatorname{LSE}_{j\in [C]}\!\left(q^\top k_j/\sqrt d\right)
    = \log\sum_{j\in [C]}\exp(q^\top k_j/\sqrt d).
\]
The first-order Taylor approximation of $F_C$ at some reference query $\hat q_r$ gives the tangent $t_r(q) = q^\top \hat k_r/\sqrt d + \hat b_r$, where the compressed key and bias are
\begin{align*}
    \hat k_r &= \operatorname{Attn}(\hat q_r, K_C, K_C) 
              = \sum_{j \in [C]} \frac{\exp(\hat q_r^\top k_j/\sqrt d) k_j}{\sum_{j' \in [C]} \exp(\hat q_r^\top k_{j'}/\sqrt d)}, \\
    \hat b_r &= F_C(\hat q_r) - \hat q_r^\top \hat k_r/\sqrt d.
\end{align*}
Provided a set of tangents representing a given chunk, its LSE for a future query $q$ is approximated by $\max_{r \in [R]} t_r(q)$. To summarize a complete node at level $\ell>1$ in the tree, which represents $C^\ell$ original keys, we apply the same construction to its $C$ child summaries for each reference query, including the child biases in the attention scores and adding their attention-weighted biases to the entropy of the merge weights.\footnote{ When the same reference query is used throughout a subtree, this merge recovers the summary computed directly from its original keys. With chunk-dependent RoPE rotations, it instead gives an affine lower bound that need not be tangent to the original chunk's log attention mass.} For models with GQA or other KV-head groupings, we construct one tree per KV head, shared by the query heads in its group. To provide some intuition for the variety of geometries, \cref{fig:architecture-geometry} illustrates trees varying chunk size, depth, and number of reference queries.

\subsection{Training the Module}

Our chunk summarization technique relies upon a suitable choice of reference queries.  Our approach is to learn them during a short KL-warmup phase; these are in fact the only learnable component in our ZHA mechanism. We find it sufficient to learn a single set of $R$ reference queries per tree, shared by all nodes across the sequence, at all levels in the tree, and, in the case of GQA, by the query heads sharing the tree. With a reasonable number of reference queries $R$, the ZHA parameters add a negligible increase to the parameter count of the models we test.\footnote{ For example, with R16 and one tree per KV head in Qwen3.6-35B-A3B, the parameter count increases by approximately $0.00024\%$ (81,920 / 34.66B).} For models that incorporate RoPE, we rotate the shared reference queries to the last token position of each node's span, at every tree level. Let $\hat q_{r,c}$ denote the reference query $r$ rotated for node $c$. Before merging the children of a parent node $P$, we evaluate all child summaries at the parent's rotated reference query $\hat q_{r,P}$. Each child $c$ was originally summarized at its own rotated reference query $\hat q_{r,c}$. To evaluate its affine approximation at the parent's query, we add $\tfrac{1}{\sqrt d}\langle \hat k_{r,c}, \hat q_{r,P}-\hat q_{r,c}\rangle$ to its estimated log attention mass. This correction is simply the compressed key's score at the parent's reference query minus its score at the child's reference query.

From a trained dense-attention model, the ZHA conversion procedure is remarkably simple and compute-efficient. In particular, we freeze the base model (including all native attention components) and learn the reference query parameters by matching the predicted chunk probabilities to the true attention distribution pooled into chunks at each level. For a query $q$, let $a_j(q)$ denote its dense attention weight on key $j$, and let $\mathcal{P}_\ell(q)$ partition the original key indices visible to $q$ into chunks $\mathcal{C}$ at tree level $\ell$.
The target and predicted chunk probabilities are
\[
    p_{\ell,\mathcal{C}}(q) = \sum_{j\in \mathcal{C}} a_j(q),
    \quad
    \hat p_{\ell,\mathcal{C}}(q) = \frac{\exp s_{\ell,\mathcal{C}}(q)}{\sum_{\mathcal{C}'\in\mathcal{P}_\ell(q)}\exp s_{\ell,\mathcal{C}'}(q)},
\]
where $s_{\ell,\mathcal{C}}(q)=\max_{r\in[R]}t_{\ell,\mathcal{C},r}(q)$ is the chunk's approximate log attention mass.\footnote{ For the current incomplete chunk, we use its exact log attention mass in place of $s_{\ell,\mathcal{C}}(q)$.} We supervise the shared reference queries at every tree level by minimizing
\[
    \mathcal{L}_{\mathrm{KL}} = \sum_{\ell=1}^{L}\mathbb{E}_q\!\left[D_{\mathrm{KL}}\!\left(p_\ell(q)\,\middle\|\,\hat p_\ell(q)\right)\right].
\]
Here $L$ is the number of tree levels, and the expectation averages over eligible queries, attention heads, and model layers at each tree level.
In practice, learning the reference query parameters requires only a billion tokens, and can be done at sequence lengths significantly shorter than the pretrained backbone's full context length.\footnote{ For the reported Qwen3.6-35B-A3B and Nemotron-3.5-Lightning-30B-A3B runs, we trained for 1B tokens at 32K context length, using a peak learning rate of $3.2\times10^{-2}$ and the WSD schedule described in Appendix~\ref{app:training-compute}.}

\subsection{Routing Through the Tree}

Having constructed the tree, we must now sparsely traverse it during inference. A given query retrieves from the tree by scoring nodes and selecting a subset of nodes at each level, recursively descending by repeating this process for the selected nodes' children, until terminating at a sparse subset of the original KV, from which we compute the attention output.

A core feature of our approach is the combination of learned attention-mass summaries with adaptive top-$p$ selection at every level of the hierarchy (described in \cref{sec:motivation}).
Top-$p$ selection, previously applied to sparse attention by \citet{lin2025twilight}, allows the selection size to adapt to the attention distribution and sequence length, in contrast to fixed-selection methods such as top-$K$. If only a few nodes contain most of the mass, top-$p$ is not obliged to select marginal nodes to satisfy a fixed $K$ budget. Conversely, if attention mass is near-uniform, a large fraction of the nodes at a given level can be selected for expansion. We posit that this property enables length extrapolation beyond that seen in other methods, which we confirm for two open-source models in \cref{sec:evaluation}. The hierarchy also keeps the ordering cost of top-$p$ selection manageable. Pruning at each level limits the number of candidate nodes considered at the next, reducing the size of subsequent ordering operations. In practice, we find that these candidate sets remain small enough to make top-$p$ selection feasible at every level of the tree.

The concrete routing procedure is as follows.
Each $q$ denotes one query head's query; selection and child expansion remain independent across query heads, including those sharing a KV tree. Let $\mathcal{A}_\ell(q)=\{\mathcal{C}_1,\ldots,\mathcal{C}_{n_\ell}\}$ denote the complete candidate chunks considered at level $\ell$, indexed in decreasing order of score $s_{\ell,\mathcal{C}}(q)$. Note that the ordering and $n_\ell=|\mathcal{A}_\ell(q)|$ vary depending on $q$, and there may be incomplete chunks at each level in the tree depending on the exact sequence length. At each level, we pick the top-$p$ nodes; i.e., for chosen $p \in (0, 1)$, we select chunks
\[
    \mathcal{S}_\ell(q)=\{\mathcal{C}_1,\ldots,\mathcal{C}_{m_\ell(q)}\},
\]
where
\[
    m_\ell(q)=\min\left\{m\in[n_\ell]:\frac{\sum_{i=1}^{m}\exp s_{\ell,\mathcal{C}_i}(q)}{\sum_{i=1}^{n_\ell}\exp s_{\ell,\mathcal{C}_i}(q)}\ge p\right\}.
\]
Thus, the selected chunks form the smallest prefix containing at least a fraction $p$ of the estimated attention mass among the current candidates.
Let $\mathcal{U}_\ell(q)$ denote the complete chunks at level $\ell$ with no complete parent. Starting with all complete top-level chunks, $\mathcal{A}_L(q)=\mathcal{U}_L(q)$, we pass the selected chunks' children and any parentless chunks to the next level:
\[
    \mathcal{A}_{\ell-1}(q)=\left(\bigcup_{\mathcal{C}\in\mathcal{S}_\ell(q)}\operatorname{ch}(\mathcal{C})\right)\cup\mathcal{U}_{\ell-1}(q),
    \qquad \ell=L,\ldots,2,
\]
where $\operatorname{ch}(\mathcal{C})$ denotes the child chunks of $\mathcal{C}$. We exclude the first key (index $0$) from the tree construction and instead always retain it\footnote{In many models, the first token acts as an attention sink, receiving a large share of attention mass regardless of content; we found it beneficial to always attend to it.} together with the visible keys in the current chunk, $\mathcal{C}_{\mathrm{cur}}(q)$, giving the token indices
\[
    \mathcal{I}(q)=\{0\}\cup\left(\bigcup_{\mathcal{C}\in\mathcal{S}_1(q)}\mathcal{C}\right)\cup \mathcal{C}_{\mathrm{cur}}(q).
\]
During decode, query heads sharing a KV head jointly load the union of their final selected leaf chunks; this sharing does not merge their upper-level routing selections or attention masks.\footnote{For block prefill, loads are shared across query positions within each head's tile, with per-query masks retained.} Each head computes standard scaled dot-product attention over its own selected KV:
\[
    o(q)=\operatorname{Attn}\!\left(q,K_{\mathcal{I}(q)},V_{\mathcal{I}(q)}\right).
\]
\section{Evaluations} \label{sec:evaluation}

We evaluate ZHA on a suite of reasoning and retrieval tasks with context lengths extending to hundreds of thousands of tokens, demonstrating that the module can match the dense model while attending to a small fraction of the total KV cache. Across these settings, ZHA is impressively robust to choice of geometry, whose effect on retrieval accuracy is nonetheless highly predictable by a simple model. These results suggest that ZHA enables efficient retrieval from past context and that this retrieval remains effective at very long contexts. 

\subsection{Reasoning and Retrieval}

\begin{figure*}[!t]
    \centering
    \includegraphics[width=\textwidth]{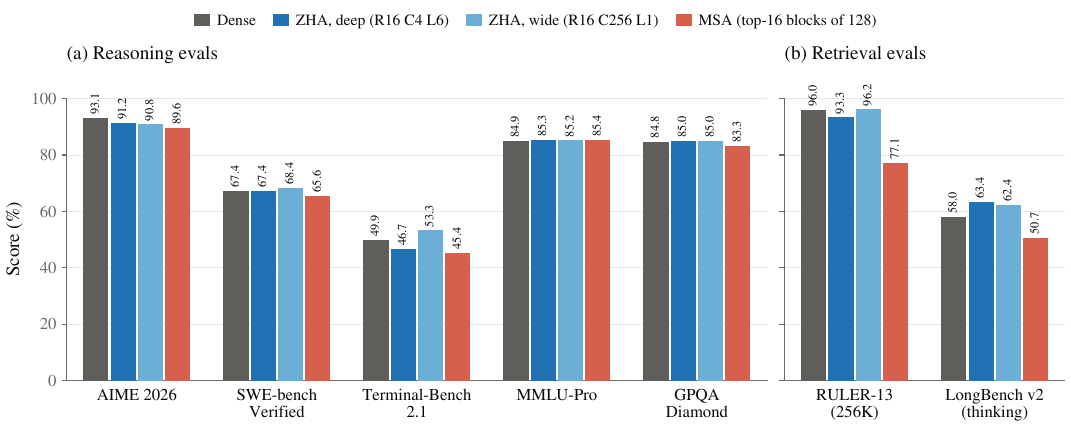}
    \caption{(a) Reasoning and (b) retrieval evaluation scores with dense attention, MSA, and deep (C4~L6) or wide (C256~L1) ZHA ($p=0.95$, R16, converted from Qwen3.6-35B-A3B). Appendix~\ref{app:training-evaluation} gives evaluation details and results across routing thresholds. We observe that ZHA with both wide and deep geometries maintains performance even at high sparsity levels, while MSA suffers performance degradation especially on retrieval tasks such as RULER and LongBench~v2. }
    \label{fig:evaluation-qwen}
\end{figure*}

For the primary suite of evaluations, we convert Qwen3.6-35B-A3B \citep{qwen36model} with ZHA routers of varying tree geometry. To start, we train both deep (chunk size 4, six tree levels; C4 L6) and wide (chunk size 256, one tree level; C256 L1) routers, each with 16 reference queries (R16) and trained on 1B tokens with the backbone frozen (for more training details, please see Appendix~\ref{app:training-compute}). For our evaluations, we vary $p \in \{0.8, 0.9, 0.95\}$, constant at every tree level. Both routers perform within noise of the dense baseline on reasoning, agentic coding, and long-context benchmarks. The exception is RULER-13 for the deep router, which scores 93.3\% versus 96.0\% for dense; however, this is just a 3\% drop in performance while attending to 1\%-6\% of the original KV cache. See \cref{fig:evaluation-qwen} for the full suite of evaluation results for ZHA (deep and wide) and dense.

We also compare ZHA against MiniMax Sparse Attention (MSA)~\citep{MSA}, which selects the top 16 blocks of 128 tokens using a learned indexer. We use the same Qwen backbone and train MSA's indexer on 40B tokens using only the KL warmup stage, with the backbone frozen. MSA stays close to the deep and wide ZHA routers on AIME 2026, MMLU-Pro, and GPQA Diamond, but falls well behind both routers on long-context retrieval. On LongBench~v2, where MSA's fixed budget amounts to roughly 1\% of the context, the deep router attends to a similar 1.1\% of decode keys yet scores 12.7 percentage points higher, even though MSA was trained on 40x the token budget of ZHA. For comparison, MSA is also included in \cref{fig:evaluation-qwen}.

Finally, we also train ZHA onto Nemotron-3.5-Lightning-30B-A3B \citep{nemotron35}, to assess applicability to an architecture without explicit positional encodings. Generally Nemotron performs comparably to its baseline on the agentic coding benchmarks, but suffers moderately on the RULER benchmark. Full results for Nemotron are reported in Appendix~\ref{app:evaluation-nemotron}.

\subsection{Effect of Geometry and Sparsity} \label{sec:evaluation-geometry}

We have shown that ZHA is competitive with dense attention, whether the tree is wide (with chunk size 256) or deep (with 6 levels), and that these variants remain competitive even while attending to a fraction of the native KV cache. However, the space of tree geometries, spanned by choice of reference count $R$, chunk size $C$, tree depth $L$, and routing threshold $p$, is combinatorially complex. Despite preliminary evidence of robustness, it is easy to see why variation along any of the axes above may have a direct effect on performance: each changes the effective compression and sparsity in different ways, and each may introduce error in the router's approximation to the true top-$p$ set. Moreover, depth and width introduce potentially qualitatively different kinds of error into the approximation. Here, we explore the effect of geometry and sparsity along a few particularly revealing slices of the design space; then, to demonstrate the predictability of performance as a function of these variables, we develop a simple yet accurate model of RULER accuracy in terms of $(R, C, L, p)$ that we fit with just four parameters.

\begin{figure*}[!t]
    \centering
    \includegraphics[width=\textwidth]{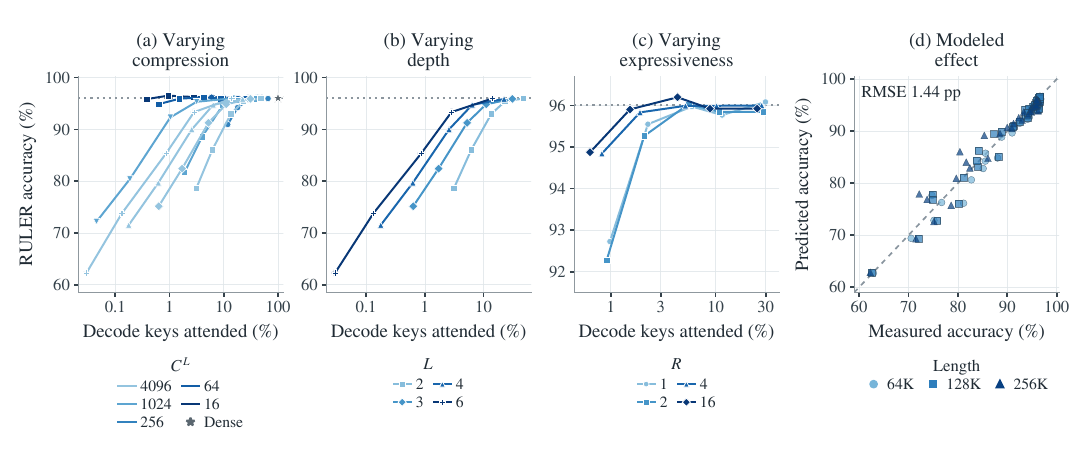}
    \caption{Geometry and sparsity have predictable effects on RULER accuracy (mean across tasks). Greater sparsity without degrading performance is achieved by: (a) decreasing compression $C^L$, (b) increasing depth at fixed compression ($C^L=4096$), or (c) making key summaries more expressive (with increasing reference-query number $R$) at fixed chunk size and depth. (The horizontal dotted lines in panels (a)--(c) show dense attention.) (d) For the simple geometry-to-performance model, predictions versus measurements across 64K, 128K, and 256K, holding out each length in turn (RMSE: 1.44 percentage points).}
    \label{fig:geometry-ruler-model}
\end{figure*}

\paragraph{Robustness across geometries.} To more thoroughly probe the effect on performance, we sweep a wide range of geometries by varying $R$, $C$, and $L$ (this yields 28 different configs; see \cref{fig:geometry-coverage} in Appendix~\ref{app:geometry} for the 28 used for fitting). For each, we use the same 1B-token KL warmup and then evaluate on RULER (at lengths of 64K, 128K, and 256K) while varying $p \in \{0.7, 0.8, 0.9, 0.95, 0.99\}$. Across 13 tasks at 256K, the aggregate RULER accuracy as a function of percentage of decode keys attended is shown in \cref{fig:geometry-ruler-model}(a), color-coded by the top-level node span $C^L$, our nominal compression factor. The primary observation is that, when attending to between 10\% and 30\% of keys, retrieval performance is robust to choice of geometry. Beyond this, performance generally degrades with increasing key sparsity, with the rate depending on both top-level compression and its decomposition into $C$ and $L$. \Cref{fig:geometry-ruler-model}(c) isolates the effect of reference count $R$ at fixed C8 and L2. \Cref{fig:geometry-ruler-pareto} in Appendix~\ref{app:geometry} shows the 256K RULER Pareto frontiers across geometries for decode keys attended and modeled speedup over dense.

\paragraph{Depth at fixed compression.} Because ZHA's subquadratic scaling properties rely on increasing tree depth, a critical question for ZHA's promise is whether depth specifically leads to sharp degradation in retrieval. To isolate the effect of tree depth $L$ from top-level compression $C^L$, \cref{fig:geometry-ruler-model}(b) shows RULER accuracy versus decode keys attended for a slice of ZHA geometries with fixed $C^L = 4096$. In fact, it appears that increasing depth enables sustaining effective retrieval out to greater sparsity and that the deeper geometries are more performant at a fixed sparsity level. However, this alone does not suggest that ZHA can indefinitely increase depth while scaling to longer contexts (as this increases top-level compression $C^L$ at fixed $C$), but it does provide strong evidence that depth itself is not the limiting factor. Notably, to our knowledge, our approach is the first hierarchical attention variant to have demonstrated convincing scaling to depths greater than 1 while maintaining close parity to dense attention on challenging mathematical reasoning, agentic coding, and long-context retrieval evaluations. 

\paragraph{Predicting retrieval accuracy.} To quantify and generalize the accuracy-sparsity tradeoff, we develop a simple model predicting RULER accuracy from input $(R, C, L, p)$.
The model is
\begin{equation}
\begin{aligned}
    d_\ell(R,C) &= \log\!\left(1+\frac{C^\ell}{R^r}\right),\\
    \widehat A &= A_{\mathrm{dense}}(N)\,
    \prod_{\ell=1}^{L}\exp\!\left[-a(1-p)^b d_\ell^{h}\right].
\end{aligned}
\label{eq:geometry-accuracy-model}
\end{equation}
Intuitively, the term $\log(1+C^\ell/R^r)$ represents compression difficulty at level $\ell$, increasing with node span and decreasing with reference capacity. The product compounds per-level accuracy retention over the tree.
The factor $(1-p)^b$ accounts for the effect of sparsity, whose penalty can grow nonlinearly.\footnote{We constrain $1\leq b\leq2$ to retain flexibility while avoiding unstable low-$p$ extrapolation for near-ceiling tasks.} We fit four parameters: $a$, which specifies the overall penalty strength; $b$, which controls sensitivity to sparsity; $h$, which controls sensitivity to each level's difficulty; and $r$, which controls reference capacity. Each exponential maps a level's penalty to an accuracy-retention factor; their product scales the dense-attention accuracy $A_{\mathrm{dense}}(N)$ at that sequence length ($N$). Aside from length-dependence inherited from this baseline, sequence length does not enter this simple model.

We fit the model using 28 geometries from the sweep across $R\in\{1,2,4,16\}$, evaluated for 100 RULER examples per task at $p\in\{0.7,0.8,0.9,0.95\}$ and lengths (64K, 128K, 256K), together with $p=0.99$ evaluations for 13 of these geometries. Fitting uses settings complete at all three lengths; the empirical panels include complete 256K settings. We fit the four parameters (constraining $a>0$, $1\leq b\leq2$, $0.2\leq h\leq6$, and $0\leq r\leq6$) by nonlinear least squares on accuracy (in percentage points). Here we fit to the 13-task mean, and Appendix~\ref{app:geometry} fits each task independently with the same form. To evaluate held-out prediction accuracy (for the aggregate and per-task fits separately) and particularly length transfer, we measure predicted versus actual RULER accuracy on a held-out sequence length while fitting on the other two lengths; we do this separately for the three sequence lengths, and for each we fit on 250 cells (125 geometry-threshold settings at 2 lengths) and evaluate on 125 cells. Each fold reuses the same geometries and thresholds at the other lengths and uses the measured dense baseline at the held-out length, so this tests transfer of the relative accuracy penalty across lengths conditional on that baseline. \Cref{fig:geometry-ruler-model}(d) shows all 150 R16 leave-one-length-out predictions pooled across the three lengths, with an RMSE of 1.44 percentage points (the 256K subset has an RMSE of 1.65 percentage points).
The relatively low prediction error of such a simple model suggests that the accuracy varies surprisingly smoothly with sparsity and tree geometry over the fitted range. 
\section{Speedup} \label{sec:speedup}

ZHA is designed to achieve fundamentally subquadratic latency scaling, in contrast to the variety of chunk-sparse methods that yield significant but constant-factor speedups. Yet, however principled this design, its practicality depends on whether it can run efficiently on GPUs. While we do not release optimized kernels in this work, we do investigate the question of efficient GPU implementation via a detailed roofline model that enables extrapolation of ZHA's latency scaling in comparison to dense attention and other baselines.

\begin{figure*}[!tbp]
    \centering
    \includegraphics[width=\textwidth]{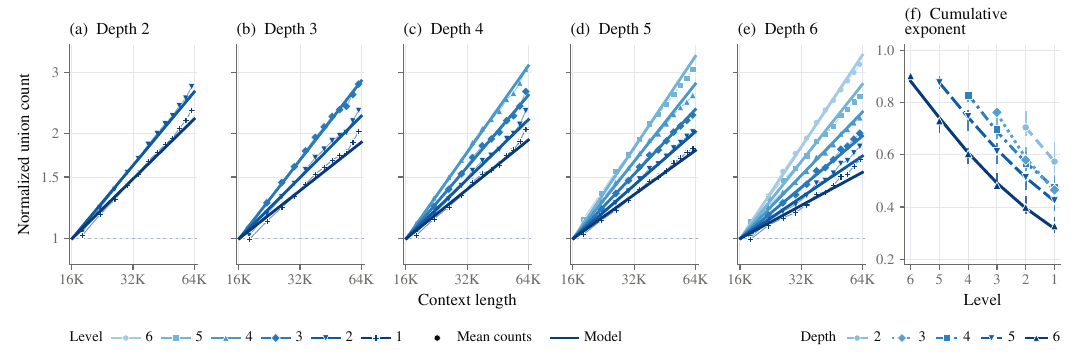}
    \caption{Per-level selection scales sublinearly in the ZHA router with predictable power-law exponents. (a)--(e): For ZHA geometries with depths L2--L6 (C4, R16, $p=0.8$), per-level selection counts (normalized by first-bin count per-level) on AIME problems fitted by power-law scaling curves. (f): Per-level exponents fitted by the cumulative top-down model, demonstrating that the across-depth variation in their exponents is well-described by a simple model that accumulates top-down contributions.}
    \label{fig:sparsity-scaling}
\end{figure*}

\subsection{Roofline Model}

Although we leave the full details to Appendix~\ref{app:roofline} (which provides a treatment of the top-$p$ selection algorithm among other specifics), we briefly describe the roofline model we use to estimate how ZHA latency scales with sequence length. Suppose $T$ is the context length, $L(T)$ the ultimate tree depth, and $\ell$ the routing level. The modeled ZHA attention-subsystem latency $\bm{t}_{\mathrm{ZHA}}$ is given by a sum reflecting its serial execution through the tree:
\begin{equation}
    \begin{aligned}
        \bm{t}_{\mathrm{ZHA}}(T) &= t_{\mathrm{tree}}(T)+\lambda_A+\bm{\Phi}_{\mathrm{attn}}(T)\\
        &\quad+\sum_{\ell=1}^{L(T)}
        \left[\lambda_\ell+\bm{\Phi}_{\mathrm{routing},\ell}(T)
        +\bm{\Phi}_{\mathrm{scan},\ell}(T)\right].
    \end{aligned}
    \label{eq:speedup-overall-model}
\end{equation}
Above, $t_{\mathrm{tree}}$ accounts for prefill construction or amortized decode maintenance, including launches. $\lambda_\ell$, $\lambda_A$ measure the per-level and final-attention launch overheads, respectively. The three kinds of execution stages are reference scoring $\bm{\Phi}_{\mathrm{routing},\ell}$, node selection $\bm{\Phi}_{\mathrm{scan},\ell}$, and retained-token exact attention $\bm{\Phi}_{\mathrm{attn}}$.

Each constituent kernel or pass $\bm{\Phi}_x$ (indexed by $x$) follows a three-term roofline model accounting for compute and memory bandwidth along with a worker-wave term $\bm{t}_{x,\mathrm{worker\ waves}}$ that accounts for finite block concurrency:
\begin{equation}
    \bm{\Phi}_x(T)=\max\left\{
    \begin{aligned}
        &\mathbf{FLOPs}_x(T)/P_x,\\
        &\mathbf{bytes}_x(T)/\mathrm{BW}_x,\\
        &\bm{t}_{x,\mathrm{worker\ waves}}(T)
    \end{aligned}\right\}.
    \label{eq:speedup-stage-roofline}
\end{equation}
Above, $\mathbf{FLOPs}_x$ denotes the floating-point operation count and $P_x$ the effective FLOP/s (scalar for scan); similarly, $\mathbf{bytes}_x$ denotes the HBM traffic and $\mathrm{BW}_x$ the effective bytes/s. For stages with sequential kernels or passes, the execution times are summed.

Though the compute and memory terms are familiar to any roofline model, it is worth noting why the worker-wave term is of particular importance for ZHA. This term is
\begin{equation}
    \bm{t}_{x,\mathrm{worker\ waves}}(T)
    =\bm{\tau}_x(T)\left\lceil
    \frac{\bm{N}_{\mathrm{tiles},x}(T)}{N_{\mathrm{SM}}c_x}
    \right\rceil,
    \label{eq:speedup-worker-waves}
\end{equation}
where \(\bm{N}_{\mathrm{tiles},x}\) is the number of thread blocks, \(N_{\mathrm{SM}}c_x\) the number that can execute concurrently, and \(\bm{\tau}_x\) the per-block execution time at that concurrency. The ceiling therefore counts the sequential waves needed to execute the stage. This matters for ZHA because pruning at one level reduces work at the next, potentially leaving individual stages with limited parallelism and thus idle SMs in the higher levels of the tree.

\subsection{Sparsity Scaling}

Predicting latency therefore requires understanding how the number of nodes selected at each level scales with sequence length. Extrapolating this behavior to longer sequences requires observing how the selected top-$p$ set scales at each level, and then modeling how this scaling varies across levels, particularly as we increase the tree's ultimate depth.

\paragraph{Empirical per-level exponents.}
Inspired by the robustly sublinear scaling observed in the top-$p$ selection set for dense models (see \cref{sec:motivation}), we fit a power law to the number of nodes selected per level. We fit these curves with per-level KV-group union counts collected across heads and layers of ZHA generations to AIME 2026 questions (response lengths 16K--64K), with geometries of C4, R16, and depths L2--L6 (a subset of the Qwen3.6-35B-A3B conversions from \cref{sec:evaluation-geometry}).

For a fixed geometry and routing threshold $p$, let $S_\ell(T)$ denote the mean selected-node count per query at context length $T$ and level $\ell$, counting the union within each KV group at every level and averaging over groups (with $\ell=L$ at the top and $\ell=1$ at the leaf chunks). Our power-law summary $\widehat{S}_\ell(T)$ has the form
\begin{equation}
    \widehat{S}_\ell(T)=c_\ell T^{\alpha_\ell},
    \label{eq:sparsity-selection-power-law}
\end{equation}
where $c_\ell$ is a constant factor and $\alpha_\ell$ is the scaling exponent that we fit to the empirical selection counts.\footnote{In particular, we estimate observed exponents $\alpha_\ell$ by fitting each problem's log counts against log context length with $1/T$ weights, then taking the median slope across problems at each level. See Appendix~\ref{app:sparsity-scaling} for more details.} \Cref{fig:sparsity-scaling}(a)--(e) shows the empirical number of selected nodes for the (color-coded) levels across fixed-$C$, different-depth geometries. We also plot lines denoting each level's fitted depth-model power law, demonstrating good quality of fit to the selection counts.

\paragraph{Modeled per-level exponents.}
In order to extrapolate ZHA's latency to longer sequences with increasing tree depth, we develop a model for how the exponent varies per-level. Using the same variable-depth geometries as a case study, we begin by modeling per-level exponents given some fixed ultimate depth $L$. Notice that, in \cref{fig:sparsity-scaling}, the scaling exponent is highest at the top level and decreases proceeding down each tree, consistently for each geometry.\footnote{\Cref{fig:sparsity-scaling} shows $p=0.8$; the corresponding data for $p \in \{0.7, 0.9, 0.95\}$ may be found in \cref{fig:sparsity-thresholds} in Appendix~\ref{app:sparsity-scaling} and broadly corroborates these same trends. The exception is the deepest geometry (C4, L6, R16) at $p=0.7$, whose exponents rise again toward the leaf levels. This setting achieves only 44.0\% accuracy on AIME, compared with 93.1\% for dense attention, so its anomalous scaling should be interpreted in the context of severe accuracy degradation.} This pattern inspires our approach to modeling the per-level exponent $\widehat \alpha_\ell$ as a product of top-down contributions by top-level exponent $\beta$ and per-level factors $\beta_\ell$:
\begin{equation}
    \begin{aligned}
        \varepsilon_\ell &= 2\varepsilon\left(1-2^{-(L-\ell)}\right),\\
        \beta_\ell &= \beta+\varepsilon_\ell,\qquad \varepsilon_L=0,\\
        \widehat{\alpha}_\ell &= \prod_{k=\ell}^{L}\beta_k,\qquad
        \widehat{\alpha}_L=\beta.
    \end{aligned}
    \label{eq:sparsity-exponent-compounding}
\end{equation}
This product includes the current level and every level above it: $\widehat{\alpha}_{L-1}=\beta_{L-1}\beta$ and $\widehat{\alpha}_{L-2}=\beta_{L-2}\beta_{L-1}\beta$. To account for sparsity introduced by higher-level pruning and KV-group unioning, $\varepsilon_\ell$ is a saturating additive correction to the factor $\beta_\ell$. We fit the overall shape parameters $(\beta,\varepsilon)$ separately for each geometry and $p$.\footnote{Specifically, we fit $(\beta,\varepsilon)$ for each geometry by equally weighted least squares against the empirical exponents $\alpha_\ell$, allowing signed $\varepsilon$ while keeping every $\beta_\ell$ positive, with an independent amplitude $c_\ell$ calibrated to pooled mean counts by $1/T$-weighted least squares in log space at each level. See Appendix~\ref{app:sparsity-scaling} for more details.} In Appendix~\ref{app:sparsity-scaling}, we compare to more general additive corrections, from a constant additive factor to increasing linear drift.

\Cref{fig:sparsity-scaling}(f) shows the empirically estimated per-level exponents $\alpha_\ell$ along with their fitted cumulative form $\widehat{\alpha}_\ell$ for all five depths; note that this form accurately captures the across-level exponent trends at $p=0.8$. Appendix~\ref{app:sparsity-scaling} completes the comparison across all four routing thresholds and gives the problem counts, full-resolution fitting, display binning, and bootstrap details.

\begin{figure}[!htbp]
    \centering
    \includegraphics[width=\linewidth]{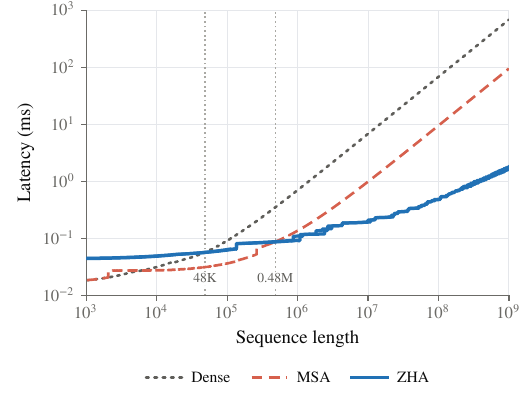}
    \caption{For a fixed top-$p$, and hence sparsity level, ZHA (here, R1, C8, $p=0.75$) scales subquadratically according to our decode roofline model, eventually dramatically outperforming both dense attention and the MSA sparse attention variant. Importantly, we have not empirically validated the scaling at these ultra-long context lengths due to the lack of existing LLMs which can operate at these contexts. Because of this, while we can obtain subquadratic latency, it is unclear to what extent reasoning and retrieval performance degrades at the level of depths required for these ultra-long contexts. Latencies are modeled per-token decode times for the attention subsystem of one layer at batch size 1.}
    \label{fig:speedup-latency}
\end{figure} 

Crucially, to predict sparsity scaling for longer sequences, we must also extrapolate the within-tree profiles above to increasingly deeper trees. Given the sparsity of available data (20 measurements, spanning five depths per $p$), we present upper and lower bounds rather than attempt a tight fit. In particular, we parameterize the top exponent as
\begin{equation}
    \widehat{\beta}(L,p) = b_0+b_p p + \left[d_0+d_p p\right] L.
    \label{eq:sparsity-top-exponent-depth}
\end{equation}
The constant-top scenario sets $d_0=d_p=0$; the linear-top scenario fits all four coefficients directly to the observed top-level exponents $\alpha_L$, independently of the within-tree profile fit. To isolate the effect of ultimate depth, we fit this model on the top-level exponents across the variable-depth geometries (the same 20 measurements referred to above); Appendix~\ref{app:sparsity-scaling} reports the exact coefficients and validation procedure. From this limited data, linear-top is better than constant-top, though neither provides a particularly good fit. However, they jointly provide a loose bound on the top-level exponent with increasing depth, which we use to extrapolate ZHA's latency out to very long contexts.

\subsection{Latency Comparison}

Given this roofline model and selection scaling framework, we compare the theoretical latency scaling of ZHA to dense attention and MSA for sequence lengths up to $10^9$ tokens. For this analysis, we restrict to a lightweight ZHA geometry (R1, C8, $p=0.75$; Qwen 35B with BF16) that retains high performance across our long-context reasoning and retrieval evaluation suite (see \cref{fig:evaluation-qwen-small-router} in Appendix~\ref{app:training-evaluation}; these results use L2), and compare its modeled latency to dense attention and MSA. To model the per-level exponent extrapolation with depth, we transfer the C4, R16 AIME exponent fits and calibrate count amplitudes to R1, C8, L2 AIME measurements. At each sequence length, we choose the ultimate tree depth that minimizes latency. As shown in \cref{fig:speedup-latency}, the model predicts that ZHA becomes faster than dense attention at approximately 48K tokens (at L1) and MSA at 0.48M tokens (at L2). At $10^9$ tokens, these speedups reach about $373$--$417\times$ and $52$--$58\times$, respectively, with selected depth L5. \Cref{fig:speedup-latency-thresholds} in Appendix~\ref{app:sparsity-scaling} compares $p=0.7,0.75,0.8$. 

\section{Analysis}
\label{sec:analysis}

Several design choices in the development of ZHA rely upon assumptions about the underlying attention distribution, and therefore how best to approximate it with subquadratic compute. Here, we empirically investigate two such phenomena: the dynamic sparsity introduced by top-$p$, and the learned distribution and usage dynamics of the router's reference queries.

\subsection{Variable Sparsity}

In contrast to fixed top-$K$ thresholds such as MSA, ZHA's top-$p$ threshold dynamically varies the number of attended keys. Not only does this facilitate proper scaling of retrieval with sequence length, but it also enables adaptive retrieval depending on the query and preceding context. The utility of the latter depends on the extent to which the underlying attention dynamics have differing sparsity depending on the setting and task, and whether the model can adaptively route through the tree according to these dynamics.

First, we compare attention sparsity across the evaluation domains, which require retrieval patterns ranging from narrow lookup to diffuse aggregation. At context lengths of approximately 64K tokens, 90\% of a query head’s dense attention mass is concentrated in a median of 2.3\% of 16-key chunks on RULER, 11.7\% on SWE-bench Verified, and 15.0\% on fiction books.\footnote{We exclude key 0 and the current partial chunk, which ZHA always attends to, and record dense attention along the router's own decode. Medians are taken over decode steps, the 10 full-attention layers, and query heads within each sequence, then across sequences.} A representative ZHA geometry (R16, C16, L3), also at $p=0.9$, retains 0.44\%, 5.0\%, and 7.4\% of chunks, respectively. These chunks contain a median of 80\%, 78\%, and 78\% of dense attention mass, with individual RULER tasks ranging from 74--82\%. Though the router is consistently sparser than the dense top-$p$ set (reflecting iterative pruning across the three levels), the pattern across tasks is consistent. Furthermore, the variation for both dense and ZHA matches our intuition: in RULER, its synthetic filler carries little that the answer depends on; in agentic contexts like SWE-bench, the relevant information is spread over code, tool outputs, and earlier reasoning; and in fiction books, the information is diffuse throughout the entire context.

RULER provides a particularly controlled setting in which to isolate how specific tasks induce different attention distributions such as either diffuse or concentrated. As such, we compare the size of ZHA's retrieved key set to the geometric mean over all 13 RULER subtasks. The tasks fall into three sparsity tiers: tasks with a single target in repetitive filler (\texttt{variable\_tracking}, \texttt{niah\_single\_1}) read about half as many keys as the typical task; needles in essay text and document QA sit near the middle (0.82--1.26$\times$); and tasks that must scan many candidates, either aggregating items (\texttt{cwe}, \texttt{fwe}) or distinguishing many look-alike needles (\texttt{niah\_multikey\_2/3}), read about 1.6$\times$ as many. To concretely illustrate how ZHA achieves variable sparsity, \Cref{fig:traversal-strip-combined} depicts a representative attention head's traversal through the tree for \texttt{niah\_single\_1} and \texttt{niah\_multikey\_2}; both traversals ultimately attend to the correct answer-containing chunk, but the former tree is far more aggressively pruned than the latter. Note that this adaptive sparsity is impossible for a fixed-budget top-$K$ mechanism, which instead forces a tradeoff between a large budget wasting keys on concentrated inputs, and a small one missing context needed for diffuse retrieval. ZHA's top-$p$ threshold avoids this choice entirely.

\subsection{Reference-Query Structure}

Another question concerns the nature of the learned structure in the reference queries, which comprise the only learned components in the router and therefore are critical for its efficacy. Recall that each reference query produces a compressed key and bias defining an affine approximation to the chunk’s log attention mass. Taking the maximum over these approximations yields a piecewise-affine approximation to the true function. In this setup, each reference query serves as a tangent plane's center point, and therefore it might reflect some central tendency in the query distribution. However, this maximum over the reference-query-mediated scores in the KL training objective does not explicitly enforce balanced usage or distinct reference query representations. In the worst case, they might simply collapse while still adding a constant-factor computational cost.

\begin{figure}[!t]
  \centering
  \includegraphics[width=\linewidth]{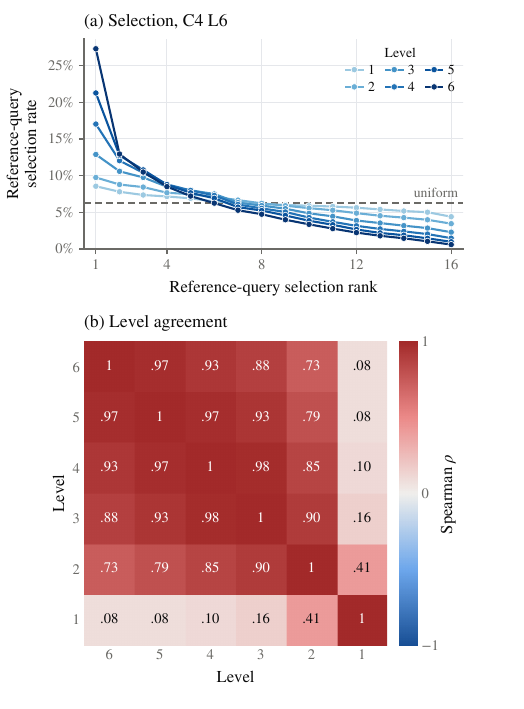}
  \caption{Reference-query selection rates per tree level for the C4~L6 ZHA router on natural text. (a) Within each KV group and level, reference queries are independently ranked by selection frequency. Each point shows the level's share of selections at that rank, averaged over layers and KV groups. Higher levels concentrate selections among fewer reference queries, while lower levels use them more uniformly. (b) Spearman correlation between levels' selection rates. Higher levels have strongly aligned reference-query rankings. Agreement generally weakens as levels become further separated, with the chunk level showing the greatest divergence from the upper levels.}
  \label{fig:qref-selection-rank}
\end{figure}

Thus, a reference query is only useful insofar as it is selected by the max-pooling operation. We therefore measure the selection rates of reference queries across levels of the tree, using a sample of length-32K natural language documents from our training mix.\footnote{133 documents of at least length-32K in one training-mix shard; these are chat-formatted, mostly long mathematical reasoning, plus general chat, agentic coding, and long-document QA. At every 8th query position, we count which reference query wins each kept node. As the routers may have seen this data, \cref{app:qref-level-agreement} repeats the analysis on RULER.} The reference queries develop a coherent pattern of utilization both within and between tree levels; this is summarized in \cref{fig:qref-selection-rank}. In particular, the selection rate among reference queries within a level follows a gradual decay, which becomes increasingly concentrated among a few reference queries at higher tree levels and more balanced toward the leaves. Furthermore, comparing the rank correlation in the selection rate orderings between levels, the higher levels consistently use the same reference queries most and least commonly, whereas the lowest level's usage pattern diverges, suggesting different selection dynamics depending on the level's compression factor. The across-level usage patterns are further broken down in Appendix~\ref{app:qref-level-agreement}, demonstrating rich selection-rate structure in which reference query usage varies smoothly across levels.

\begin{figure}
  \centering
  \includegraphics[width=0.80\linewidth]{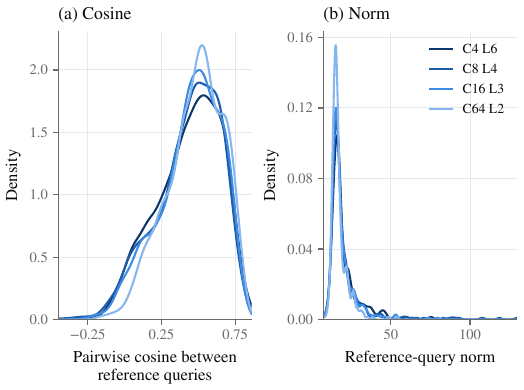}
  \caption{(a) Cosine similarity distribution of every pair of reference queries within a KV group; (b) norm distribution of every reference query.  Reference queries within a KV group become slightly more similar and their norms more concentrated as the tree gets shallower and its chunks larger.}
  \label{fig:qref-cos-norm}
\end{figure}

Moreover, we investigate the learned structure, using a deep geometry (R16, C4, L6) with moderate sparsity ($p=0.9$) as a case study. First, \cref{fig:qref-cos-norm} summarizes their geometric variation. The pairwise cosine similarities among the 16 reference queries within a KV group, averaging 0.44 across all 20 KV groups, show that they occupy a cone of directions rather than collapsing to identical vectors. This structure emerges across geometries; independently trained routers with varying chunk sizes and depths learn closely-matching reference queries (\cref{app:reference-similarity}). This similarity across geometries is clearest for the most frequently selected references. The reference query norms have some spread while remaining fairly concentrated. Interestingly, selection also favors smaller-norm reference queries. Since smaller norms lead to smoother weighting, it may indicate that more uniform averaging of keys becomes advantageous at higher levels. Together, the distribution in cosine similarity and norms suggests that the reference queries have organized into a coherent and non-collapsed geometric structure.

\section{Discussion} \label{sec:discussion}

In this paper, we have introduced ZHA, a tree-structured hierarchical sparse attention method that can achieve asymptotically subquadratic attention scaling, subject to specific constraints on the tree geometry and the top-$p$ value. Moreover, we have shown that ZHA can be applied to existing base models in a short distillation phase with only a billion tokens, and it maintains performance on challenging long-context reasoning, agentic coding, and retrieval tasks. 

While empirically ZHA appears to perform very well for contexts up to 256K, performance does degrade at increasingly higher levels of sparsity, and thus ultimately at longer context lengths. Here, we present perhaps the simplest possible version of the tree-structured indexer idea, which only learns reference queries in a separate independent KL-distillation phase. There are potentially many improvements that could be made to the core algorithm or training recipe to improve its performance. For instance, chunk boundaries could be dynamically learned instead of set arbitrarily; we could introduce a recency bias into the tree so that it becomes more detailed closer to the current token, rather than uniformly distributing mass over the sequence; and we could finetune the base model parameters on long-context training sets with a highly aggressive ZHA geometry to let the rest of the model learn to compensate for the degradation caused by the sparsity. We could also augment the tree with compressed value summaries, allowing nodes pruned during routing to contribute coarse attention outputs. This would preserve approximate visibility over the full KV cache span, combining token-level attention over selected regions with summary-level contributions from pruned subtrees.

\paragraph{Parameter scaling.}
Our method is surprisingly simple, requiring only a single KL-warmup phase in which we train a set of reference queries (these are the only learnable parameters we add) to match the attention mass induced by regular dense attention. Although we initially tried using ideas from \citet{hils} such as a reference-query LoRA and an explicit factorization of the attention probabilities across different levels of the tree, we found that these methods added complexity and did not improve performance in our preliminary ablations. 

One interesting trend we noted during our initial development of ZHA is that the performance of ZHA improves with model scale. We found that small models under 1B parameters often struggled significantly with RULER and other retrieval tasks during conversion to ZHA because the original attention scores we were distilling from were very noisy themselves. Our success with distillation improved dramatically when moving to larger models in the 4B scale and up. This makes us optimistic that ZHA will continue to improve with further increases to model scale.

\paragraph{Positional encoding.}
Another interesting question concerns the impact of different positional encoding schemes on the effectiveness of ZHA. The primary experiments we ran were on the Qwen model series which uses partial RoPE, as were our experiments on the structure of naturalistic data in \cref{fig:dense-top-p-growth}. However, we find that RoPE induces a very particular structure into the attention scores over long contexts, where most heads represent a strong recency bias while a few heads appear to primarily attend to the lowest RoPE frequencies and are sensitive much more uniformly across the context. NoPE attention, by contrast, can induce a more diffuse attention structure, and thus may extrapolate or give different asymptotic scalings for ZHA than RoPE models. In Appendix~\ref{app:evaluation-nemotron}, we also tested the hybrid-NoPE Nemotron-3.5 models and found that ZHA retains reasoning and coding performance close to dense, but exhibits larger retrieval losses. Therefore, the long-context extrapolation capabilities of ZHA with NoPE remain uncertain. 

\paragraph{Asymptotic length scaling.}
The most important limitation in our results is that, although we have extrapolated latency out to millions of tokens of context and beyond, we have not empirically validated the evaluation performance of ZHA at these context lengths. The primary limitation is finding models with sufficiently long contexts that are small enough to adapt without substantial computational expense. In future work, we plan to scale up the method to frontier open-source models and determine whether our method continues to scale to 1 million tokens of context and beyond. 

Of course, this problem of evaluating performance at contexts much beyond 1 million tokens is not unique to ZHA, and is in fact shared by all attention-based models. Though this in part derives from data limitations, it also points back to the inspiration for sparse attention methods: attention's quadratic scaling makes long-context training and inference prohibitively expensive. Beyond this computational bottleneck, however, it is still unclear whether dense attention itself is even performant at longer sequence lengths. Prior work has in fact found that softmax can be ``dispersive'', in the sense that it upweights noise within the preceding context, leading to increasingly uniform attention scores that eventually prevent effective retrieval \citep{SoftmaxNotEnough}.

Therefore, an argument may be made that sparsity is not only necessary for computational efficiency, but also to continue to scale language modeling to ultra-long contexts. It may be the case that sparsity's pruning effectively ``denoises'' the past context, sidestepping the dispersiveness of dense attention and uncovering the truly relevant history to be retrieved. Indeed, our finding that top-$p$ grows like a sublinear power law is highly suggestive that the subset of information that is truly relevant for predicting the next token has fundamentally different scaling properties than dense retrieval of the entire context.

A core assumption of our work is that this top-$p$ set contains sufficient information to enable effective language modeling. ZHA's empirical success in part validates this assumption; however, it is admittedly unclear whether this will remain the case indefinitely. Similarly, it is unclear whether top-$p$, and the underlying predictive information that we suppose it picks up, grows sublinearly to indefinite context lengths.

Supposing that retrieving the true top-$p$ set is ``all you need'', the remaining source of error is properly routing to that set. Due to asymptotic divergence of different power-law curves, precisely matching the true scaling of the top-$p$ set is a surprisingly subtle point. In short, it turns out that our method of applying top-$p$ iteratively down a growing tree requires carefully tuning the per-level $p$, and failing to do so results in two possibilities. First, fixing $p<1$ as we do currently, the ``cumulative'' $p_{\text{eff}} = p^L \to 0$ as depth $L \to \infty$. In this case, the retrieval state asymptotes towards some constant size and performance necessarily degrades. Alternatively, if we rescaled per-level $p$ to $p^{1/L}$ to counteract this effect, we could then observe the opposite pathology where $p$ tends to 1 as depth increases, leading to all of the nodes at the higher levels being selected and the upper levels of the tree effectively becoming redundant. If this occurs then we return to quadratic scaling and hence the diffusive pathology of fully dense attention.

To remain subquadratic in the limit while also maintaining performance, then, we need some method to scale $p$ dynamically with depth such that the effective $p$ at the leaf nodes decreases, but at a slow enough rate that it tracks the decrease in predictive information in the sequence with increasing context length, similar to that we found in \cref{fig:dense-top-p-growth}. This was roughly a power-law scaling. However, this depends not on the architecture, but on the intrinsic distribution of the data and thus is deeply dependent upon the statistical structure of ultra-long context data. Our analysis suggests that naturalistic data does exhibit an amenable scaling, but we only investigated data up to 256K context, and thus extrapolating this to hundreds of millions or billions of tokens could fail. Whether or not naturalistic text data continues to exhibit power-law self-similar structure at these ultra-long scales thus remains an important avenue for future work.

\FloatBarrier
\bibliographystyle{style/tmlr}
\bibliography{references}

\begin{thebibliography}{49}
\providecommand{\natexlab}[1]{#1}
\providecommand{\url}[1]{\texttt{#1}}
\expandafter\ifx\csname urlstyle\endcsname\relax
  \providecommand{\doi}[1]{doi: #1}\else
  \providecommand{\doi}{doi: \begingroup \urlstyle{rm}\Url}\fi

\bibitem[Ainslie et~al.(2023)Ainslie, Lee-Thorp, De~Jong, Zemlyanskiy,
  Lebr{\'o}n, and Sanghai]{GQA}
Joshua Ainslie, James Lee-Thorp, Michiel De~Jong, Yury Zemlyanskiy, Federico
  Lebr{\'o}n, and Sumit Sanghai.
\newblock {GQA}: Training generalized multi-query transformer models from
  multi-head checkpoints.
\newblock In \emph{Proceedings of the 2023 Conference on Empirical Methods in
  Natural Language Processing}, pp.\  4895--4901, 2023.

\bibitem[Bai et~al.(2024)Bai, Tu, Zhang, Peng, Wang, Lv, Cao, Xu, Hou, Dong,
  Tang, and Li]{longbenchv2}
Yushi Bai, Shangqing Tu, Jiajie Zhang, Hao Peng, Xiaozhi Wang, Xin Lv, Shulin
  Cao, Jiazheng Xu, Lei Hou, Yuxiao Dong, et~al.
\newblock {LongBench v2}: Towards deeper understanding and reasoning on
  realistic long-context multitasks.
\newblock \emph{arXiv preprint arXiv:2412.15204}, 2024.

\bibitem[Bai et~al.(2026)Bai, Dong, Jiang, Lv, Du, Zeng, Tang, and
  Li]{IndexCache}
Yushi Bai, Qian Dong, Ting Jiang, Xin Lv, Zhengxiao Du, Aohan Zeng, Jie Tang,
  and Juanzi Li.
\newblock {IndexCache}: Accelerating sparse attention via cross-layer index
  reuse.
\newblock \emph{arXiv preprint arXiv:2603.12201}, 2026.

\bibitem[Cai et~al.(2024)Cai, Zhang, Gao, Liu, Li, Liu, Lu, Xiong, Dong, Hu,
  et~al.]{PyramidKV}
Zefan Cai, Yichi Zhang, Bofei Gao, Yuliang Liu, Yucheng Li, Tianyu Liu, Keming
  Lu, Wayne Xiong, Yue Dong, Junjie Hu, et~al.
\newblock {PyramidKV}: Dynamic {KV} cache compression based on pyramidal
  information funneling.
\newblock \emph{arXiv preprint arXiv:2406.02069}, 2024.

\bibitem[Chowdhury et~al.(2024)Chowdhury, Aung, Shern, Jaffe, Sherburn,
  Starace, Mays, Dias, Aljubeh, Glaese, Jimenez, Yang, Ho, Patwardhan, Liu, and
  Madry]{chowdhury2024swebenchverified}
Neil Chowdhury, James Aung, Chan~Jun Shern, Oliver Jaffe, Dane Sherburn, Giulio
  Starace, Evan Mays, Rachel Dias, Marwan Aljubeh, Mia Glaese, et~al.
\newblock Introducing {SWE-bench Verified}, 2024.
\newblock URL \url{https://openai.com/index/introducing-swe-bench-verified/}.

\bibitem[Dao \& Gu(2024)Dao and Gu]{dao2024transformersssmsgeneralizedmodels}
Tri Dao and Albert Gu.
\newblock Transformers are {SSMs}: Generalized models and efficient algorithms
  through structured state space duality, 2024.
\newblock URL \url{https://arxiv.org/abs/2405.21060}.

\bibitem[Dao et~al.(2022)Dao, Fu, Ermon, Rudra, and R{\'e}]{flashattention}
Tri Dao, Daniel~Y. Fu, Stefano Ermon, Atri Rudra, and Christopher R{\'e}.
\newblock {FlashAttention}: Fast and memory-efficient exact attention with
  {IO}-awareness.
\newblock \emph{Advances in Neural Information Processing Systems},
  35:\penalty0 16344--16359, 2022.

\bibitem[{DeepSeek-AI}(2026)]{deepseekai2026deepseekv4highlyefficientmilliontoken}
{DeepSeek-AI}.
\newblock {DeepSeek-V4}: Towards highly efficient million-token context
  intelligence, 2026.
\newblock URL \url{https://arxiv.org/abs/2606.19348}.

\bibitem[{DeepSeek-AI} et~al.(2025)]{deepseekr1}
{DeepSeek-AI} et~al.
\newblock {DeepSeek-R1}: Incentivizing reasoning capability in {LLMs} via
  reinforcement learning.
\newblock \emph{arXiv preprint arXiv:2501.12948}, 2025.

\bibitem[Feng et~al.(2026)Feng, Lv, Cao, Xie, and Zhou]{AdaKV}
Yuan Feng, Junlin Lv, Yukun Cao, Xike Xie, and S~Kevin Zhou.
\newblock {Ada-KV}: Optimizing {KV} cache eviction by adaptive budget
  allocation for efficient {LLM} inference.
\newblock \emph{Advances in Neural Information Processing Systems},
  38:\penalty0 113152--113188, 2026.

\bibitem[Figliolia et~al.(2025)Figliolia, Alonso, Iyer, Anthony, and
  Millidge]{figliolia2025compressed}
Tomas Figliolia, Nicholas Alonso, Rishi Iyer, Quentin Anthony, and Beren
  Millidge.
\newblock Compressed convolutional attention: Efficient attention in a
  compressed latent space.
\newblock \emph{arXiv preprint arXiv:2510.04476}, 2025.

\bibitem[Fu et~al.(2025)Fu, Song, Wang, Wu, Zheng, Zhang, Xu, Wei, Xu, and
  Zhao]{fu2025slidingwindowattentiontraining}
Zichuan Fu, Wentao Song, Yejing Wang, Xian Wu, Yefeng Zheng, Yingying Zhang,
  Derong Xu, Xuetao Wei, Tong Xu, and Xiangyu Zhao.
\newblock Sliding window attention training for efficient large language
  models, 2025.
\newblock URL \url{https://arxiv.org/abs/2502.18845}.

\bibitem[Glorioso et~al.(2024{\natexlab{a}})Glorioso, Anthony, Tokpanov,
  Golubeva, Shyam, Whittington, Pilault, and Millidge]{glorioso2024zamba2}
Paolo Glorioso, Quentin Anthony, Yury Tokpanov, Anna Golubeva, Vasudev Shyam,
  James Whittington, Jonathan Pilault, and Beren Millidge.
\newblock The {Zamba2} suite: Technical report.
\newblock \emph{arXiv preprint arXiv:2411.15242}, 2024{\natexlab{a}}.

\bibitem[Glorioso et~al.(2024{\natexlab{b}})Glorioso, Anthony, Tokpanov,
  Whittington, Pilault, Ibrahim, and Millidge]{glorioso2024zamba}
Paolo Glorioso, Quentin Anthony, Yury Tokpanov, James Whittington, Jonathan
  Pilault, Adam Ibrahim, and Beren Millidge.
\newblock Zamba: A compact {7B} {SSM} hybrid model.
\newblock \emph{arXiv preprint arXiv:2405.16712}, 2024{\natexlab{b}}.

\bibitem[Gu \& Dao(2024)Gu and Dao]{gu2024mambalineartimesequencemodeling}
Albert Gu and Tri Dao.
\newblock Mamba: Linear-time sequence modeling with selective state spaces,
  2024.
\newblock URL \url{https://arxiv.org/abs/2312.00752}.

\bibitem[Hsieh et~al.(2024)Hsieh, Sun, Kriman, Acharya, Rekesh, Jia, Zhang, and
  Ginsburg]{ruler}
Cheng-Ping Hsieh, Simeng Sun, Samuel Kriman, Shantanu Acharya, Dima Rekesh, Fei
  Jia, Yang Zhang, and Boris Ginsburg.
\newblock {RULER}: What's the real context size of your long-context language
  models?
\newblock \emph{arXiv preprint arXiv:2404.06654}, 2024.

\bibitem[Hu et~al.(2024)Hu, Tu, Han, He, Cui, Long, Zheng, Fang, Huang, Zhao,
  Zhang, Thai, Zhang, Wang, Yao, Zhao, Zhou, Cai, Zhai, Ding, Jia, Zeng, Li,
  Liu, and Sun]{hu2024minicpm}
Shengding Hu, Yuge Tu, Xu~Han, Chaoqun He, Ganqu Cui, Xiang Long, Zhi Zheng,
  Yewei Fang, Yuxiang Huang, Weilin Zhao, et~al.
\newblock {MiniCPM}: Unveiling the potential of small language models with
  scalable training strategies.
\newblock \emph{arXiv preprint arXiv:2404.06395}, 2024.
\newblock URL \url{https://arxiv.org/abs/2404.06395}.

\bibitem[Hu et~al.(2026)Hu, Wei, Gu, Zhang, Liang, Li, Zhu, Wang, Han, Bai,
  et~al.]{hils}
Xiang Hu, Xinyu Wei, Hao Gu, Minshen Zhang, Tian Liang, Huayang Li, Lei Zhu,
  Yan Wang, Sirui Han, Yushi Bai, et~al.
\newblock Hierarchical sparse attention done right: Toward infinite context
  modeling.
\newblock \emph{arXiv preprint arXiv:2607.02980}, 2026.

\bibitem[Huang et~al.(2026)Huang, Gon{\c{c}}alves, Alvetreti, Li, Han, Ponti,
  Martins, and Treviso]{dash}
Yuxiang Huang, Nuno~MT Gon{\c{c}}alves, Federico Alvetreti, Lei Li, Xu~Han,
  Edoardo~M Ponti, Andr{\'e}~FT Martins, and Marcos~V Treviso.
\newblock {DashAttention}: Differentiable and adaptive sparse hierarchical
  attention.
\newblock \emph{arXiv preprint arXiv:2605.18753}, 2026.

\bibitem[Hwang et~al.(2025)Hwang, Wang, and Gu]{HNet}
Sukjun Hwang, Brandon Wang, and Albert Gu.
\newblock Dynamic chunking for end-to-end hierarchical sequence modeling.
\newblock \emph{arXiv preprint arXiv:2507.07955}, 2025.

\bibitem[Jelassi et~al.(2024)Jelassi, Brandfonbrener, Kakade, and
  Malach]{jelassi2024repeat}
Samy Jelassi, David Brandfonbrener, Sham~M. Kakade, and Eran Malach.
\newblock Repeat after me: Transformers are better than state space models at
  copying.
\newblock In \emph{Proceedings of the 41st International Conference on Machine
  Learning}, volume 235, pp.\  21502--21521, 2024.
\newblock URL \url{https://proceedings.mlr.press/v235/jelassi24a.html}.

\bibitem[Jimenez et~al.(2024)Jimenez, Yang, Wettig, Yao, Pei, Press, and
  Narasimhan]{jimenez2024swebench}
Carlos~E. Jimenez, John Yang, Alexander Wettig, Shunyu Yao, Kexin Pei, Ofir
  Press, and Karthik~R. Narasimhan.
\newblock {SWE}-bench: Can language models resolve real-world {GitHub} issues?
\newblock In \emph{The Twelfth International Conference on Learning
  Representations}, 2024.
\newblock URL \url{https://openreview.net/forum?id=VTF8yNQM66}.

\bibitem[Kim et~al.(2026)Kim, Han, and Yun]{kim2026fastkvzipefficientaccurate}
Jang-Hyun Kim, Dongyoon Han, and Sangdoo Yun.
\newblock Fast {KVzip}: Efficient and accurate {LLM} inference with gated {KV}
  eviction, 2026.
\newblock URL \url{https://arxiv.org/abs/2601.17668}.

\bibitem[Lahoti et~al.(2026)Lahoti, Li, Chen, Wang, Bick, Kolter, Dao, and
  Gu]{lahoti2026mamba3improvedsequencemodeling}
Aakash Lahoti, Kevin~Y. Li, Berlin Chen, Caitlin Wang, Aviv Bick, J.~Zico
  Kolter, Tri Dao, and Albert Gu.
\newblock Mamba-3: Improved sequence modeling using state space principles,
  2026.
\newblock URL \url{https://arxiv.org/abs/2603.15569}.

\bibitem[Lai et~al.(2026)Lai, Xu, Yang, Chen, Xu, Zeng, Li, Sun, Zhu, Zhang,
  et~al.]{MSA}
Xunhao Lai, Weiqi Xu, Yufeng Yang, Qiaorui Chen, Yang Xu, Lunbin Zeng, Xiaolong
  Li, Haohai Sun, Haichao Zhu, Vito Zhang, et~al.
\newblock {MiniMax} sparse attention.
\newblock \emph{arXiv preprint arXiv:2606.13392}, 2026.

\bibitem[Lee et~al.(2025)Lee, Park, Lee, Suh, Kim, Jeong, Kim, Lee, Jeon, and
  Hwang]{lee2025hip}
Heejun Lee, Geon Park, Youngwan Lee, Jaduk Suh, Jina Kim, Wonyong Jeong, Bumsik
  Kim, Hyemin Lee, Myeongjae Jeon, and Sung~Ju Hwang.
\newblock A training-free sub-quadratic cost transformer model serving
  framework with hierarchically pruned attention.
\newblock In \emph{International Conference on Learning Representations}, 2025.
\newblock URL \url{https://arxiv.org/abs/2406.09827}.

\bibitem[Lieber et~al.(2024)Lieber, Lenz, Bata, Cohen, Osin, Dalmedigos,
  Safahi, Meirom, Belinkov, Shalev-Shwartz, et~al.]{lieber2024jamba}
Opher Lieber, Barak Lenz, Hofit Bata, Gal Cohen, Jhonathan Osin, Itay
  Dalmedigos, Erez Safahi, Shaked Meirom, Yonatan Belinkov, Shai
  Shalev-Shwartz, et~al.
\newblock Jamba: A hybrid transformer-{Mamba} language model.
\newblock \emph{arXiv preprint arXiv:2403.19887}, 2024.

\bibitem[Lin et~al.(2025)Lin, Tang, Yang, Wang, Tang, Tian, Stoica, Han, and
  Gao]{lin2025twilight}
Chaofan Lin, Jiaming Tang, Shuo Yang, Hanshuo Wang, Tian Tang, Boyu Tian, Ion
  Stoica, Song Han, and Mingyu Gao.
\newblock {Twilight}: Adaptive attention sparsity with hierarchical top-$p$
  pruning.
\newblock In \emph{Advances in Neural Information Processing Systems},
  volume~38, 2025.
\newblock URL \url{https://arxiv.org/abs/2502.02770}.

\bibitem[Liu et~al.(2024)Liu, Feng, Wang, Wang, Liu, Zhao, Dengr, Ruan, Dai,
  Guo, et~al.]{liu2024deepseek}
Aixin Liu, Bei Feng, Bin Wang, Bingxuan Wang, Bo~Liu, Chenggang Zhao, Chengqi
  Dengr, Chong Ruan, Damai Dai, Daya Guo, et~al.
\newblock {DeepSeek-V2}: A strong, economical, and efficient mixture-of-experts
  language model.
\newblock \emph{arXiv preprint arXiv:2405.04434}, 2024.

\bibitem[Liu et~al.(2025)Liu, Mei, Lin, Xue, Wang, Xu, Wu, Zhang, Lin, Dong,
  et~al.]{DSV3.2}
Aixin Liu, Aoxue Mei, Bangcai Lin, Bing Xue, Bingxuan Wang, Bingzheng Xu,
  Bochao Wu, Bowei Zhang, Chaofan Lin, Chen Dong, et~al.
\newblock {DeepSeek-V3.2}: Pushing the frontier of open large language models.
\newblock \emph{arXiv preprint arXiv:2512.02556}, 2025.

\bibitem[Liu et~al.(2023)Liu, Desai, Liao, Wang, Xie, Xu, Kyrillidis, and
  Shrivastava]{Scissorhands}
Zichang Liu, Aditya Desai, Fangshuo Liao, Weitao Wang, Victor Xie, Zhaozhuo Xu,
  Anastasios Kyrillidis, and Anshumali Shrivastava.
\newblock Scissorhands: Exploiting the persistence of importance hypothesis for
  {LLM} {KV} cache compression at test time.
\newblock \emph{Advances in Neural Information Processing Systems},
  36:\penalty0 52342--52364, 2023.

\bibitem[Loshchilov \& Hutter(2019)Loshchilov and
  Hutter]{loshchilov2019decoupled}
Ilya Loshchilov and Frank Hutter.
\newblock Decoupled weight decay regularization.
\newblock In \emph{International Conference on Learning Representations}, 2019.
\newblock URL \url{https://arxiv.org/abs/1711.05101}.

\bibitem[{MathArena}(2026)]{matharena2026}
{MathArena}.
\newblock {AIME 2026} dataset, 2026.
\newblock URL
  \url{https://huggingface.co/datasets/MathArena/aime_2026/tree/d2de22f3c656b4f56cf8981212186377d1e23bc3}.

\bibitem[{NVIDIA}(2026)]{nemotron35}
{NVIDIA}.
\newblock {NVIDIA Nemotron 3.5 Super and Lightning}, 2026.
\newblock URL
  \url{https://research.nvidia.com/labs/nemotron/Nemotron-3.5-Super-and-Lightning/}.

\bibitem[Qiu et~al.(2026)Qiu, Wang, Zheng, Huang, Wen, Yang, Men, Yu, Huang,
  Huang, et~al.]{qiu2026gated}
Zihan Qiu, Zekun Wang, Bo~Zheng, Zeyu Huang, Kaiyue Wen, Songlin Yang, Rui Men,
  Le~Yu, Fei Huang, Suozhi Huang, et~al.
\newblock Gated attention for large language models: Non-linearity, sparsity,
  and attention-sink-free.
\newblock \emph{Advances in Neural Information Processing Systems},
  38:\penalty0 100092--100118, 2026.

\bibitem[{Qwen Team}(2026)]{qwen36model}
{Qwen Team}.
\newblock {Qwen3.6-35B-A3B} model card, 2026.
\newblock URL \url{https://huggingface.co/Qwen/Qwen3.6-35B-A3B}.

\bibitem[Rein et~al.(2023)Rein, Hou, Stickland, Petty, Pang, Dirani, Michael,
  and Bowman]{gpqa}
David Rein, Betty~Li Hou, Asa~Cooper Stickland, Jackson Petty, Richard~Yuanzhe
  Pang, Julien Dirani, Julian Michael, and Samuel~R. Bowman.
\newblock {GPQA}: A graduate-level {Google}-proof {Q\&A} benchmark.
\newblock \emph{arXiv preprint arXiv:2311.12022}, 2023.

\bibitem[Su et~al.(2024)Su, Ahmed, Lu, Pan, Bo, and Liu]{su2024roformer}
Jianlin Su, Murtadha Ahmed, Yu~Lu, Shengfeng Pan, Wen Bo, and Yunfeng Liu.
\newblock {RoFormer}: Enhanced transformer with rotary position embedding.
\newblock \emph{Neurocomputing}, 568:\penalty0 127063, 2024.

\bibitem[Tang et~al.(2026)Tang, Qin, Pan, Li, and Liu]{pisa}
Bohao Tang, Zhen Qin, Yuqi Pan, Zheng Li, and Pengfei Liu.
\newblock Block sparse attention with log-linear complexity.
\newblock \emph{arXiv preprint arXiv:2609.31093}, 2026.

\bibitem[Team et~al.(2025)Team, Zhang, Lin, Yao, Hu, Meng, Liu, Men, Yang, Li,
  et~al.]{team2025kimi}
Kimi Team, Yu~Zhang, Zongyu Lin, Xingcheng Yao, Jiaxi Hu, Fanqing Meng,
  Chengyin Liu, Xin Men, Songlin Yang, Zhiyuan Li, et~al.
\newblock {Kimi Linear}: An expressive, efficient attention architecture.
\newblock \emph{arXiv preprint arXiv:2510.26692}, 2025.

\bibitem[{Terminal-Bench Team}(2026)]{terminalbench21}
{Terminal-Bench Team}.
\newblock {Terminal-Bench 2.1}, 2026.
\newblock URL \url{https://www.tbench.ai/news/terminal-bench-2-1}.

\bibitem[Vaswani et~al.(2017)Vaswani, Shazeer, Parmar, Uszkoreit, Jones, Gomez,
  Kaiser, and Polosukhin]{vaswani2017attention}
Ashish Vaswani, Noam Shazeer, Niki Parmar, Jakob Uszkoreit, Llion Jones,
  Aidan~N. Gomez, {\L}ukasz Kaiser, and Illia Polosukhin.
\newblock Attention is all you need.
\newblock In \emph{Advances in Neural Information Processing Systems},
  volume~30, 2017.
\newblock URL \url{https://arxiv.org/abs/1706.03762}.

\bibitem[Veli{\v{c}}kovi{\'c} et~al.(2024)Veli{\v{c}}kovi{\'c},
  Perivolaropoulos, Barbero, and Pascanu]{SoftmaxNotEnough}
Petar Veli{\v{c}}kovi{\'c}, Christos Perivolaropoulos, Federico Barbero, and
  Razvan Pascanu.
\newblock Softmax is not enough (for sharp size generalisation).
\newblock \emph{arXiv preprint arXiv:2410.01104}, 2024.

\bibitem[Wang et~al.(2024)Wang, Ma, Zhang, Ni, Chandra, Guo, Ren, Arulraj, He,
  Jiang, et~al.]{mmlupro}
Yubo Wang, Xueguang Ma, Ge~Zhang, Yuansheng Ni, Abhranil Chandra, Shiguang Guo,
  Weiming Ren, Aaran Arulraj, Xuan He, Ziyan Jiang, et~al.
\newblock {MMLU-Pro}: A more robust and challenging multi-task language
  understanding benchmark.
\newblock \emph{arXiv preprint arXiv:2406.01574}, 2024.

\bibitem[Washbourne et~al.(2026)Washbourne, Iyer, Figliolia, Zheng,
  Lorig-Roach, Yang, Yuvraj, Anthony, Tokpanov, Yang, et~al.]{zaya1}
Robert Washbourne, Rishi Iyer, Tomas Figliolia, Henry Zheng, Ryan Lorig-Roach,
  Sungyeon Yang, Pritish Yuvraj, Quentin Anthony, Yury Tokpanov, Xiao Yang,
  et~al.
\newblock {ZAYA1-8B} technical report.
\newblock \emph{arXiv preprint arXiv:2605.05365}, 2026.

\bibitem[Williams et~al.(2009)Williams, Waterman, and
  Patterson]{williams2009roofline}
Samuel Williams, Andrew Waterman, and David Patterson.
\newblock Roofline: An insightful visual performance model for multicore
  architectures.
\newblock \emph{Communications of the ACM}, 52\penalty0 (4):\penalty0 65--76,
  2009.
\newblock \doi{10.1145/1498765.1498785}.

\bibitem[Yuan et~al.(2025)Yuan, Gao, Dai, Luo, Zhao, Zhang, Xie, Wei, Wang,
  Xiao, et~al.]{NSA}
Jingyang Yuan, Huazuo Gao, Damai Dai, Junyu Luo, Liang Zhao, Zhengyan Zhang,
  Zhenda Xie, Yuxing Wei, Lean Wang, Zhiping Xiao, et~al.
\newblock Native sparse attention: Hardware-aligned and natively trainable
  sparse attention.
\newblock In \emph{Proceedings of the 63rd Annual Meeting of the Association
  for Computational Linguistics (Volume 1: Long Papers)}, pp.\  23078--23097,
  2025.

\bibitem[Zan et~al.(2026)Zan, Zhang, Tan, Liu, Wang, Li, Ma, Wu, Lu, Li,
  et~al.]{LongCat}
Wen Zan, Jiaqi Zhang, Jianchao Tan, Hong Liu, Cunguang Wang, Xiang Li, Duyue
  Ma, Guanyu Wu, Yifan Lu, Fengcun Li, et~al.
\newblock {LongCat} sparse attention: Taming the lightning via streaming-aware
  hierarchical cross-layer indexing.
\newblock \emph{arXiv preprint arXiv:2608.01662}, 2026.

\bibitem[Zhang et~al.(2023)Zhang, Sheng, Zhou, Chen, Zheng, Cai, Song, Tian,
  R{\'e}, Barrett, et~al.]{H2O}
Zhenyu Zhang, Ying Sheng, Tianyi Zhou, Tianlong Chen, Lianmin Zheng, Ruisi Cai,
  Zhao Song, Yuandong Tian, Christopher R{\'e}, Clark Barrett, et~al.
\newblock {H2O}: Heavy-hitter oracle for efficient generative inference of
  large language models.
\newblock \emph{Advances in Neural Information Processing Systems},
  36:\penalty0 34661--34710, 2023.

\end{thebibliography}

\clearpage
\onecolumn
\appendices
\startcontents[appendix]
{\begin{center}
    \Large Appendix
\end{center}}
{\hypersetup{linkcolor=black}\printcontents[appendix]{}{1}{\setcounter{tocdepth}{2}}}
\clearpage

\section{Top-\texorpdfstring{$p$}{p} Scaling by Layer and Head}
\label[appendix]{app:top-p-scaling}

\subsection{Dense Top-\texorpdfstring{$p$}{p} Sets by Layer and Head}
\label[appendix]{app:top-p-dense-heads}

\begin{figure}[H]
    \centering
    \includegraphics[width=\linewidth]{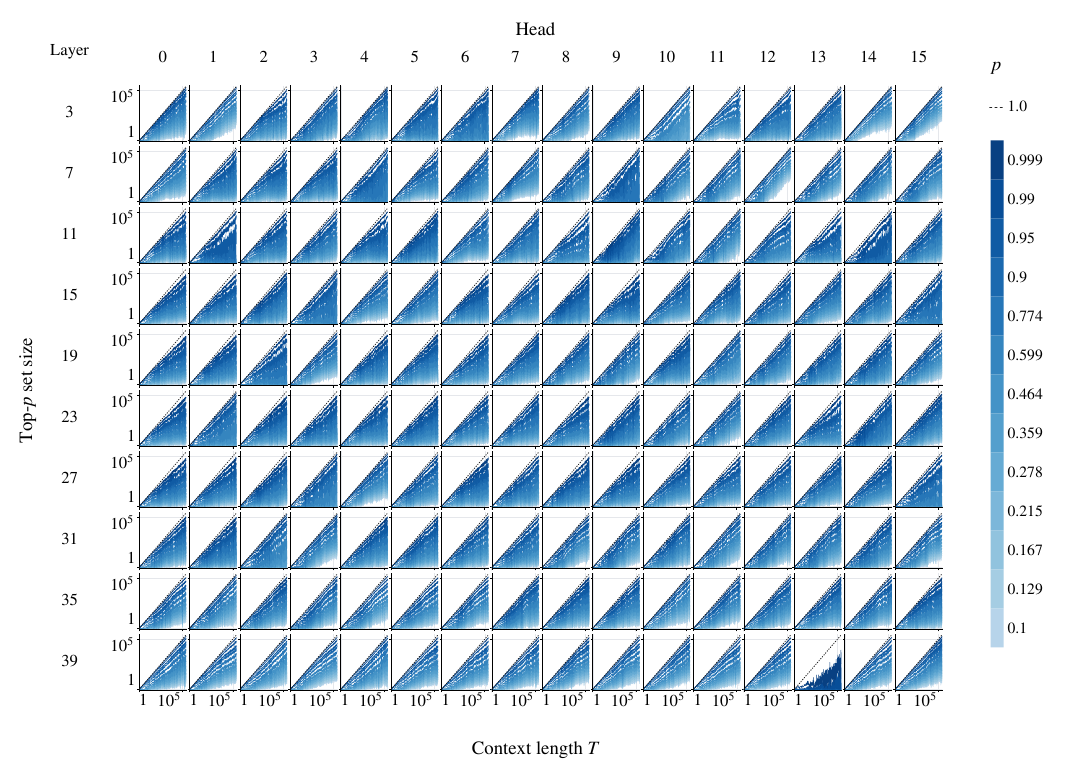}
\caption{Dense top-$p$ set size for individual query heads in Qwen3.6-35B-A3B, using the same six-book cohort and thresholds as \cref{fig:dense-top-p-growth}.
Rows identify full-attention layers and columns identify query heads, with zero-based indices.
Each panel averages only across books.
All panels share logarithmic axis limits; colors denote $p$ and the dotted black line gives the $p=1$ reference.}
    \label{fig:dense-top-p-heads}
\end{figure}

In \cref{sec:motivation}, we average dense top-$p$ set sizes over layers and query heads.
\Cref{fig:dense-top-p-heads} shows all 10 full-attention layers and 16 query heads of Qwen3.6-35B-A3B separately, averaging only across the six books.
Common logarithmic axes allow comparison with the linear $p=1$ reference, which retains all keys.
Sublinear trends appear throughout the full-attention stack, with substantial variation in set size and growth rate.
The aggregate trend is thus not confined to a few sparse heads, but does not imply a common exponent.

\subsection{Affine Versus Power-Law Fits on Rollouts}
\label[appendix]{app:top-p-rollout-fits}

\Cref{tab:rollout-linear-preference} provides a complementary comparison using ZHA leaf-level selection counts, not dense top-$p$ set sizes.
Each count unions selected leaf nodes across the eight query heads sharing a KV head and excludes mandatory sink and local tokens.
Here, $C$ is the checkpoint's base KV chunk size, $Q$ is the query-tile width, and $p$ is the routing mass threshold.

We fit each rollout--layer--KV-head trace separately, averaging tile counts in 256-query bins.
Both models minimize squared log-count residuals on the same observations.
Raw and cumulative-maximum targets are fitted separately, with running maxima computed within each rollout before bin averaging.
The Raw and Cum-max columns each give the affine win rate among 9,600 paired fits.

The preferred model depends on the configuration.
At $p=0.9$ and $(C,Q)=(8,16)$, the affine win rate is $3.76\%$ for raw counts and $5.36\%$ for cumulative-maximum counts.
For $(C,Q)=(256,256)$ at the same threshold, the corresponding rates are $99.97\%$ and $100\%$.
Power laws are therefore preferred in many configurations, but not uniformly across checkpoint geometries and query-tile widths.

\begin{table}[H]
\centering
\footnotesize
\setlength{\tabcolsep}{3pt}
\renewcommand{\arraystretch}{1.05}
\captionsetup{font=small}
\caption{Percentage of individual rollout--layer--head fits for which the affine model beats the power law, $\rho=(1-R^2_{\mathrm{lin}})/(1-R^2_{\mathrm{pow}})<1$, evaluated in log-$y$ space at the leaf level.
Raw and Cum-max denote raw and cumulative-maximum targets.
Each percentage uses 9,600 paired fits (30 problems $\times$ 16 rollouts $\times$ 10 layers $\times$ 2 KV heads), without averaging across rollouts before fitting.}
\label{tab:rollout-linear-preference}
\begin{subtable}[t]{0.32\textwidth}
\centering
\caption{$Q=1$}
\label{tab:rollout-linear-preference-q1}
\begin{tabular}{rrrr}
\toprule
$C$ & $p$ & Raw (\%) & Cum-max (\%) \\
\midrule
4 & 0.70 & $56.43$ & $39.18$ \\
4 & 0.80 & $11.95$ & $8.59$ \\
4 & 0.90 & $11.27$ & $6.77$ \\
4 & 0.95 & $11.94$ & $7.99$ \\
\addlinespace[2pt]
8 & 0.70 & $12.50$ & $8.67$ \\
8 & 0.80 & $9.11$ & $4.61$ \\
8 & 0.90 & $10.38$ & $6.82$ \\
8 & 0.95 & $10.40$ & $7.35$ \\
\addlinespace[2pt]
16 & 0.70 & $5.86$ & $2.94$ \\
16 & 0.80 & $7.34$ & $5.00$ \\
16 & 0.90 & $11.52$ & $7.28$ \\
16 & 0.95 & $9.57$ & $8.13$ \\
\addlinespace[2pt]
64 & 0.70 & $1.64$ & $0.72$ \\
64 & 0.80 & $4.41$ & $1.81$ \\
64 & 0.90 & $7.49$ & $4.07$ \\
64 & 0.95 & $8.09$ & $15.56$ \\
\addlinespace[2pt]
256 & 0.80 & $6.09$ & $41.63$ \\
256 & 0.90 & $12.18$ & $83.65$ \\
256 & 0.95 & $29.55$ & $97.17$ \\
\bottomrule
\end{tabular}
\end{subtable}\hfill
\begin{subtable}[t]{0.32\textwidth}
\centering
\caption{$Q=16$}
\label{tab:rollout-linear-preference-q16}
\begin{tabular}{rrrr}
\toprule
$C$ & $p$ & Raw (\%) & Cum-max (\%) \\
\midrule
4 & 0.70 & $49.83$ & $30.26$ \\
4 & 0.80 & $5.89$ & $2.30$ \\
4 & 0.90 & $4.42$ & $3.25$ \\
4 & 0.95 & $4.36$ & $6.69$ \\
\addlinespace[2pt]
8 & 0.70 & $6.89$ & $2.67$ \\
8 & 0.80 & $2.95$ & $2.17$ \\
8 & 0.90 & $3.76$ & $5.36$ \\
8 & 0.95 & $3.73$ & $8.79$ \\
\addlinespace[2pt]
16 & 0.70 & $1.80$ & $1.43$ \\
16 & 0.80 & $2.26$ & $3.57$ \\
16 & 0.90 & $3.25$ & $7.08$ \\
16 & 0.95 & $3.76$ & $17.36$ \\
\addlinespace[2pt]
64 & 0.70 & $0.47$ & $1.96$ \\
64 & 0.80 & $1.69$ & $16.67$ \\
64 & 0.90 & $10.81$ & $53.89$ \\
64 & 0.95 & $24.41$ & $83.21$ \\
\addlinespace[2pt]
256 & 0.80 & $46.68$ & $96.72$ \\
256 & 0.90 & $74.31$ & $99.84$ \\
256 & 0.95 & $90.93$ & $100.00$ \\
\bottomrule
\end{tabular}
\end{subtable}\hfill
\begin{subtable}[t]{0.32\textwidth}
\centering
\caption{$Q=32$}
\label{tab:rollout-linear-preference-q32}
\begin{tabular}{rrrr}
\toprule
$C$ & $p$ & Raw (\%) & Cum-max (\%) \\
\midrule
4 & 0.70 & $47.78$ & $27.04$ \\
4 & 0.80 & $4.15$ & $1.91$ \\
4 & 0.90 & $3.15$ & $3.88$ \\
4 & 0.95 & $3.39$ & $8.42$ \\
\addlinespace[2pt]
8 & 0.70 & $5.59$ & $2.39$ \\
8 & 0.80 & $1.82$ & $2.49$ \\
8 & 0.90 & $2.66$ & $6.55$ \\
8 & 0.95 & $3.21$ & $10.94$ \\
\addlinespace[2pt]
16 & 0.70 & $1.41$ & $1.60$ \\
16 & 0.80 & $1.84$ & $4.13$ \\
16 & 0.90 & $2.32$ & $10.47$ \\
16 & 0.95 & $7.54$ & $22.64$ \\
\addlinespace[2pt]
64 & 0.70 & $0.55$ & $8.65$ \\
64 & 0.80 & $4.31$ & $35.40$ \\
64 & 0.90 & $22.18$ & $74.70$ \\
64 & 0.95 & $51.15$ & $93.01$ \\
\addlinespace[2pt]
256 & 0.80 & $67.74$ & $98.88$ \\
256 & 0.90 & $89.13$ & $99.96$ \\
256 & 0.95 & $96.57$ & $100.00$ \\
\bottomrule
\end{tabular}
\end{subtable}\par\medskip

\begin{subtable}[t]{0.32\textwidth}
\centering
\caption{$Q=64$}
\label{tab:rollout-linear-preference-q64}
\begin{tabular}{rrrr}
\toprule
$C$ & $p$ & Raw (\%) & Cum-max (\%) \\
\midrule
4 & 0.70 & $44.09$ & $22.85$ \\
4 & 0.80 & $2.85$ & $1.88$ \\
4 & 0.90 & $2.32$ & $4.98$ \\
4 & 0.95 & $2.83$ & $10.80$ \\
\addlinespace[2pt]
8 & 0.70 & $4.32$ & $2.22$ \\
8 & 0.80 & $1.26$ & $2.97$ \\
8 & 0.90 & $2.26$ & $8.55$ \\
8 & 0.95 & $3.24$ & $14.33$ \\
\addlinespace[2pt]
16 & 0.70 & $1.17$ & $1.90$ \\
16 & 0.80 & $1.47$ & $5.08$ \\
16 & 0.90 & $3.08$ & $14.16$ \\
16 & 0.95 & $12.45$ & $30.91$ \\
\addlinespace[2pt]
64 & 0.70 & $3.11$ & $29.47$ \\
64 & 0.80 & $19.73$ & $62.84$ \\
64 & 0.90 & $53.82$ & $90.22$ \\
64 & 0.95 & $78.69$ & $98.55$ \\
\addlinespace[2pt]
256 & 0.80 & $85.46$ & $99.73$ \\
256 & 0.90 & $96.49$ & $100.00$ \\
256 & 0.95 & $99.34$ & $100.00$ \\
\bottomrule
\end{tabular}
\end{subtable}\hfill
\begin{subtable}[t]{0.32\textwidth}
\centering
\caption{$Q=128$}
\label{tab:rollout-linear-preference-q128}
\begin{tabular}{rrrr}
\toprule
$C$ & $p$ & Raw (\%) & Cum-max (\%) \\
\midrule
4 & 0.70 & $39.41$ & $18.83$ \\
4 & 0.80 & $2.06$ & $2.00$ \\
4 & 0.90 & $2.11$ & $6.66$ \\
4 & 0.95 & $2.72$ & $11.85$ \\
\addlinespace[2pt]
8 & 0.70 & $3.29$ & $2.23$ \\
8 & 0.80 & $1.05$ & $3.83$ \\
8 & 0.90 & $2.32$ & $9.64$ \\
8 & 0.95 & $7.48$ & $17.28$ \\
\addlinespace[2pt]
16 & 0.70 & $1.05$ & $2.58$ \\
16 & 0.80 & $1.40$ & $6.04$ \\
16 & 0.90 & $8.27$ & $19.96$ \\
16 & 0.95 & $24.26$ & $45.98$ \\
\addlinespace[2pt]
64 & 0.70 & $13.19$ & $43.44$ \\
64 & 0.80 & $40.36$ & $75.22$ \\
64 & 0.90 & $73.65$ & $94.26$ \\
64 & 0.95 & $90.56$ & $99.28$ \\
\addlinespace[2pt]
256 & 0.80 & $95.44$ & $99.98$ \\
256 & 0.90 & $99.39$ & $100.00$ \\
256 & 0.95 & $99.95$ & $100.00$ \\
\bottomrule
\end{tabular}
\end{subtable}\hfill
\begin{subtable}[t]{0.32\textwidth}
\centering
\caption{$Q=256$}
\label{tab:rollout-linear-preference-q256}
\begin{tabular}{rrrr}
\toprule
$C$ & $p$ & Raw (\%) & Cum-max (\%) \\
\midrule
4 & 0.70 & $32.93$ & $15.39$ \\
4 & 0.80 & $1.54$ & $2.04$ \\
4 & 0.90 & $2.26$ & $8.24$ \\
4 & 0.95 & $3.94$ & $11.57$ \\
\addlinespace[2pt]
8 & 0.70 & $2.46$ & $2.59$ \\
8 & 0.80 & $1.17$ & $4.68$ \\
8 & 0.90 & $4.08$ & $11.08$ \\
8 & 0.95 & $15.96$ & $22.85$ \\
\addlinespace[2pt]
16 & 0.70 & $1.27$ & $3.31$ \\
16 & 0.80 & $2.70$ & $7.78$ \\
16 & 0.90 & $20.52$ & $29.79$ \\
16 & 0.95 & $49.83$ & $62.56$ \\
\addlinespace[2pt]
64 & 0.70 & $33.61$ & $58.35$ \\
64 & 0.80 & $63.20$ & $83.81$ \\
64 & 0.90 & $87.25$ & $96.84$ \\
64 & 0.95 & $96.89$ & $99.64$ \\
\addlinespace[2pt]
256 & 0.80 & $99.42$ & $100.00$ \\
256 & 0.90 & $99.97$ & $100.00$ \\
256 & 0.95 & $100.00$ & $100.00$ \\
\bottomrule
\end{tabular}
\end{subtable}
\end{table}

\subsection{Nemotron Top-\texorpdfstring{$p$}{p} Scaling}
\begin{figure}[H]
    \centering
    \includegraphics[width=0.5\linewidth]{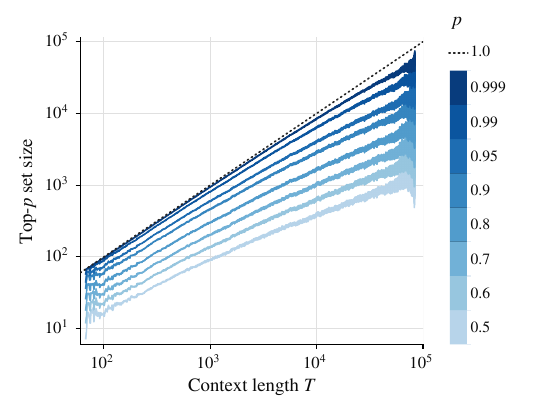}
\caption{Dense top-$p$ set size for Nemotron-3.5-Lightning-30B-A3B using the same six-book cohort as \cref{fig:dense-top-p-growth}.}
    \label{fig:dense-top-p-nemotron-heads}
\end{figure}

\Cref{fig:dense-top-p-nemotron-heads} repeats the dense analysis for Nemotron-3.5-Lightning-30B-A3B on the same six books.
Here, we find that while top-$p$ set sizes again grow sublinearly, the stronger log-log curvature and late-context fluctuations make a single power law less descriptive.
Sublinear growth thus need not follow a clean power law.
Although the models differ in positional encoding, this comparison does not isolate it as the cause.

\section{Training and Evaluation Details}
\label[appendix]{app:training-evaluation}

\subsection{Training Setup}
\label[appendix]{app:training-compute}

\Cref{tab:training-stats} summarizes the training regime for the four ZHA routers and the MSA indexer used in our evaluations.
Qwen denotes Qwen3.6-35B-A3B and Nemotron denotes Nemotron-3.5-Lightning-30B-A3B.

\begin{table}[H]
  \centering
  \small
  \setlength{\tabcolsep}{4pt}
  
  \begin{tabular}{lccccc}
    \toprule
     & \multicolumn{2}{c}{Qwen ZHA} & \multicolumn{2}{c}{Nemotron ZHA} & Qwen MSA \\
    \cmidrule(lr){2-3} \cmidrule(lr){4-5} \cmidrule(lr){6-6}
     & Deep & Wide & Deep & Wide & Index \\
    \midrule
    Trained module & ref.\ queries & ref.\ queries & ref.\ queries & ref.\ queries & index proj.\ + norms \\
    Geometry & C4 L6 & C256 L1 & C4 L6 & C256 L1 & B128 top-16 \\
    Sparse-attention layers & 10 / 40 & 10 / 40 & 6 / 52 & 6 / 52 & 10 / 40 \\
    Trainable params & 81{,}920 & 81{,}920 & 24{,}576 & 24{,}576 & 7.87M \\
    Frozen backbone params & 34.66B & 34.66B & 31.58B & 31.58B & 34.66B \\
    \midrule
    Sequence length & 32{,}768 & 32{,}768 & 32{,}768 & 32{,}768 & 32{,}768 \\
    Batch, rows (tokens) & 16 (0.52M) & 16 (0.52M) & 16 (0.52M) & 16 (0.52M) & 320 (10.5M) \\
    Steps & 1{,}900 & 1{,}900 & 1{,}900 & 1{,}900 & 3{,}815 \\
    Training tokens & 0.996B & 0.996B & 0.996B & 0.996B & 40.0B \\
    \midrule
    AdamW $\epsilon$ & $1{\times}10^{-15}$ & $1{\times}10^{-15}$ & $1{\times}10^{-15}$ & $1{\times}10^{-15}$ & $1{\times}10^{-8}$ \\
    Peak learning rate & $3.2{\times}10^{-2}$ & $3.2{\times}10^{-2}$ & $3.2{\times}10^{-2}$ & $3.2{\times}10^{-2}$ & $3{\times}10^{-4}$ \\
    LR schedule & WSD & WSD & WSD & WSD & cosine \\
    Weight decay & 0 & 0 & 0 & 0 & 0.1 / 0 \\
    Gradient clipping (norm) & 1.0 & 1.0 & 1.0 & 1.0 & 1.0 \\
    \midrule
    Final KL (ZHA: level sum) & 1.172 & 0.242 & 0.850 & 0.179 & 0.120 \\
    \bottomrule
  \end{tabular}

\caption{Training setup and compute for the evaluated ZHA routers and MSA index.
Backbones are frozen; ZHA trains only reference queries.
Deep and wide ZHA use R16.}
  \label{tab:training-stats}
\end{table}

\paragraph{Training recipe.}
ZHA is trained by minimizing KL to dense attention over tree nodes at each level. This is a local objective applied to each attention block independently. By default, unless otherwise specified, ZHA shares 16 reference queries per KV group. The same reference queries are used for all levels of the tree. We train at a context length of 32,768 tokens using document-masked attention. Unless otherwise specified, the backbone is in BF16 and the routing parameters are in FP32. We train for 1,900 steps, totaling approximately 1 billion tokens, and we use the final checkpoint of each training run for evaluation.

For MSA, we use 128-token blocks and select the top 16 blocks per query. We perform warmup by minimizing token-level KL to dense attention over all causal keys; max pooling within blocks is used for selection.

\paragraph{Optimization.}
We train with the AdamW optimizer~\citep{loshchilov2019decoupled} using $\beta=(0.9,0.95)$ and gradient-norm clipping at 1.0.
ZHA's reference-query learning rate is $3.2\times10^{-2}$, with no weight decay.
We use the warmup--stable--decay (WSD) learning rate schedule~\citep{hu2024minicpm} with 20 warmup steps and linear decay over the final 20\% of steps to 1\% of the peak learning rate. 
MSA uses a peak learning rate of $3\times10^{-4}$, 200 warmup steps, and cosine decay to 10\% of peak; weight decay is 0.1 on projections and zero on norm gains. The MSA paper does not report its optimizer settings. We chose ours from the best of three 40B-token pilot runs that varied the learning rate, batch size and schedule.
Nemotron's stored FP32 MoE correction biases are frozen and excluded from the trainable parameter count.

\paragraph{Data accounting.} For our Qwen and Nemotron ZHA training runs, we sample approximately 1B tokens (3.45M documents) from a 100B-token mid-training mixture we used for the ZAYA1-8B model~\citep{zaya1} for each model. We retokenized the data with the appropriate tokenizer. We trained ZHA for a single epoch over this sampled corpus, and MSA for 40B tokens. We chose mid-training data since we believed it was likely similar to the data distribution used for final mid-training and SFT stages of the Qwen and Nemotron models prior to RL. 

\paragraph{Training KL.} When we report the final KL, we are averaging the last 100 steps. 
We report ZHA's total sums over tree levels; \cref{tab:training-kl-levels} gives the individual values, with level $\ell=1$ denoting the raw KV cache chunks. The MSA value is its token-level objective, averaged over query positions, KV groups, and MSA layers. These differently aggregated objectives should not be compared directly across depths or methods.

\begin{table}[H]
  \centering
  \small
  \begin{tabular}{lrrrrrr}
    \toprule
    Router & $\ell=1$ & $\ell=2$ & $\ell=3$ & $\ell=4$ & $\ell=5$ & $\ell=6$ \\
    \midrule
    Qwen deep & 0.021 & 0.097 & 0.180 & 0.253 & 0.302 & 0.319 \\
    Qwen wide & 0.242 & -- & -- & -- & -- & -- \\
    Nemotron deep & 0.016 & 0.069 & 0.137 & 0.192 & 0.217 & 0.221 \\
    Nemotron wide & 0.179 & -- & -- & -- & -- & -- \\
    \bottomrule
  \end{tabular}
\caption{Final per-level ZHA training KL, averaged over logged points in the last 100 steps.
}
  \label{tab:training-kl-levels}
\end{table}

\subsection{Evaluation Protocols}
\label[appendix]{app:evaluation-protocols}

\begin{table}[H]
    \centering
\caption{Evaluation datasets, scoring metrics, and inference protocols.}
    \label{tab:eval-protocol}
    \small
    \setlength{\tabcolsep}{3pt}
    \renewcommand{\arraystretch}{1.15}
    \begin{tabular}{@{}p{0.16\textwidth}p{0.23\textwidth}p{0.14\textwidth}p{0.21\textwidth}p{0.18\textwidth}@{}}
        \hline
        \textbf{Benchmark} & \textbf{Evaluated subset} & \textbf{Reported metric} & \textbf{Generation budget} & \textbf{Decoding / harness} \\
        \hline
        AIME 2026~\citep{matharena2026}\textsuperscript{1} & 30 problems, 16 samples each & avg@16 (\%) & 81,920 output tokens & Thinking; temperature 1.0, top-$p$ 0.95, top-$k$ 20, presence penalty 1.5 \\
        SWE-bench Verified~\citep{jimenez2024swebench,chowdhury2024swebenchverified} & 500 tasks, 1 run each & Resolved tasks (\%) & 32,768 tokens/turn; 204,800-token context; 250 steps & mini-swe-agent 2.4.6; bash text actions; thinking retained across turns; temperature 1.0, top-$p$ 0.95, top-$k$ 20 \\
        Terminal-Bench 2.1~\citep{terminalbench21}\textsuperscript{2} & All 89 tasks, 5 attempts each & Mean per-task success rate (\%) & 81,920 tokens/turn; 262,144-token context; 3\,h per trial & Harbor 0.24 / Terminus-2; thinking retained across turns; temperature 1.0, top-$p$ 0.95, top-$k$ 20 \\
        MMLU-Pro~\citep{mmlupro} & Full test split: 12,032 questions; zero-shot CoT; 1 sample each & Accuracy (\%) & 81,920 output tokens & Thinking; temperature 1.0, top-$p$ 0.95, top-$k$ 20 \\
        GPQA Diamond~\citep{gpqa} & All 198 questions, 8 samples each & avg@8 (\%) & 81,920 output tokens & Thinking; temperature 1.0, top-$p$ 0.95, top-$k$ 20 \\
        RULER-13~\citep{ruler} & 13 tasks $\times$ 100 examples at 256K; 64K and 128K in appendix & Mean task score (\%) & Official per-task caps: 30--128 tokens & Greedy; thinking disabled \\
        LongBench v2~\citep{longbenchv2}\textsuperscript{3} & 205 questions with original contexts $\geq$128K tokens; 1 sample each & Accuracy using official answer keys (\%) & 32,768 reasoning + 4,096 answer tokens; prompts capped at 225,280 tokens & Native thinking; temperature 1.0, top-$p$ 0.95, top-$k$ 20, presence penalty 1.5 \\
        \hline
    \end{tabular}
    \par\smallskip
    \begin{minipage}{0.97\textwidth}
        \footnotesize
        \textsuperscript{1}Uses the pinned MathArena dataset revision containing the problem-10 statement error.\par
        \textsuperscript{2}Uses the Qwen3.6 model-card protocol: flat 3\,h per trial and 32 CPUs / 48\,GB per container.\par
        \textsuperscript{3}Prompts exceeding the cap retain their first and last halves.
    \end{minipage}
\end{table}

\paragraph{Evaluation protocols.}
ZHA is applied during both prefill and decoding in our evaluations. AIME~2026 reports avg@16 over 30 problems, grading the final boxed answer after reasoning.
SWE-bench Verified reports resolved tasks out of 500.
Terminal-Bench~2.1 averages five attempts per task over 89 tasks with a three-hour agent limit.
MMLU-Pro uses 12,032 questions, and GPQA Diamond uses 198 questions with eight samples each.
RULER-13 averages 13 official task scores at 256K tokens, with 100 examples per task.
LongBench~v2 uses 205 questions with contexts $\geq$128K tokens, a 32K-token reasoning budget, and a 4K-token answer budget.

\subsection{Qwen3.6-35B-A3B}

\begin{figure}[!ht]
  \centering
  \includegraphics[width=\textwidth]{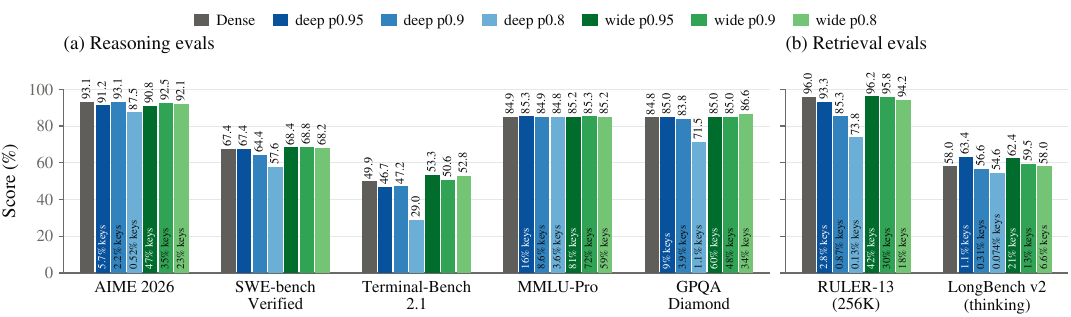}
\caption{Qwen3.6-35B-A3B scores and decode key shares across routing thresholds for deep (C4~L6, blue) and wide (C256~L1, green) ZHA, with R16.
Darker shades indicate higher $p$. (a) Reasoning; (b) retrieval.
Labels above bars show the benchmark scores and labels inside bars give key shares where available.} 
  \label{fig:evaluation-qwen-thresholds}
\end{figure}

\begin{figure}[!ht]
  \centering
  \includegraphics[width=\textwidth]{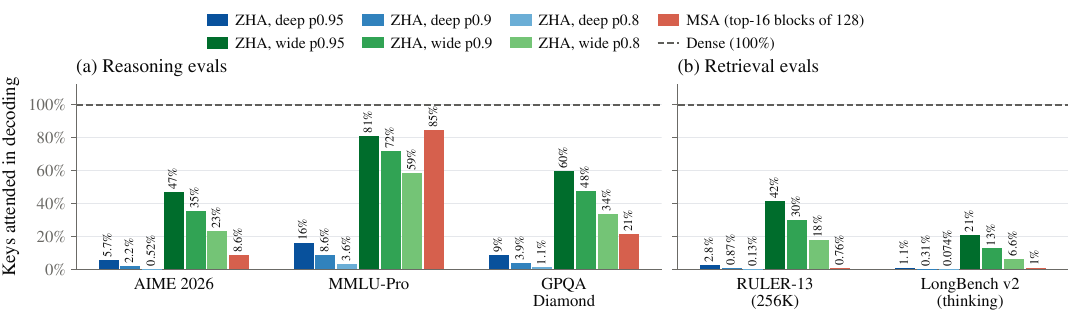}
  \caption{Qwen3.6-35B-A3B decode key shares for deep (C4~L6, blue) and wide (C256~L1, green) ZHA (R16) at $p\in\{0.95,0.9,0.8\}$, compared with MSA (top-16 blocks of 128 tokens). (a) Reasoning; (b) retrieval. Darker shades indicate higher $p$; the dashed line marks dense attention (100\%).}
  \label{fig:evaluation-qwen-key-share}
\end{figure}

\Cref{fig:evaluation-qwen-thresholds} extends the main-text dense vs ZHA comparisons at $p=0.95$ to routing thresholds $p\in\{0.95,0.9,0.8\}$.
Each threshold is applied uniformly at each tree level.
The figure reports both benchmark scores and, where available, the share of keys attended during decoding, showing the tradeoff between accuracy and keys attended. Furthermore it demonstrates how sparsity compounds with depth: at the same routing threshold \(p\), the deep router achieves greater sparsity than the wide router.
We also compare the sparsity of both ZHA geometries with MSA in \cref{fig:evaluation-qwen-key-share}. 
For both figures, blue bars denote the deep router and green bars the wide router; darker shades indicate higher thresholds.

\clearpage
\paragraph{Small-reference router.}\mbox{}\par

\begin{figure}[H]
  \centering
  \includegraphics[width=\linewidth]{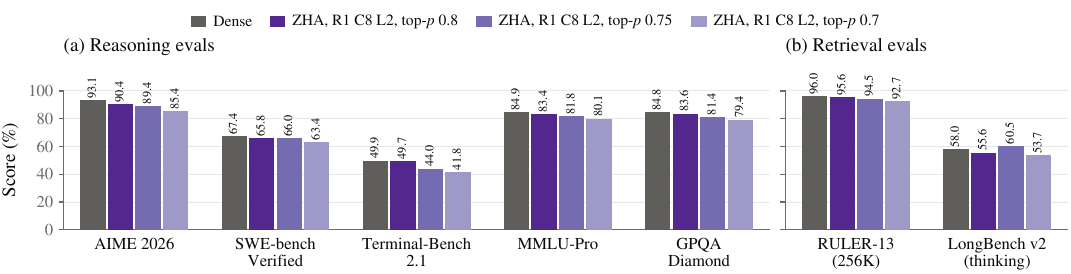}
  \caption{Dense attention vs.\ a ZHA router with one reference query per KV group, 8-token chunks and two levels (R1~C8~L2) on Qwen3.6-35B-A3B, with uniform routing top-$p$ 0.8, 0.75 and 0.7. (a) Reasoning; (b) retrieval.}
  \label{fig:evaluation-qwen-small-router}
\end{figure}

\paragraph{Router precision.}\mbox{}\par

\begin{figure}[H]
  \centering
  \includegraphics[width=\linewidth]{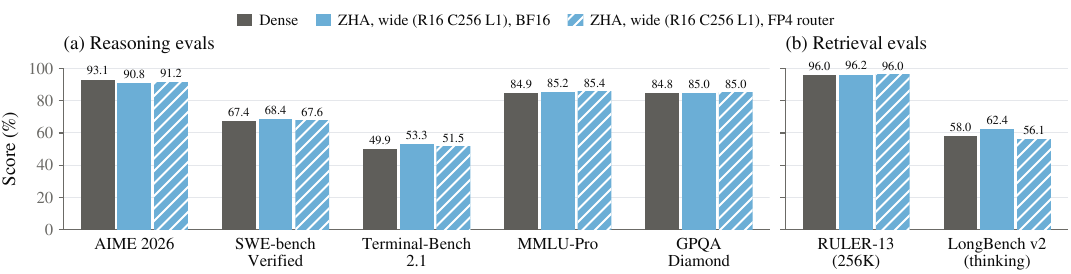}
  \caption{Dense attention vs.\ the wide ZHA router (16 reference queries per KV group, 256-token chunks, one level; top-$p$ 0.95) on Qwen3.6-35B-A3B, with the router in BF16 and in NVFP4 (router operands, reference-key cache and scoring in FP4; backbone, KV cache and the sparse attention in BF16). (a) Reasoning; (b) retrieval.}
  \label{fig:evaluation-qwen-fp4}
\end{figure}

\subsection{Nemotron-3.5-Lightning-30B-A3B}
\label[appendix]{app:evaluation-nemotron}

\begin{figure}[!ht]
  \centering
  \includegraphics[width=\textwidth]{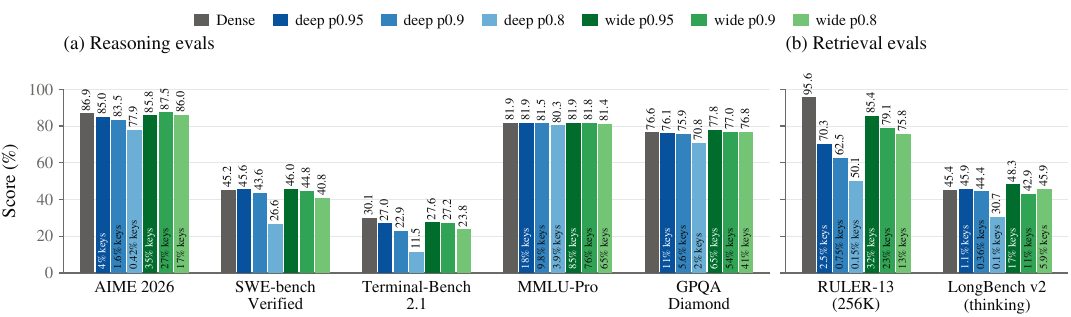}
    \caption{Nemotron-3.5-Lightning-30B-A3B scores and decode key shares across routing thresholds for deep (C4~L6, blue) and wide (C256~L1, green) ZHA, with R16.
    Darker shades indicate higher $p$. (a) Reasoning; (b) retrieval.
    Labels above bars show the benchmark scores and labels inside bars give key shares where available.}
  \label{fig:evaluation-nemotron-thresholds}
\end{figure}

On Nemotron-3.5-Lightning-30B-A3B, which has no explicit positional encodings, ZHA retains performance close to the dense baseline on the reasoning and coding benchmarks. On RULER-13, however, at $p=0.95$, the deep and wide geometries score 25.3 and 10.2 percentage points below the dense baseline, respectively.
\Cref{fig:evaluation-nemotron-thresholds} shows results across routing thresholds.

\clearpage
\section{Geometry, Retrieval Accuracy, and Efficiency}
\label[appendix]{app:geometry}

Here we provide supplementary material for the performance analysis of a variety of ZHA geometries and sparsities, including further distinguishing their performance per retrieval task and studying which geometries are on the sparsity--performance Pareto frontier.

\subsection{Geometry Sweep}

\begin{figure}[H]
  \centering
  \includegraphics[width=\textwidth]{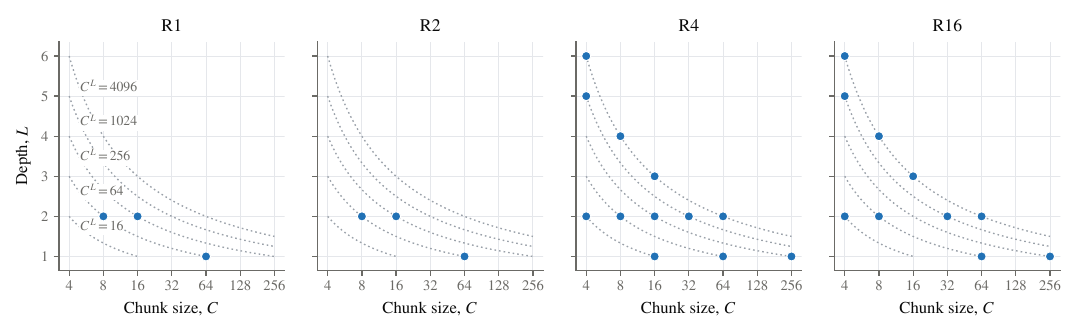}
  \caption{The 28 geometries used in the RULER fits. Each panel fixes the reference count $R$; dotted curves mark constant top-level compression $C^L$. These geometries are evaluated at $p\in\{0.7,0.8,0.9,0.95,0.99\}$ and sequence lengths of 64K, 128K, and 256K.}
  \label{fig:geometry-coverage}
\end{figure}

We are careful to cover a wide range of geometries to obtain enough coverage on the effect of each variable in $(R, C, L, p)$. \cref{fig:geometry-coverage} summarizes the 28 geometries used in our fits, spanning reference counts $R\in\{1,2,4,16\}$, chunk sizes $C$ from 4 to 256, and depths $L$ from 1 to 6. We focus on coverage of top-level compression $C^L$, ranging from 16 to 4096; matched-compression comparisons varying the decomposition into $C$ and $L$; and repeated $(C,L)$ pairs across $R$ to isolate the effect of reference-query expressiveness. Each geometry is evaluated at $p\in\{0.7,0.8,0.9,0.95,0.99\}$ and sequence lengths of 64K, 128K, and 256K.

\subsection{Effect of Geometry on Per-Task RULER Accuracy}

\Cref{sec:evaluation-geometry} modeled retrieval accuracy as a function of ZHA geometry and sparsity using aggregate accuracy across the 13 RULER tasks. However, these tasks vary greatly in their retrieval dynamics and difficulty. Some tasks require not only sharp but narrow retrieval in order to ignore a large set of distractor needles, while other tasks require multi-needle or multi-hop retrieval. As such, it is meaningful to study the accuracy of ZHA geometries per RULER task, as this may reveal particular weaknesses of different geometries and inform which provide the best tradeoff of speed and performance.

\begin{figure}[H]
    \centering
    \includegraphics[width=\textwidth,height=0.75\textheight,keepaspectratio]{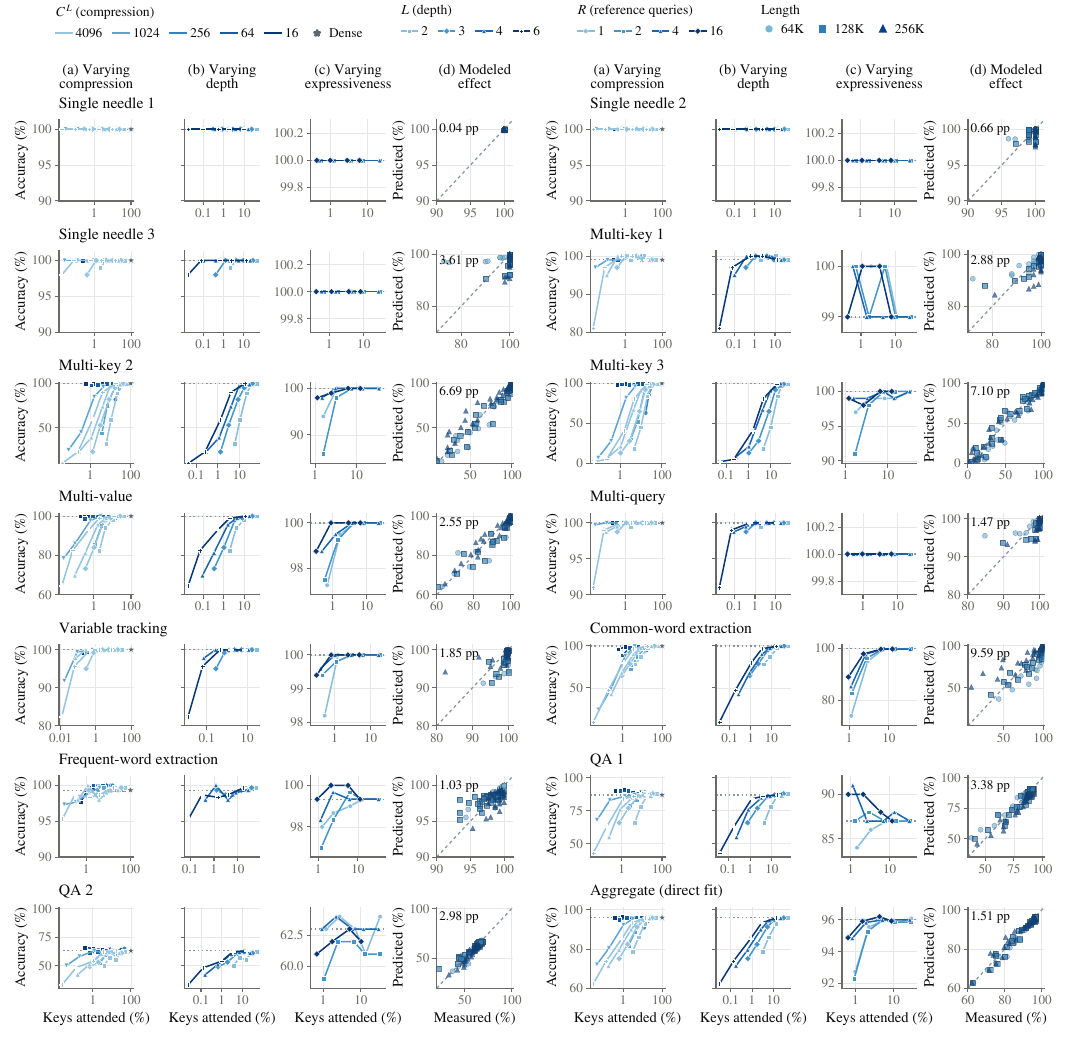}
    \caption{Per-task geometry, sparsity, and prediction accuracy for Qwen3.6-35B-A3B with reference-query-only routing; the final group shows the direct aggregate fit.
    Panels (a), (b), and (d) fix R16.
    (a) Measured accuracy at 256K versus decode keys attended across 11 geometries, colored by $C^L$.
    (b) The matched $C^L=4096$ geometries, colored by depth $L$.
    (c) The C8, L2 slice at 256K, colored by reference count $R$.
    Curves sweep $p\in\{0.7,0.8,0.9,0.95\}$, with $p=0.99$ included where available; markers denote depth in (a,b) and $R$ in (c).
    Key shares use each task's native decode counters (pooled counters for the aggregate), and the key-share axes are logarithmic.
    Dotted horizontal lines show dense attention.
    (d) All 150 leave-one-length-out predictions per response over 64K, 128K, and 256K, colored and marked by held-out length; inset values give RMSE in percentage points and dashed diagonals show equality.
    Each task has four independent parameters in \cref{eq:geometry-accuracy-model}, fitted at the other two lengths across $R\in\{1,2,4,16\}$ using the measured dense baseline at the held-out length.
    Accuracy uses 100 examples per task; vertical-axis ranges vary by task.}
    \label{fig:geometry-ruler-tasks}
\end{figure}

\Cref{fig:geometry-ruler-tasks}(a)--(c) show the same views of geometry's effect on retrieval accuracy that separate the effect of varying compression ($C^L$), depth ($L$), and per-node expressiveness ($R$); panel (d) shows the simple geometry-to-performance model's quality of fit to the empirical RULER accuracy data (in this case, per task). The first observation is the variety in the ``lossless sparsity'' (i.e., without accuracy loss) permitted by each task; some admit good performance with a fraction of a percent of dense keys attended (e.g., single needle 1) whereas others see decay by 10\% of keys attended (e.g., common-word extraction). This reflects that ZHA's efficacy is bounded by the inherent compressibility of the underlying data. Nevertheless, note also that the same trends with geometry remain, and are consistent across geometries: greater compression can hurt performance, and increasing depth or expressiveness enables greater sparsity at fixed performance. Furthermore, though there is some variability in the fitting quality per task (often when there is little spread in accuracy, whether high or low), the geometry-to-performance model is broadly a good fit to the empirical RULER accuracy across tasks and sequence lengths.

\subsection{Geometry--Performance Pareto Frontier}

\begin{figure}[H]
  \centering
  \includegraphics[width=\textwidth]{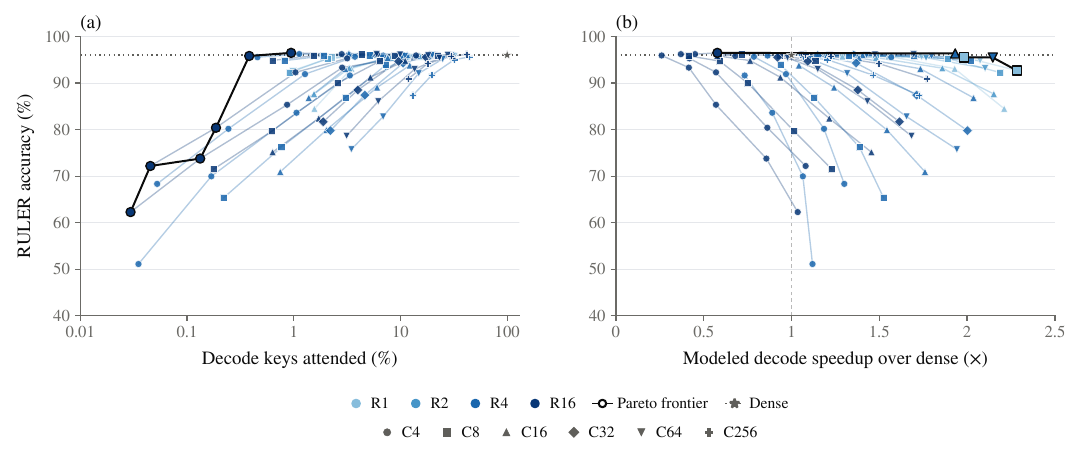}
  \caption{Measured RULER-13 accuracy at 256K versus (a) measured decode key share and (b) modeled decode-attention speedup over dense for Qwen3.6-35B-A3B (115 settings, 28 geometries).
Speedups are H100 BF16 attention-layer projections at batch size one, not measured end-to-end timings; missing higher-level union counts use the slower modeled scenario.
Colors denote $R$, markers $C$, and curves sweep $p$ within each geometry.
Black lines mark Pareto frontiers; dotted horizontal lines show dense accuracy, and the vertical guide marks $1\times$.}
  \label{fig:geometry-ruler-pareto}
\end{figure}

To obtain a holistic view of the aggregate effect of different geometries on retrieval performance, we study the 28 geometries above together in \cref{fig:geometry-ruler-pareto}, comparing accuracy with decode keys attended and modeled decode-attention speedup. At 256K, the high-accuracy end of the sparsity frontier favors R16~C4~L2, which achieves $95.8\%$ accuracy at $p=0.7$ while attending to $0.38\%$ of keys. However, the deeper L5--L6 variants reach still lower key fractions but with substantially lower accuracy. The modeled speedup frontier instead favors shallower or lower-$R$ trees: at $p=0.7$, R4~C16~L1 gives $96.4\%$ accuracy at $1.93\times$, R4~C64~L1 gives $95.5\%$ at $2.15\times$, and R1~C8~L2 gives $92.7\%$ at $2.28\times$. That these shallower or more lightweight trees are favored over the deeper variants reflects that minimizing the attended key count alone does not maximize speedup, as routing and selection also contribute to attention cost. 

\clearpage
\section{Roofline Model}
\label[appendix]{app:roofline}

We use roofline models to compute the theoretical latencies of various attention mechanisms to estimate their optimized real-world speeds.
Specifically, we build models for dense attention, ZHA, and MSA.
We study prefill and (amortized) decode.

\subsection{Dense}

We begin with dense attention, using the same execution variables as ZHA below.
At fixed workload dimensions, these quantities are deterministic.
\begin{table}[H]
\centering
\caption{Variables for the dense-attention roofline model.}
\label{tab:roofline-dense-variables}
\normalsize
\begin{tabular}{l|p{0.72\linewidth}}
    Variable & Definition \\
    \hline
    $B$ & Batch size. \\
    $T,\ Q(T)$ & Context and query lengths; $Q(T)=1$ in decode and $Q(T)=T$ in full prefill. \\
    $H_q,\ H_{\mathrm{KV}},\ G$ & Query and KV head counts, and GQA group size $G=H_q/H_{\mathrm{KV}}$. \\
    $d_k,\ d_v$ & Key and value dimensions per head. \\
    $s_{\mathrm{QKV}},\ s_{\mathrm{accum}}$ & Bytes per stored Q/K/V or output element, and per accumulator or softmax statistic. \\
    $x$ & Kernel stage: attention, split-KV partials, or merge. \\
    $\lambda_x,\ \lambda_A$ & Per-kernel launch overhead; $\lambda_A=\lambda_{\mathrm{attn}}$ for unsplit attention. \\
    $\Phi_x(T)$ & Execution latency of stage $x$, excluding launches. \\
    $\mathrm{FLOPs}_x(T),\ \mathrm{bytes}_x(T)$ & Operation count and memory traffic of stage $x$. \\
    $P_x,\ \mathrm{BW}_x$ & Effective compute throughput (FLOP/s) and memory bandwidth (bytes/s). \\
    $N_{\mathrm{qtiles},x}(T)$ & Query-side tile count, before key partitioning. \\
    $N_{\mathrm{tiles},x}(T)$ & Launch-grid block count, rather than inner-loop tiles. \\
    $\kappa_{Q,x},\ \kappa_{H,x}$ & Query positions and query heads per tile within one KV group. \\
    $\kappa_{\mathrm{attn}},\ J(T)$ & Keys per split-KV partial block, and splits per query: $J(T)=\lceil T/\kappa_{\mathrm{attn}}\rceil$.
\end{tabular}
\end{table}

\paragraph{Latency.}
For each kernel, the roofline execution time is
\begin{align}
    \Phi_x(T)
    &= \max\left\{
        \frac{\mathrm{FLOPs}_x(T)}{P_x},
        \frac{\mathrm{bytes}_x(T)}{\mathrm{BW}_x}
    \right\}.
\end{align}
Dense attention has no tree construction, routing, or selection.
Its unsplit latency is simply
\begin{equation}
    t_{\mathrm{dense}}(T)=\lambda_A+\Phi_{\mathrm{attn}}(T).
    \label{eq:roofline-dense-unsplit-latency}
\end{equation}
For split-KV decode or noncausal attention, the partial and merge kernels give
\begin{equation}
    t_{\mathrm{dense},\mathrm{split}}(T)
    =\lambda_{\mathrm{partial}}+\Phi_{\mathrm{partial}}(T)
     +\lambda_{\mathrm{merge}}+\Phi_{\mathrm{merge}}(T).
    \label{eq:roofline-dense-split-latency}
\end{equation}
Each stage uses its own effective compute throughput and memory bandwidth.

\paragraph{Work and traffic.}
With rectangular query/head tiling, the query-side count is
\begin{align}
    N_{\mathrm{qtiles},x}(T)
    &= B H_{\mathrm{KV}}
       \left\lceil\frac{Q(T)}{\kappa_{Q,x}}\right\rceil
       \left\lceil\frac{G}{\kappa_{H,x}}\right\rceil
    = \begin{cases}
    B H_\mathrm{KV} & \text{grouped-head decode $(Q(T)=1, \kappa_{H,x}=G)$} \\
    B H_q \left\lceil \frac{T}{\kappa_{Q,x}} \right\rceil & \text{prefill $(Q(T)=T, \kappa_{H,x}=1)$}
    \end{cases}
\end{align}
The $\kappa_{\bullet,x}$ are kernel tile sizes. %

For unsplit attention over all $T$ keys, the stage counts are
\begin{align}
    N_{\mathrm{tiles},\mathrm{attn}}(T)
    &=N_{\mathrm{qtiles},\mathrm{attn}}(T), \\
    \mathrm{FLOPs}_{\mathrm{attn}}(T)
    &=2B Q(T)H_q(d_k+d_v)T, \\
    \mathrm{bytes}_{\mathrm{attn}}(T)
    &=s_{\mathrm{QKV}}(d_k+d_v)
      \left[N_{\mathrm{qtiles},\mathrm{attn}}(T)T+B Q(T)H_q\right].
\end{align}

For causal prefill with one-head query tiles, let $n(T)=\lceil T/\kappa_{Q,\mathrm{attn}}\rceil$.
Set $\mathrm{FLOPs}_{\mathrm{attn}}(T)=BH_q(d_k+d_v)T(T+1)$ and replace the KV-read count $N_{\mathrm{qtiles},\mathrm{attn}}(T)T$ in $\mathrm{bytes}_{\mathrm{attn}}(T)$ by
\begin{equation}
    BH_q\left[\frac{\kappa_{Q,\mathrm{attn}}n(T)(n(T)-1)}{2}+T\right].
\end{equation}
The query and output traffic is unchanged.

For split-KV decode or noncausal attention, the counts are
\begin{align}
    N_{\mathrm{tiles},\mathrm{partial}}(T)
    &= N_{\mathrm{qtiles},\mathrm{partial}}(T) J(T),
       \qquad N_{\mathrm{tiles},\mathrm{merge}}(T) = N_{\mathrm{qtiles},\mathrm{merge}}(T), \label{eq:roofline-split-tiles}\\
    \mathrm{FLOPs}_{\mathrm{partial}}(T)
    &= 2 B Q(T) H_q (d_k + d_v) T,
       \qquad \mathrm{FLOPs}_{\mathrm{merge}}(T) = B Q(T) H_q \left[2 J(T) d_v + 3 J(T) - 1\right], \label{eq:roofline-split-flops}\\
    \mathrm{bytes}_{\mathrm{partial}}(T)
    &= s_{\mathrm{QKV}} (d_k + d_v) N_{\mathrm{qtiles},\mathrm{partial}}(T) T
       + B Q(T) H_q \left[s_{\mathrm{QKV}} d_k + s_{\mathrm{accum}} J(T) (d_v + 2)\right], \label{eq:roofline-split-partial-bytes}\\
    \mathrm{bytes}_{\mathrm{merge}}(T)
    &= B Q(T) H_q \left[s_{\mathrm{accum}} J(T) (d_v + 2) + s_{\mathrm{QKV}} d_v\right]. \label{eq:roofline-split-merge-bytes}
\end{align}
For the merge, $P_{\mathrm{merge}}$ is an effective scalar arithmetic rate.
Query loads are assumed cached across splits.
Each partial writes a $d_v$-element vector and two softmax statistics at $s_{\mathrm{accum}}$ bytes per element; the merge reads them and writes the final output at $s_{\mathrm{QKV}}$ bytes per element.
Both launches are included explicitly.

The existing worker-aware calibration was checked against native FlashAttention~\citep{flashattention} on $576$ held-out H100 cases ($144$/row).
Each uses $B=1$, $H_q=32$; $d_k=d_v\in\{64,128,256\}$ and $H_{\mathrm{KV}}\in\{1,4,8,32\}$, with $T=2^9,\ldots,2^{20}$.
The error is $e=100(t_{\mathrm{model}}/t_{\mathrm{Flash}}-1)$; calibration uses separate cases.
\begin{table}[H]
\centering
\begin{tabular}{ll|rrr}
    Mode & Storage & Median $e$ & Median $|e|$ & $90$th percentile $|e|$ \\
    \hline
    Causal prefill & BF16 & $-0.9\%$ & $15.2\%$ & $57.2\%$ \\
    Causal prefill & FP16 & $-1.0\%$ & $16.5\%$ & $54.9\%$ \\
    Decode & BF16 & $12.9\%$ & $12.9\%$ & $54.1\%$ \\
    Decode & FP16 & $13.2\%$ & $13.3\%$ & $53.8\%$
\end{tabular}
\end{table}
Note that decode includes the profiled split/merge geometry.
These are all attention-kernel timings, not end-to-end model timings.

\clearpage
\begin{figure}[p]
    \centering
    \captionsetup{font=small}
    \begin{subfigure}[t]{0.48\textwidth}
        \centering
        \includegraphics[page=1,width=\linewidth,height=0.36\textheight,keepaspectratio]{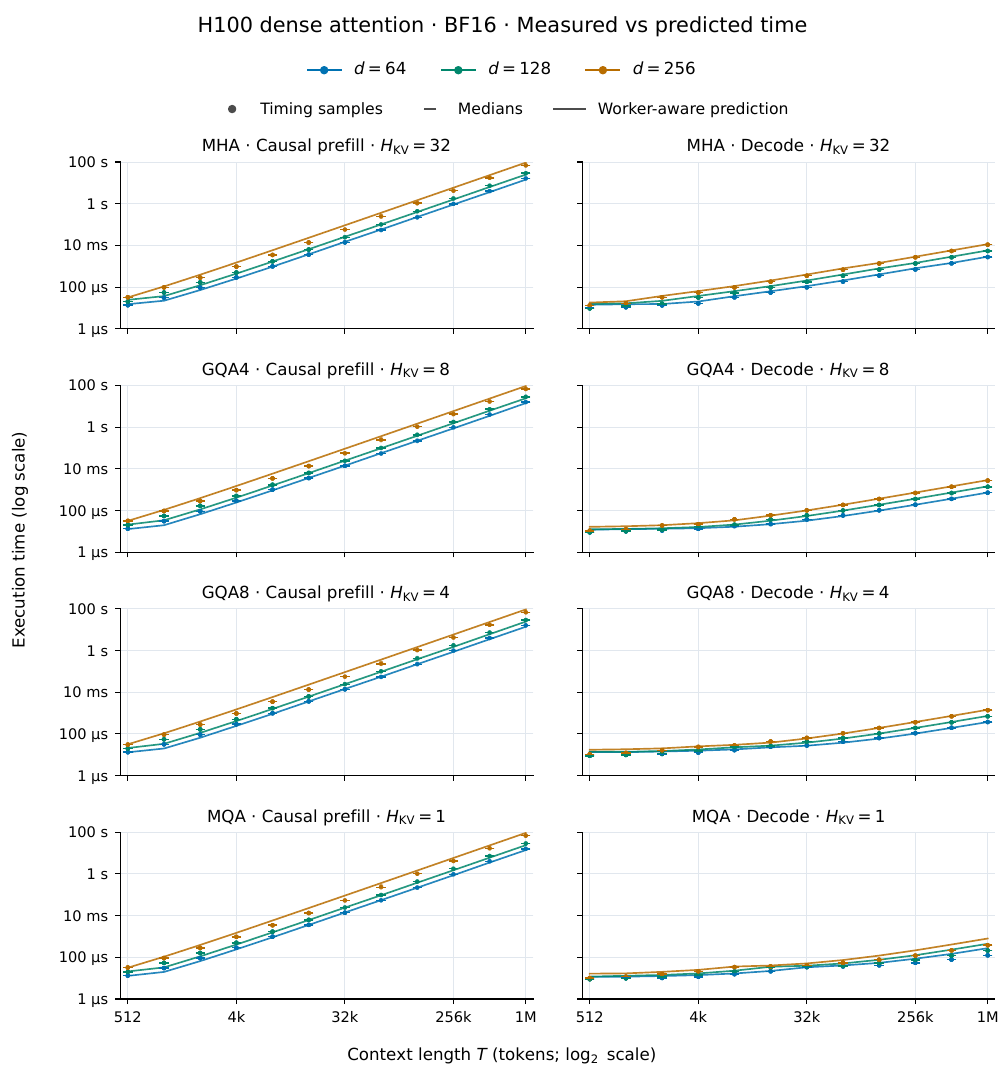}
        \caption{BF16 latency.}
        \label{fig:roofline-dense-bf16-latency}
    \end{subfigure}\hfill
    \begin{subfigure}[t]{0.48\textwidth}
        \centering
        \includegraphics[page=2,width=\linewidth,height=0.36\textheight,keepaspectratio]{figures/attention-roofline/dense-validation.pdf}
        \caption{FP16 latency.}
        \label{fig:roofline-dense-fp16-latency}
    \end{subfigure}

    \medskip
    \begin{subfigure}[t]{0.48\textwidth}
        \centering
        \includegraphics[page=3,width=\linewidth,height=0.36\textheight,keepaspectratio]{figures/attention-roofline/dense-validation.pdf}
        \caption{BF16 timing/prediction.}
        \label{fig:roofline-dense-bf16-ratio}
    \end{subfigure}\hfill
    \begin{subfigure}[t]{0.48\textwidth}
        \centering
        \includegraphics[page=4,width=\linewidth,height=0.36\textheight,keepaspectratio]{figures/attention-roofline/dense-validation.pdf}
        \caption{FP16 timing/prediction.}
        \label{fig:roofline-dense-fp16-ratio}
    \end{subfigure}
    \caption{Held-out dense-attention latency and timing divided by prediction.
    Within each subfigure, rows are MHA, GQA4, GQA8, and MQA; columns are causal prefill and decode.
    Colors distinguish $d\in\{64,128,256\}$.
    Dots show all recorded timing samples and short ticks mark medians; solid lines in the latency views show worker-aware predictions.
    Each CUDA-event sample averages a replay batch, not an individual invocation.
    The lower views show $t_{\mathrm{Flash}}/t_{\mathrm{model}}$; the dashed $1\times$ line denotes exact prediction, and values below it indicate overprediction.
    Matching views retain the same layout, colors, marks, and axis limits across precisions; the ratio views share the same ratio scale.}
    \label{fig:roofline-dense-validation}
\end{figure}

\clearpage

\subsection{ZHA}
We now extend the dense model to ZHA, including tree construction, routing, and selection.
We model it as
\begin{align}
    \bm{t}_{\mathrm{ZHA}}(T)
    &= t_{\mathrm{tree}}(T)
    + \sum_{\ell=1}^{L(T)}
    \underbrace{ \left[\lambda_\ell + \bm{\Phi}_{\mathrm{routing},\ell}(T) + \bm{\Phi}_{\mathrm{scan},\ell}(T)\right] }_\text{Level $\ell$ of tree}
    + \underbrace{\lambda_A + \bm{\Phi}_{\mathrm{attn}}(T)}_\text{Final attention}.
    \label{eq:roofline-zha-latency}
\end{align}
We use $\lambda$ to represent kernel launch overhead and $\bm{\Phi}$ to capture execution time, accounting for compute, memory bandwidth, and available parallel workers.
Boldface indicates data-dependent random quantities, including deterministic special cases.
Workload dimensions, kernel choices, and calibration coefficients are fixed, not random, in this model.
Context-length dependence is explicit, in particular, $Q(T)$ and $L(T)$ are deterministic functions set by the workload and tree configuration, whereas dependence on the input and query is implicit, with routing parameters fixed.
For fusion, replace routing and selection by one execution term with its own worker and launch parameters.
Variable definitions are provided in \cref{tab:roofline-zha-variables,tab:roofline-worker-variables,tab:roofline-tree-variables}.
We choose this model for a few reasons:
\begin{enumerate}
\item The sum over levels captures the inherent \emph{serial} processing through the depth of the tree.
\item It separates kernel launch overhead from execution time, i.e., it does not assume perfect pipelining, which may be challenging with our architecture.
\item It captures data-dependent sparsity and latency.
\item It accounts for potential under-utilization of a GPU at higher levels of the tree.
\end{enumerate}
\begin{table}[H]
\centering
\caption{Variables for the ZHA roofline model.}
\label{tab:roofline-zha-variables}
\small
\begin{tabular}{l|p{0.72\linewidth}}
    Variable & Definition \\
    \hline
    $B$ & Batch size. \\
    $T,\ Q(T)$ & Context and query lengths; $Q(T)=1$ in decode and $Q(T)=T$ in full prefill. \\
    $H_q,\ H_{\mathrm{KV}}$ & Numbers of query heads and key/value heads, respectively. \\
    $d_k,\ d_v$ & Query/key and value dimensions per head, respectively. \\
    $G$ & Query heads per KV head (GQA group size): $G=H_q/H_{\mathrm{KV}}$. \\
    $s_{\mathrm{QKV}}$ & Bytes per stored Q/K/V, compressed-key, reference-bias, or routing-score element; BF16 baseline: $s_{\mathrm{QKV}}=2$. \\
    $s_{\mathrm{accum}}$ & Bytes per stored accumulator or softmax/tree statistic element. \\
    $R$ & Number of reference queries. \\
    $C$ & Tree chunk size: tokens per leaf chunk and children per internal node. \\
    $L(T),\ \ell$ & Configured tree depth at context length $T$ and level index, respectively. A fixed-depth configuration has constant $L(T)$. \\
    $p,\ K$ & Selection threshold and fixed-count selection budget, respectively, when used by the chosen selector. \\
    $\bm{t}_{\mathrm{ZHA}}(T)$ & Total modeled attention-subsystem latency, including tree work and launches. \\
    $t_{\mathrm{tree}}(T)$ & Deterministic tree-construction cost for prefill or amortized maintenance cost for decode, including launches. \\
    $\bm{\Phi}_{\mathrm{routing},\ell}(T)$ & Routing execution; scores $R$ references for each of the $\bm{K}_\ell(T)$ candidate nodes at level $\ell$. \\
    $\bm{\Phi}_{\mathrm{scan},\ell}(T)$ & Selection execution; scans the scored frontier and applies selection %
    to decide which nodes to retain or expand. \\
    $\bm{\Phi}_{\mathrm{attn}}(T)$ & Exact-attention execution over the $\bm{K}_0(T)$ retained tokens, excluding launch overhead. \\
    \hline
    $\bm{K}_\ell(T)$ & Candidate nodes before selection at $\ell\ge1$; retained tokens at $\ell=0$. Per query/head or a grouped union, depending on context. \\
    $\bm{S}_\ell(T)$ & Nodes retained after selection at $\ell\ge1$. Frontier statistics must satisfy the expansion relation below; they are not independent free counts. \\
    $\lambda_\ell,\ \lambda_A$ & Total launch overhead for routing/selection at level $\ell$ and final attention, respectively; sum the actual kernel launches.
\end{tabular}
\par\smallskip

\end{table}

With $C$ children per retained node, expansion gives
\begin{align}
    \bm{K}_{\ell-1}(T) &= C\,\bm{S}_\ell(T),
    \qquad 1\le\ell\le L(T).
\end{align}
At $\ell=1$, the children are raw tokens.
Partially filled nodes contribute only their actual children.

\subsubsection{Worker-aware Execution}
For a sparse stage $x$, with the routing level $\ell$ implicit when applicable, we use the same worker-aware maximum:
\begin{align}
    \bm{\Phi}_x(T)
    &= \max\left(
        \bm{t}_{x,\mathrm{worker\ waves}}(T),
        \frac{\mathbf{FLOPs}_x(T)}{P_x},
        \frac{\mathbf{bytes}_x(T)}{\mathrm{BW}_x}
    \right).
\end{align}
This is a standard roofline model except for the $\bm{t}_{x,\mathrm{worker\ waves}}(T)$ term.
In $\mathbf{bytes}_x(T)$, each stored Q/K/V, compressed-key, reference-bias, or routing-score element contributes $s_{\mathrm{QKV}}$ bytes per transfer.
Accumulation precision is separate from storage width; intermediates kept on-chip incur no HBM traffic.
The throughput, occupancy, and block-latency parameters correspond to each stage's operand and accumulation types.
For worker waves, we use
\begin{align}
    \bm{t}_{x,\mathrm{worker\ waves}}(T)
    &= \bm{\tau}_x(T)\left\lceil
        \frac{\bm{N}_{\mathrm{tiles},x}(T)}{N_{\mathrm{SM}}c_x}
      \right\rceil.
\end{align}
For fixed work per block, $\bm{\tau}_x(T)=\tau_x$ is constant in this model.
\begin{table}[H]
\centering
\caption{Kernel and worker variables for the roofline model.}
\label{tab:roofline-worker-variables}
\small
\begin{tabular}{l|p{0.72\linewidth}}
    Variable & Definition \\
    \hline
    $x$ & Kernel label: routing, scan, attention, split partial/merge, fusion, or tree summarization/merge. Sparse stage quantities also carry $\ell$ when level-specific. \\
    $\bm{\Phi}_x(T)$ & Sparse-stage execution latency, excluding kernel launches. \\
    $\bm{t}_{x,\mathrm{worker\ waves}}(T)$ & Worker-wave contribution to sparse-stage execution latency. \\
    $\mathbf{FLOPs}_x(T),\ \mathbf{bytes}_x(T)$ & Sparse-stage operation count and memory traffic (bytes). \\
    $\bm{N}_{\mathrm{tiles},x}(T)$ & Product of launch-grid dimensions, not inner-loop tiles. In 1-D: $\lceil \bm{n}_x(T)/\kappa_x\rceil$. Unsplit attention is a deterministic special case. \\
    $\bm{n}_x(T)$ & Work-item count for a generic 1-D partition. \\
    $N_{\mathrm{qtiles},x}(T)$ & Deterministic query-side tile count, before key/frontier partitioning. \\
    $N_{\mathrm{SM}}$ & Number of GPU execution units: SMs on NVIDIA, CUs on AMD. \\
    $\bm{Q}(T),\ \bm{K}(T),\ \bm{V}(T)$ & Query, key, and value matrices; $\bm{Q}(T)$ is distinct from the scalar query length $Q(T)$. \\
    $\bm{K}_c(T)$ & Compressed-key matrix used in routing. \\
    $\bm{k}_i,\ \bm{k}_i^{(\ell)},\ i$ & Raw-token key, compressed key at level $\ell$, and key index within the level, respectively. These are stored, fixed-index keys. \\
    \hline
    $\kappa_x$ & Items per block for a generic 1-D partition. \\
    $\kappa_{Q,x},\ \kappa_{H,x}$ & Query positions and query heads per tile, respectively, within one KV group. \\
    $\kappa_{\mathrm{routing}},\ \kappa_{\mathrm{scan}}$ & Compressed keys per routing block and node scores per scan block. \\
    $\kappa_{\mathrm{attn}},\ \kappa_{\mathrm{radix}}$ & Retained tokens per split-KV block and fixed full-frontier radix pass count, respectively. \\
    $\lambda_x$ & Per-kernel launch overhead; also applies to the tree kernels. \\
    $P_x,\ \mathrm{BW}_x$ & Effective compute throughput (FLOP/s) and memory bandwidth (bytes/s). \\
    $c_x$ & Effective blocks (AMD: workgroups) resident per SM/CU; limited by registers, shared memory/LDS, and hardware limits. \\
    $\tau_x,\ \bm{\tau}_x(T)$ & Per-block execution latency at concurrency $c_x$, excluding launches: fixed $\tau_x$ for fixed work, or $\bm{\tau}_x(T)$ for data-dependent work.
\end{tabular}

\end{table}

For reference, the following are full-device hardware limits, not calibrated kernel parameters.\footnote{Sources:
\href{https://www.nvidia.com/content/dam/en-zz/Solutions/Data-Center/nvidia-ampere-architecture-whitepaper.pdf}{A100 architecture},
\href{https://developer.nvidia.com/blog/nvidia-hopper-architecture-in-depth/}{H100 architecture},
\href{https://docs.nvidia.com/cuda/ampere-tuning-guide/index.html}{Ampere tuning},
\href{https://docs.nvidia.com/cuda/hopper-tuning-guide/index.html}{Hopper tuning},
\href{https://rocm.docs.amd.com/en/latest/reference/gpu-specs.html}{AMD specifications}, and
\href{https://rocm.docs.amd.com/projects/HIP/en/docs-7.0.2/reference/hardware_features.html}{HIP hardware limits}.}
\begin{table}[H]
\centering
\small
\begin{tabular}{p{0.34\linewidth}|cccc}
    Quantity & A100 SXM4 & H100 SXM5 & MI300X & MI300A \\
    \hline
    Workers $N_{\mathrm{SM}}$ & 108 SMs & 132 SMs & 304 CUs & 228 CUs \\
    Hardware bound on $c_x$ & $\le32$ & $\le32$ & $\le32$ & $\le32$
\end{tabular}
\end{table}

For separate kernels, we use the provisional occupancy assumptions
\begin{align}
    c_x &=
    \begin{cases}
        2, & x\in\{\mathrm{routing},\mathrm{scan}\}, \\
        1, & x=\mathrm{attn}.
    \end{cases}
\end{align}
These values are not GPU-independent constants: a different GPU or kernel can change $c_x$, and $\bm{\tau}_x(T)$ must be calibrated at that concurrency.

Using the query-side tile counts defined above, use $\kappa_{\mathrm{routing}}$ compressed keys per routing block and $\kappa_{\mathrm{scan}}$ node scores per scan block:
\begin{align}
    \bm{N}_{\mathrm{tiles},\mathrm{routing},\ell}(T)
    &= N_{\mathrm{qtiles},\mathrm{routing}}(T)\left\lceil   \frac{R\bm{K}_\ell(T)}{\kappa_{\mathrm{routing}}}   \right\rceil, \\
    \bm{N}_{\mathrm{tiles},\mathrm{scan},\ell}(T)
    &= N_{\mathrm{qtiles},\mathrm{scan}}(T)\left\lceil	  \frac{\bm{K}_\ell(T)}{\kappa_{\mathrm{scan}}}   \right\rceil, \\
    \bm{N}_{\mathrm{tiles},\mathrm{attn}}(T)
    &= N_{\mathrm{qtiles},\mathrm{attn}}(T)
       \begin{cases}
           1, & \text{unsplit attention}, \\
           \left\lceil \bm{K}_0(T)/\kappa_{\mathrm{attn}}\right\rceil,
               & \text{split-KV partials}.
       \end{cases}
\end{align}
Here $\kappa_{\mathrm{attn}}$ is retained tokens per split-KV block; split-KV also needs a separately timed merge pass with $N_{\mathrm{qtiles},\mathrm{attn}}(T)$ blocks.
Unsplit attention instead loops over retained keys within each block, so $\bm{K}_0(T)$ affects $\bm{\tau}_{\mathrm{attn}}(T)$, not its block count.
FLOPs and block latency include all packed query positions and heads.
These are per-pass counts; additional selection or reduction passes add their own execution and launch costs.

\paragraph{FLOPs.}

The dominant matrix work is
\begin{align}
    \mathbf{FLOPs}_{\mathrm{routing},\ell}(T)
    &\simeq 2B Q(T) H_q R d_k\,\bm{K}_\ell(T), \\
    \mathbf{FLOPs}_{\mathrm{attn}}(T)
    &\simeq 2B Q(T) H_q(d_k+d_v)\,\bm{K}_0(T).
\end{align}
The routing contribution is $\bm{Q}(T)\bm{K}_c(T)^\top$; selection is accounted for separately.
The two attention contributions are $\bm{Q}(T)\bm{K}(T)^\top$ and multiplication of the attention weights by $\bm{V}(T)$. %

For radix selection, $\mathbf{FLOPs}_{\mathrm{scan},\ell}(T)$ counts the ordinary floating-point arithmetic of the chosen implementation. $P_{\mathrm{scan}}$ is a scalar arithmetic rate, not Tensor Core throughput.
Exponentials, bit operations, comparisons, and index handling are reflected in $\bm{\tau}_{\mathrm{scan},\ell}(T)$.

\paragraph{Bytes.} For traffic, assume attention outputs and reference biases use the same storage precision $s_{\mathrm{QKV}}$, measured in bytes.
For streaming radix selection, we have to do $\kappa_{\mathrm{radix}}$ full-frontier passes,
each reading one stored score per candidate.
The byte counts are
\begin{align}
    \mathbf{bytes}_{\mathrm{routing},\ell}(T)
    &= \underbrace{ R s_{\mathrm{QKV}} (d_k + 1) N_{\mathrm{qtiles},\mathrm{routing}}(T) \bm{K}_\ell(T) }_\text{compressed keys + biases}
       + \underbrace{ s_{\mathrm{QKV}} d_k B Q(T) H_q }_\text{input queries} \\ %
    \mathbf{bytes}_{\mathrm{scan},\ell}(T)
    &= s_{\mathrm{QKV}} \kappa_{\mathrm{radix}} B Q(T) H_q \bm{K}_\ell(T), \\
    \mathbf{bytes}_{\mathrm{attn},\mathrm{unsplit}}(T)
    &= s_{\mathrm{QKV}} (d_k + d_v)
       \left[ N_{\mathrm{qtiles},\mathrm{attn}}(T) \bm{K}_0(T) + B Q(T) H_q \right], \\
    \mathbf{bytes}_{\mathrm{attn},\mathrm{split}}(T)
    &= \mathbf{bytes}_{\mathrm{attn},\mathrm{unsplit}}(T)
       + 2 s_{\mathrm{accum}} B Q(T) H_q \left\lceil \frac{ \bm{K}_0(T) }{ \kappa_{\mathrm{attn}} } \right\rceil (d_v + 2).
\end{align}
Here $\lceil \bm{K}_0(T) / \kappa_{\mathrm{attn}} \rceil$ is the number of KV splits, and $s_{\mathrm{accum}}$ denotes the bytes per stored accumulator/statistic element.

Split-KV writes and rereads a $d_v$-element partial vector and two softmax statistics per query/head and split.
Query loads are assumed cached; partial and merge kernels are timed separately.

These leading payload counts include reference biases but omit index/mask reads, retained-ID or mask writes, quantization scales, and algorithm-specific scratch.

\subsubsection{Tree construction}

In addition to routing through the tree, we must consider the cost of constructing it.
Use the baseline stable-state merge, without Taylor/RoPE re-anchoring, with reference queries already prepared.
A leaf computes reference/key scores and weighted key sums: $4RCd_k$ leading FLOPs per query head.
A parent rescales and sums $C$ child weighted sums independently for each reference: $2RCd_k$ leading scalar FLOPs.

Per reference, retain a maximum, weight sum, entropy, and $d_k$-vector, using $s_{\mathrm{accum}}$ bytes per element for subsequent merges.
Emit normalized reference keys at $s_{\mathrm{QKV}}$ bytes per element in the same kernel, rather than an additional mean key or a later normalization pass.
The counts below include the leaf kernel's two key reads and the parent kernel's repeated maximum read.
\begin{table}[H]
\centering
\caption{Variables for tree construction and maintenance.}
\label{tab:roofline-tree-variables}
\small
\begin{tabular}{l|p{0.72\linewidth}}
    Variable & Definition \\
    \hline
    $n,\ n_\ell(T)$ & Full nodes per sequence; the balanced full-node count at level $\ell$ is $n_\ell(T)=\lfloor T/C^\ell\rfloor$. \\
    $N_{\mathrm{tiles},x}(n)$ & Deterministic launch-grid size for a tree kernel processing $n$ full nodes per sequence. \\
    $\mathrm{FLOPs}_x(n),\ \mathrm{bytes}_x(n)$ & Deterministic operation count and memory traffic for that tree kernel. \\
    $t_{\mathrm{summ}}(n),\ t_{\mathrm{merge}}(n)$ & Leaf-summarization and parent-merge latencies, respectively, including launches. \\
    $t_{\mathrm{tree}}^{\mathrm{prefill}}(T)$ & Cost of constructing all full nodes in prefill. \\
    $\overline t_{\mathrm{tree}}^{\mathrm{decode}}(T)$ & Tree-maintenance cost per decoded token, amortized over full-node completions.
\end{tabular}
\par\smallskip
Tree kernels use the free parameters in \cref{tab:roofline-worker-variables}, with $x=\mathrm{summ}$ or $x=(\mathrm{tree},\mathrm{merge})$.

\end{table}

For $n > 0$ full nodes per sequence, with $G = H_q / H_{\mathrm{KV}}$, the kernel counts are
\begin{align}
    N_{\mathrm{tiles},\mathrm{summ}}(n)
    &= B H_{\mathrm{KV}} n,
       \qquad N_{\mathrm{tiles},\mathrm{tree},\mathrm{merge}}(n) = B H_q n, \\
    \mathrm{FLOPs}_{\mathrm{summ}}(n)
    &= 4 B H_q n R C d_k,
       \qquad \mathrm{FLOPs}_{\mathrm{tree},\mathrm{merge}}(n) = 2 B H_q n R C d_k, \\
    \mathrm{bytes}_{\mathrm{summ}}(n)
    &= B H_q n \left[ 2 s_{\mathrm{QKV}} d_k (C / G + R) + s_{\mathrm{accum}} R (d_k + 3) \right], \\
    \mathrm{bytes}_{\mathrm{tree},\mathrm{merge}}(n)
    &= B H_q n R \left[ s_{\mathrm{accum}} C (d_k + 4)
       + s_{\mathrm{accum}} (d_k + 3) + s_{\mathrm{QKV}} d_k \right].
\end{align}
The kernel costs are
\begin{align}
    t_{\mathrm{summ}}(n)
    &= \lambda_{\mathrm{summ}} +
       \max\left\{
           \tau_{\mathrm{summ}} \left\lceil
               \tfrac{N_{\mathrm{tiles},\mathrm{summ}}(n)}
                    {N_{\mathrm{SM}} c_{\mathrm{summ}}}
           \right\rceil,
           \frac{\mathrm{FLOPs}_{\mathrm{summ}}(n)}{P_{\mathrm{summ}}},
           \frac{\mathrm{bytes}_{\mathrm{summ}}(n)}{\mathrm{BW}_{\mathrm{summ}}}
       \right\}, \\
    t_{\mathrm{merge}}(n)
    &= \lambda_{\mathrm{tree},\mathrm{merge}} +
       \max\left\{
           \tau_{\mathrm{tree},\mathrm{merge}} \left\lceil
               \tfrac{N_{\mathrm{tiles},\mathrm{tree},\mathrm{merge}}(n)}
                    {N_{\mathrm{SM}} c_{\mathrm{tree},\mathrm{merge}}}
           \right\rceil,
           \frac{\mathrm{FLOPs}_{\mathrm{tree},\mathrm{merge}}(n)}
                {P_{\mathrm{tree},\mathrm{merge}}},
           \frac{\mathrm{bytes}_{\mathrm{tree},\mathrm{merge}}(n)}
                {\mathrm{BW}_{\mathrm{tree},\mathrm{merge}}}
       \right\}.
\end{align}
Leaf blocks share keys across $G$ query heads; parent blocks are per query head.
Softmax, entropy, and normalization enter the calibrated block latencies; $P_{\mathrm{tree},\mathrm{merge}}$ is a scalar rate, not a Tensor Core rate.
Both costs are zero for $n=0$.

For the balanced full-node model, let $n_\ell(T)=\lfloor T/C^\ell\rfloor$.
Prefill batches all nodes at each level; decode averages one completed node per sequence every $C^\ell$ tokens:
\begin{align}
    t_{\mathrm{tree}}^{\mathrm{prefill}}(T)
    &=t_{\mathrm{summ}}(n_1(T))
      +\sum_{\ell=2}^{L(T)}t_{\mathrm{merge}}(n_\ell(T)),\\
    \overline t_{\mathrm{tree}}^{\mathrm{decode}}(T)
    &=\frac{t_{\mathrm{summ}}(1)}{C}
      +\sum_{\ell=2}^{L(T)}\frac{t_{\mathrm{merge}}(1)}{C^\ell}\\
    &=\frac{t_{\mathrm{summ}}(1)}{C}
      +\frac{t_{\mathrm{merge}}(1)}{C(C-1)}
       \left(1-C^{-(L(T)-1)}\right).
\end{align}
These counts exclude partial boundary nodes.
The decode expression is an average, not a tail latency; an unchanged resident tree has $t_{\mathrm{tree}}(T)=0$.

\subsection{MSA}
\label[appendix]{app:roofline-msa}

MSA uses one index-query head per KV group and \emph{one shared index-key head}, both of width $d_I$.
It scores every visible token, max-pools within $C$-token blocks, then selects $k$ blocks per group.
The local block occupies one of these slots. %

The shared workload and hardware variables retain the definitions in \cref{tab:roofline-dense-variables,tab:roofline-worker-variables}; \cref{tab:roofline-msa-variables} defines the additional notation used here.
The block size $C$ and selection counts below are local to MSA.
\begin{table}[H]
\centering
\caption{Additional variables for the MSA roofline model.}
\label{tab:roofline-msa-variables}
\normalsize
\begin{tabular}{l|p{0.72\linewidth}}
    Variable & Definition \\
    \hline
    $C,\ k$ & Tokens per block and maximum selected blocks per KV group, including the local block. \\
    $M(T),\ k_T$ & Visible block count and selected block count at context length $T$. \\
    $r(T),\ K(T)$ & Tokens in the final, local block and total retained tokens per KV group. \\
    $d_I,\ d_{\mathrm{model}}$ & Index-head width and model hidden width. \\
    $W_I$ & Combined index-projection width: $W_I=(H_{\mathrm{KV}}+1)d_I$. \\
    $\kappa_{\mathrm{proj}}$ & Index-projection output channels per tile. \\
    $\kappa_I$ & Tokens per index-scan tile; a multiple of $C$. \\
    $\kappa_{H,I},\ \kappa_{H,A}$ & Index-query heads and main-attention query heads per tile, respectively. \\
    $\nu_I$ & Index-query tile count in decode: $\nu_I=B\lceil H_{\mathrm{KV}}/\kappa_{H,I}\rceil$. \\
    $s,\ s_c,\ s_{\mathrm{id}}$ & Bytes per stored element ($s=s_{\mathrm{QKV}}$), pooled score, and integer block ID. \\
    $x$ & Kernel stage: index projection ($\mathrm{proj}$), index scan ($I$), top-$k$, or main attention ($A$). \\
    $N_x,\ F_x,\ V_x$ & Stage launch-grid block count, FLOPs, and memory traffic in bytes. \\
    $t_{\mathrm{sparse}}^{\mathrm{MSA}}(K)$ & Main-attention latency over $K$ retained tokens, including its launches. \\
    $k_i,\ t_j$ & Selected blocks at position $i$, $k_i=\min(k,M(i))$, and final query position in index-query tile $j$. \\
    $C_b,\ \bm n_b$ & Stored length of block $b$ and number of queries selecting it within its batch and KV group. \\
    $\kappa_{\mathrm{packed}},\ \bm a_b$ & Query-head pairs per packed tile and packed tiles for block $b$: $\bm a_b=\lceil G\bm n_b/\kappa_{\mathrm{packed}}\rceil$. \\
    $E$ & Total query--block selections, summed over batch and KV groups: $E=\sum_b\bm n_b$.
\end{tabular}
\end{table}

For decode ($Q=1$), define
\begin{align}
    M(T)&=\left\lceil T/C\right\rceil,
    & k_T&=\min(k,M(T)),\\
    r(T)&=T-C(M(T)-1),
    & K(T)&=C(k_T-1)+r(T).
\end{align}
Thus $K(T)=T$ when all blocks fit, and $K(T)\le kC$ otherwise.
The selected addresses vary, but these counts are deterministic.
The decode stage counts are:
\begin{table}[H]
\centering
\small
\begin{tabular}{l|ccc}
    Stage $x$ & $N_x$ & $F_x$ & $V_x$ \\
    \hline
    Index projections
    & $B\lceil W_I/\kappa_{\mathrm{proj}}\rceil$
    & $2Bd_{\mathrm{model}}W_I$
    & $s[d_{\mathrm{model}}W_I+B(d_{\mathrm{model}}+W_I)]$ \\
    Index + max-pool
    & $\nu_I\lceil T/\kappa_I\rceil$
    & $2BH_{\mathrm{KV}}d_IT$
    & $sd_I(\nu_IT+BH_{\mathrm{KV}})+s_cBH_{\mathrm{KV}}M$ \\
    Top-$k$
    & $BH_{\mathrm{KV}}$ & --- &
    $BH_{\mathrm{KV}}(s_cM+s_{\mathrm{id}}k_T)$
\end{tabular}
\end{table}
The two index projections are fused into one launch; cached historical index keys are not reprojected.
Projection weights and inputs, and index queries across key tiles, are assumed cached within their stage.
Only pooled block scores reach HBM, not a $T$-element token-score vector per index query.
Pooling comparisons and exp-free top-$k$ work are charged through their block latencies; they are not Tensor Core FLOPs.
The top-$k$ row assumes each group's candidate list fits on-chip.

Using the same worker-aware maximum, with stage-specific parameters,
\begin{gather}
    \Phi_x(T)=\max\left\{
        \tau_x(T)\left\lceil\tfrac{N_x(T)}{N_{\mathrm{SM}}c_x}\right\rceil,
        \frac{F_x(T)}{P_x},\frac{V_x(T)}{\mathrm{BW}_x}
    \right\},\\
    t_{\mathrm{MSA}}(T)
    =\sum_{x\in\{\mathrm{proj},I,\mathrm{top}k\}}
        [\lambda_x+\Phi_x(T)]
        +t_{\mathrm{sparse}}^{\mathrm{MSA}}(K(T)).
\end{gather}
Omit $F_x/P_x$ for top-$k$, whose scalar work is included in $\tau_{\mathrm{top}k}(M,k_T)$.
There is one top-$k$ launch, not $k_T$ launches.
For query-outer main attention, the unsplit counts are
\begin{align}
    N_A&=BH_{\mathrm{KV}}\left\lceil G/\kappa_{H,A}\right\rceil,\\
    F_A&=2BH_q(d_k+d_v)K(T),\\
    V_A&=s(d_k+d_v)\,[N_AK(T)+BH_q].
\end{align}
For split-KV, use the partial/merge counts in \crefrange{eq:roofline-split-tiles}{eq:roofline-split-merge-bytes}, replacing the \emph{key length} by $K(T)$, not the query length.
The same two-statistic partial-state convention applies.
As elsewhere, final-attention index reads, gather penalties, and padding are not separately counted; they require kernel-specific calibration.

For causal prefill, sum the visible per-position counts:
\begin{align}
    F_{\mathrm{proj}}^{\mathrm{prefill}}&=2BTd_{\mathrm{model}}W_I,\\
    F_I^{\mathrm{prefill}}&=BH_{\mathrm{KV}}d_IT(T+1),\\
    F_A^{\mathrm{prefill}}&=
        2BH_q(d_k+d_v)\sum_{i=1}^{T}K(i).
\end{align}
Pooled-score and selected-ID counts become $BH_{\mathrm{KV}}\sum_i M(i)$ and $BH_{\mathrm{KV}}\sum_i k_i$, respectively.
If index query tile $j$ ends at $t_j$, its key-read count uses $t_j$ and its key-grid count uses $\lceil t_j/\kappa_I\rceil$; do not multiply the final-position decode cost by $T$.

Native MSA prefill instead uses KV-outer scheduling.
Let $\bm n_b$ be the number of queries selecting block $b$ within its KV group, and $\bm a_b=\lceil G\bm n_b/\kappa_{\mathrm{packed}}\rceil$ its packed query tiles.
Sums below also include batch and KV group; $C_b$ is the stored block length and $E=\sum_b\bm n_b=BH_{\mathrm{KV}}\sum_i k_i$.
For one CTA per packed query tile, the partial-attention grid and traffic are
\begin{align}
    \bm N_{\mathrm{partial}}&=\sum_b\bm a_b,\\
    \mathbf{bytes}_{\mathrm{partial}}
    &=s(d_k+d_v)\sum_b C_b\bm a_b
      +sd_kGE+s_{\mathrm{accum}}(d_v+2)GE.
\end{align}
Add reverse-index/scheduling launches and scratch traffic, then combine $k_i$ partials per query using the merge model.
Native chunking or persistent schedules must use their actual launch grids, partial counts, and traffic.
Block popularity therefore affects worker waves and KV reuse, even at fixed $k$; the decode comparison below does not claim to reproduce this native prefill schedule.

\clearpage
\section{Sparsity Scaling Across Routing Thresholds}
\label[appendix]{app:sparsity-scaling}

The latency extrapolations in \cref{sec:speedup} assume a sublinear power-law scaling at each tree level that changes with depth, and therefore we fit those exponents and model their dependence on depth. Here we extend the main-text $p=0.8$ results to other thresholds, explain how these trends are estimated and validated, and connect them to the latency model.

\subsection{Estimating Per-Level Growth}

Estimating the per-level sublinear exponent from empirical selection rates requires carefully controlling for outliers in sequence length and example count. We seek typical problem-level growth, limiting the influence of unusually long generations or extreme slopes. The Qwen3.6-35B-A3B sweep uses the same 30 AIME 2026 problems, with 16 generations per setting: C4, R16, L2--L6, and $p\in\{0.7,0.8,0.9,0.95\}$ (9,600 completions).

We replay saved generations with single-query tiles to measure shared KV footprints: each selected complete node is counted once across the eight query heads in its group, separately at each position, layer, and tree level. Counts are averaged over two groups, 10 full-attention layers, and complete 256-token bins, excluding the always-retained first key and current partial chunk.

For each problem, we average available generations within each bin and fit a free-intercept log--log slope over 16K--64K (192 native bins). Objective weights $1/T$ approximate equal weighting in log context length. Requiring three complete bins leaves 29--30 problems per setting. Their median slope defines $\alpha_\ell$, weighting problems equally without imputing missing tails.

\subsection{Modeling Levels Within a Tree}

Provided empirically estimated exponents per level, we may fit our depthwise exponent model. For a fixed tree, $\beta$ sets the top exponent and $\varepsilon$ its correction toward the leaves in \cref{eq:sparsity-exponent-compounding}. Equally weighted least squares fits both to the observed $\alpha_\ell$ separately for each geometry and $p$, with signed $\varepsilon$, $0.01\leq\beta\leq1$, and $\beta_\ell\geq0.01$. Note that joint fitting can shift $\beta$ from the observed top slope, and that the factors describe combined pruning and KV-group union effects.

The parameter $r$ determines how much of the accumulated correction carries into the next lower level:
\begin{equation}
    \varepsilon_L=0,\qquad
    \varepsilon_{\ell-1}=\varepsilon+r\varepsilon_\ell,\qquad
    \varepsilon_\ell=
    \begin{cases}
        \varepsilon(1-r^{L-\ell})/(1-r), & r\ne1,\\
        \varepsilon(L-\ell), & r=1.
    \end{cases}
    \label{eq:sparsity-pruning-correction}
\end{equation}

Here $r=0$ gives a constant correction below the top, $r=1/2$ approaches $2\varepsilon$, and $r=1$ gives linear drift. Across 20 settings (80 exponent summaries), their RMSEs, weighting settings equally, are $0.0218$, $0.0130$, and $0.0156$. We favor $r=1/2$ as a compact descriptive choice that balances accuracy across settings: paired bootstraps favor $r=1/2$ over $r=0$, though not clearly over $r=1$; its advantage comes mainly from $p=0.7$, L6.

\begin{figure}[!htbp]
    \centering
    \includegraphics[width=\linewidth]{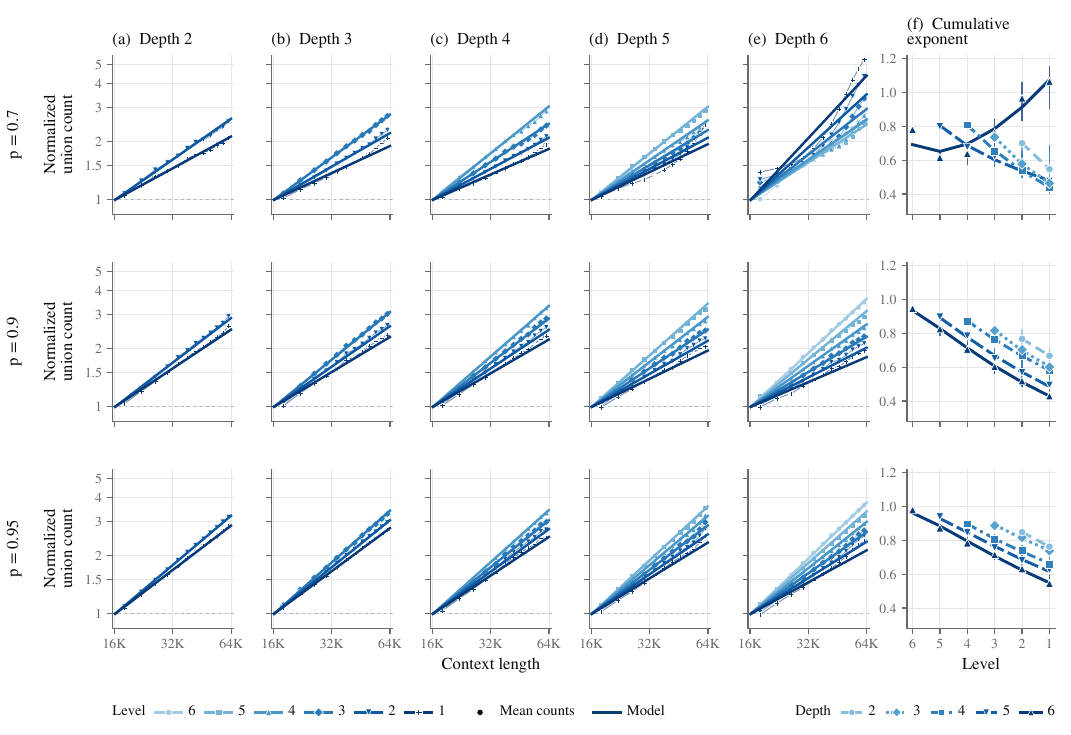}
    \caption{Per-level selection scaling for C4, R16, and $p=0.7,0.9,0.95$, extending \cref{fig:sparsity-scaling}. The first five columns show depths 2--6; the sixth compares observed median exponents with the $r=1/2$ profile. Bars show pointwise 95\% whole-problem bootstrap intervals.}
    \label{fig:sparsity-thresholds}
\end{figure}

\Cref{fig:sparsity-thresholds} shows decreasing exponent profiles at $p=0.9,0.95$; the deepest $p=0.7$ tree instead rises toward the leaves. We normalize each level and geometry by the first-bin count to make growth rates comparable. For visualizations, we further average 16 native-bin means into each of 12 display bins; fits retain native resolution. With profile slopes fixed, amplitudes $c_\ell$ fit pooled arithmetic-mean counts by $1/T$-weighted least squares in log space, weighting available generations equally within each bin.

The results for these other $p$ values broadly corroborate the main-text $p=0.8$ findings; scaling exponents decrease smoothly and predictably proceeding from the top of the tree toward the leaves. The exception is the deepest geometry (C4, L6, R16) at $p=0.7$, whose exponents rise again toward the leaf levels. This setting achieves only 44.0\% accuracy on AIME, compared with 93.1\% for dense attention, so its anomalous scaling should be interpreted in the context of severe accuracy degradation.

\subsection{Adding Tree Depth}

\begin{figure}[!htbp]
    \centering
    \includegraphics[width=\linewidth]{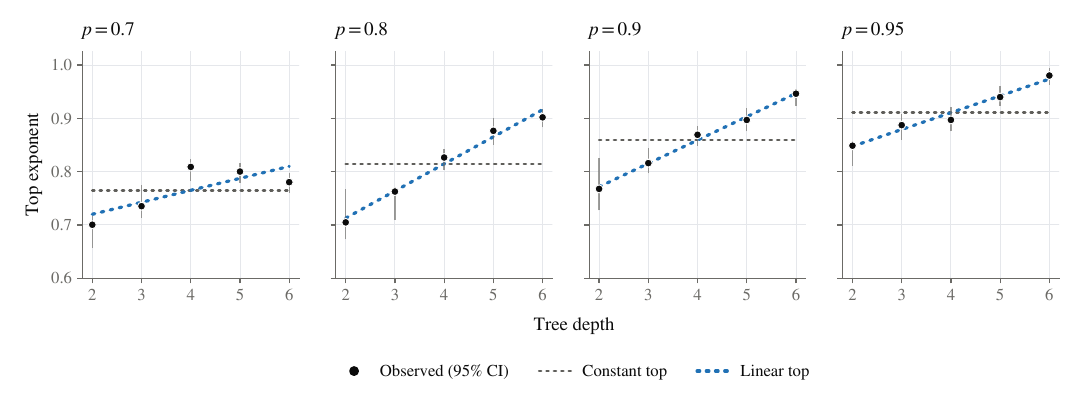}
    \caption{Observed top-level exponents versus total tree depth for C4, R16. Dotted lines fit constant (gray) and linear (blue) depth dependence separately at each $p$, using only the five top-level observations. Bars show pointwise 95\% whole-problem bootstrap intervals.}
    \label{fig:sparsity-top-exponent-depth}
\end{figure}

To estimate selection exponents within fitted trees and extrapolate them to greater depths, we must determine what happens when another level is added. To isolate this effect, \cref{fig:sparsity-top-exponent-depth} fits only the five observed top slopes $\alpha_L$ at each $p$, independently of lower levels:

\begin{equation}
    \widehat{\beta}_{\mathrm{const}}(L,p)=a_p,\qquad
    \widehat{\beta}_{\mathrm{lin}}(L,p)=a_p+b_p(L-4).
    \label{eq:sparsity-top-exponent-thresholds}
\end{equation}

The latency scenarios share the dependence on $p$. Equally weighted least squares on all 20 top slopes gives this form of \cref{eq:sparsity-top-exponent-depth}:

\begin{equation}
    \begin{aligned}
        \widehat{\beta}_{\mathrm{const}}(p)&=0.3736+0.5536p,\\
        \widehat{\beta}_{\mathrm{lin}}(L,p)&=0.3429+0.4129p
        +[0.007697+0.03518p]L.
    \end{aligned}
    \label{eq:sparsity-top-exponent-coefficients}
\end{equation}

The linear-top model follows the observed trend more closely than the constant-top model, reducing top-exponent RMSE from $0.0574$ to $0.0226$. On held-out depths or thresholds beyond the fitted settings, linear-top again improves on constant-top: RMSE $0.0317$ versus $0.0706$ for held-out depths, and $0.0442$ versus $0.0597$ for held-out thresholds.

\subsection{Robustness of the Depth Trend}

Though the trends identified appear robust, it is important to qualify that we fit in a low-data regime. The deepest trees and longest contexts have the least support: L6 has only 4--15 complete top-level candidates over 16K--64K, and fewer generations reach 64K. Many positional bins therefore do not imply many independent examples.

We quantify uncertainty with 2,000 whole-problem bootstrap draws, keeping each sampled problem's generations, bins, and levels together and recomputing the median slopes and model fits. Model comparisons pair draws across depths and thresholds and repeat the holdouts. We take the median which reduces sensitivity to extreme slopes, while still not removing differences in the length ranges covered by each problem.

\subsection{From Selection Counts to Latency}

Having fitted a model for how the sparsity exponents grow with depth, we can plug the implied per-level work into the latency model. We transfer the C4, R16 exponent scenarios by depth and calibrate count amplitudes to native R1, C8, L2 AIME measurements (480 generations per $p$), using pooled counts and the same weighted log fit. Existing C8 checks find that compression-based transfer can improve the top exponent while worsening the full profile. They also use different texts and, for R1, a different reference count, so we do introduce error by transferring between adjacent geometries.

\begin{figure}[!htbp]
    \centering
    \includegraphics[width=252pt]{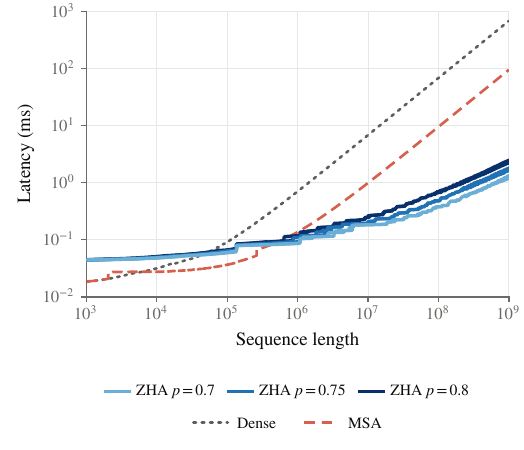}
    \caption{Modeled BF16 decode latency for dense attention, MSA, and ZHA (R1, C8) at $p=0.7,0.75,0.8$, with the H100 assumptions and nondecreasing latency-minimizing depth of \cref{fig:speedup-latency}. Latencies are modeled per-token decode times for the attention subsystem of one layer at batch size 1.}
    \label{fig:speedup-latency-thresholds}
\end{figure}

The plugged-in selection rates model the per-level expansion in the latency model. That is, the model expands parent unions to obtain candidate counts, adds nodes without complete parents, and includes the first key and partial chunk in final token reads.\footnote{At the internal 128K reference, extrapolated from 16K--64K, added-parent occupancy follows $q_{\mathrm{parent}}=1-(1-q_{\mathrm{child}})^8$, where $q$ is the selected fraction of nodes.} \Cref{fig:speedup-latency-thresholds} extends the main-text $p=0.75$ comparison to $p=0.7,0.8$; note that these are tightly clustered relative to the baselines, reflecting their different scaling exponent.

\clearpage
\section{Retrieval Traversal Examples
}
\label[appendix]{app:traversal-examples}

\begin{figure}[!htbp]
    \centering
    \includegraphics[width=\textwidth]{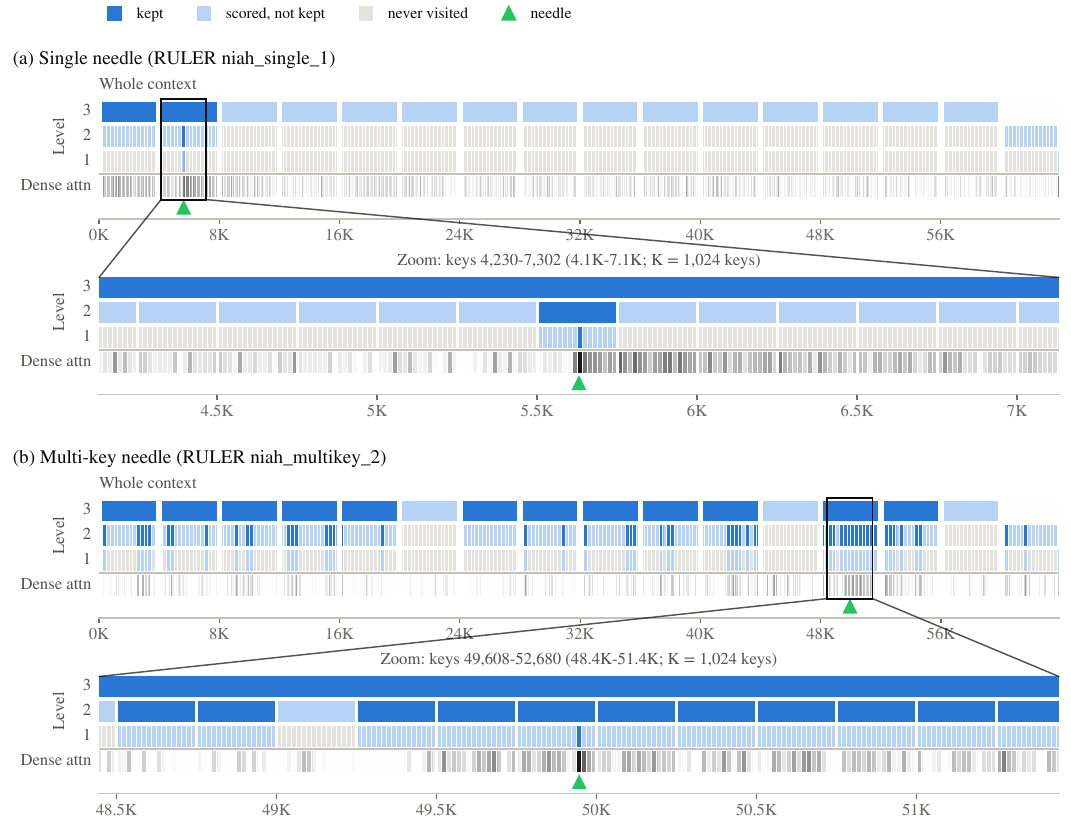}
    \captionsetup{font=small}
    \caption{Retrieval traversals at 64K tokens for (a) single-needle and (b) multi-key 2 RULER tasks (Qwen3.6-35B-A3B, C16~L3, R16, $p = 0.9$). Both show layer 35, query head 8, while emitting the first answer digit. Dark blue blocks are scored and expanded nodes; light blue blocks scored but pruned nodes; gray blocks are nodes which were never scored by the router. Dense-attention rows show chunk-level mass on a logarithmic scale. The green marker indicates the ground-truth needle location. Boxed regions are enlarged below to reveal finer traversal details. These examples illustrate task-dependent routing with top-$p$: for single-needle, the router explores few branches before reaching the answer-containing chunk; for multi-key 2, whose context contains many distractor needles, it explores substantially more upper-level nodes. Both traversals ultimately attend to a single answer-containing chunk.}
    \label{fig:traversal-strip-combined}
\end{figure}

Aggregate key shares show how much of the context ZHA attends to, but not where that budget goes within the tree.
\Cref{fig:traversal-strip-combined} follows one retrieval head through each level of a three-level router on two 64K RULER prompts.
The router's search widens depending on the task difficulty: on the multi-key prompt, whose distractor needles resemble the target, it expands more nodes at multiple regions of the context, while for the single-needle prompt the router is more targeted.
These examples come from a single layer, head and decode step; they are meant to illustrate the mechanism, not to represent typical behavior. Nevertheless, these worked examples demonstrate the power of hierarchical routing in action, where the model is able to accurately find the needle while only activating a small fraction of the total underlying KV cache. Our RULER scores in the main text imply that this retrieval is highly robust in practice, even to challenging multi-key + distractor RULER subtasks.

\clearpage

\clearpage
\section{Reference-Query Agreement Across Tree Levels}
\label[appendix]{app:qref-level-agreement}

We extend the reference-query usage analysis in \cref{fig:qref-selection-rank} to all four geometries and compare aggregate level agreement with individual layer/KV-group examples. Level $\ell$ contains nodes spanning $C^\ell$ keys for the geometries shown. Our analysis demonstrates that a similar pattern of rank-correlation between reference queries across levels holds across a range of tree geometries. 

\begin{figure}[H]
  \centering
  \includegraphics[width=\textwidth]{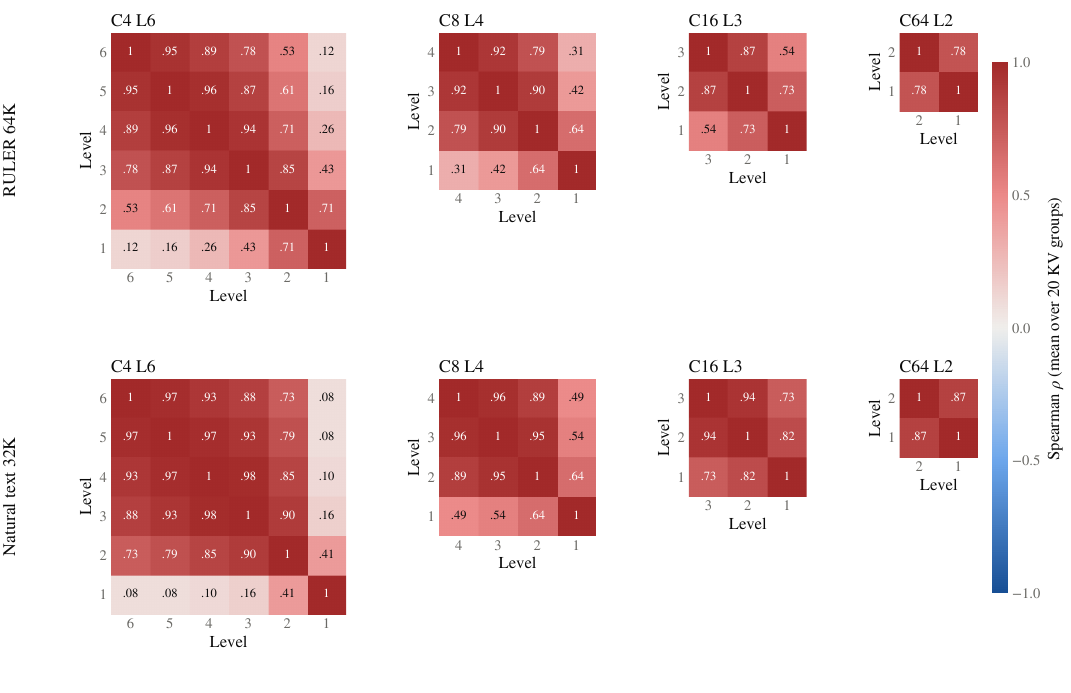}
  \caption{Agreement between tree levels in reference-query selection for Qwen3.6-35B-A3B ZHA (R16). Columns show C4~L6, C8~L4, C16~L3 and C64~L2; rows show RULER at 64K tokens and natural text at 32K tokens. For each layer/KV group, Spearman correlation is computed between the 16 reference-query selection shares at each pair of levels, then averaged over 20 groups (10 layers $\times$ 2 KV groups). Adjacent levels are similar in their reference selection rate while levels farther apart have a different selection priority. This behavior holds across all geometries.}
  \label{fig:qref-level-spearman-geoms}
\end{figure}

\clearpage
\subsection{Individual layer and KV-group examples}

\begin{figure}[H]
  \centering
  \includegraphics[width=\textwidth]{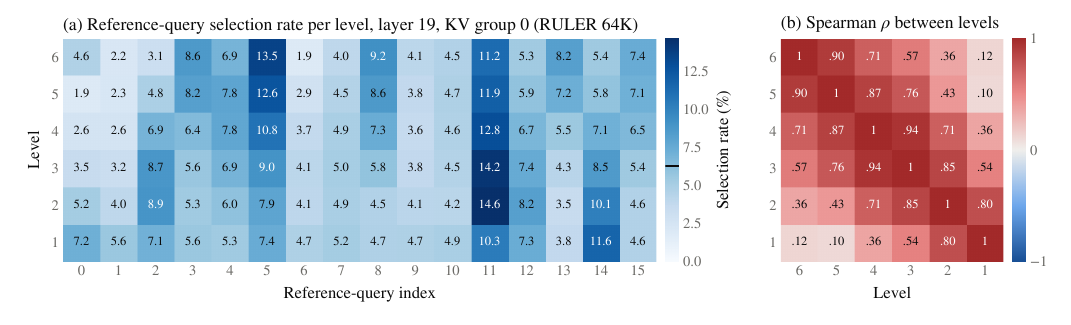}
  \caption{Typical selection behavior across levels: layer 19, KV group 0 of the C4~L6 ZHA router on RULER at 64K tokens. Adjacent levels have similar reference rankings, but agreement weakens over larger level separations; the chunk and top-level rankings are nearly unrelated ($\ell=1$ versus $\ell=6$: $\rho=0.12$).}
  \label{fig:qref-level-rates-l19kv0}
\end{figure}

\begin{figure}[H]
  \centering
  \includegraphics[width=\textwidth]{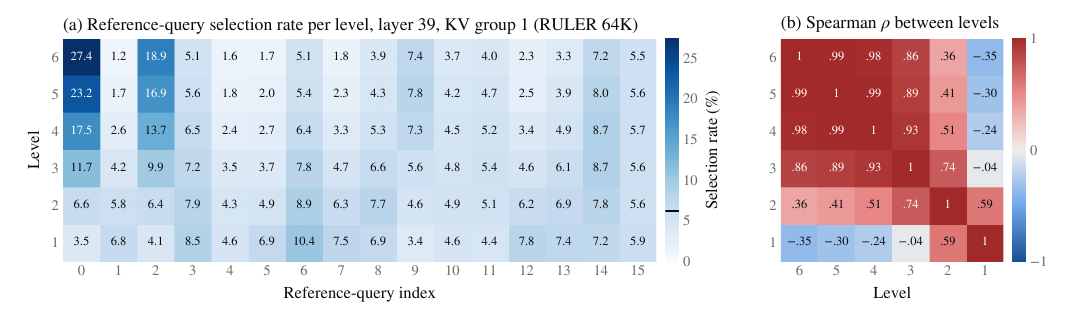}
  \caption{Outlier selection across levels: layer 39, KV group 1 of the C4~L6 ZHA router on RULER at 64K tokens. Upper levels agree closely, while $\ell=1$ reference rankings are negatively correlated with the top levels ($\ell=1$ versus $\ell=6$: $\rho=-0.35$).}
  \label{fig:qref-level-rates-l39kv1}
\end{figure}

\begin{figure}[H]
  \centering
  \includegraphics[width=\textwidth]{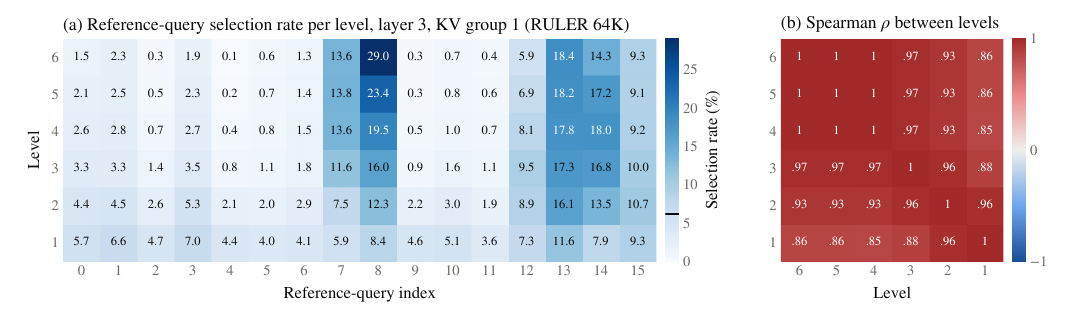}
  \caption{Outlier selection across levels: similar reference rankings throughout the tree in layer 3, KV group 1 of the C4~L6 ZHA router on RULER at 64K tokens. Chunk and top-level rankings remain strongly aligned ($\rho=0.86$).}
  \label{fig:qref-level-rates-l3kv1}
\end{figure}

We show the actual reference selection behavior per level for a typical KV group in \cref{fig:qref-level-rates-l19kv0}, and two outlier behaviors in \cref{fig:qref-level-rates-l39kv1} and \cref{fig:qref-level-rates-l3kv1}. The reference queries are shared across levels. In general, selection becomes increasingly concentrated at higher levels, while remaining more uniform at the lowest level.

\clearpage
\section{Reference-Query and Reference-Key Similarity}
\label[appendix]{app:reference-similarity}

\begin{figure}[H]
    \centering
    \includegraphics[width=\linewidth,height=0.73\textheight,keepaspectratio]{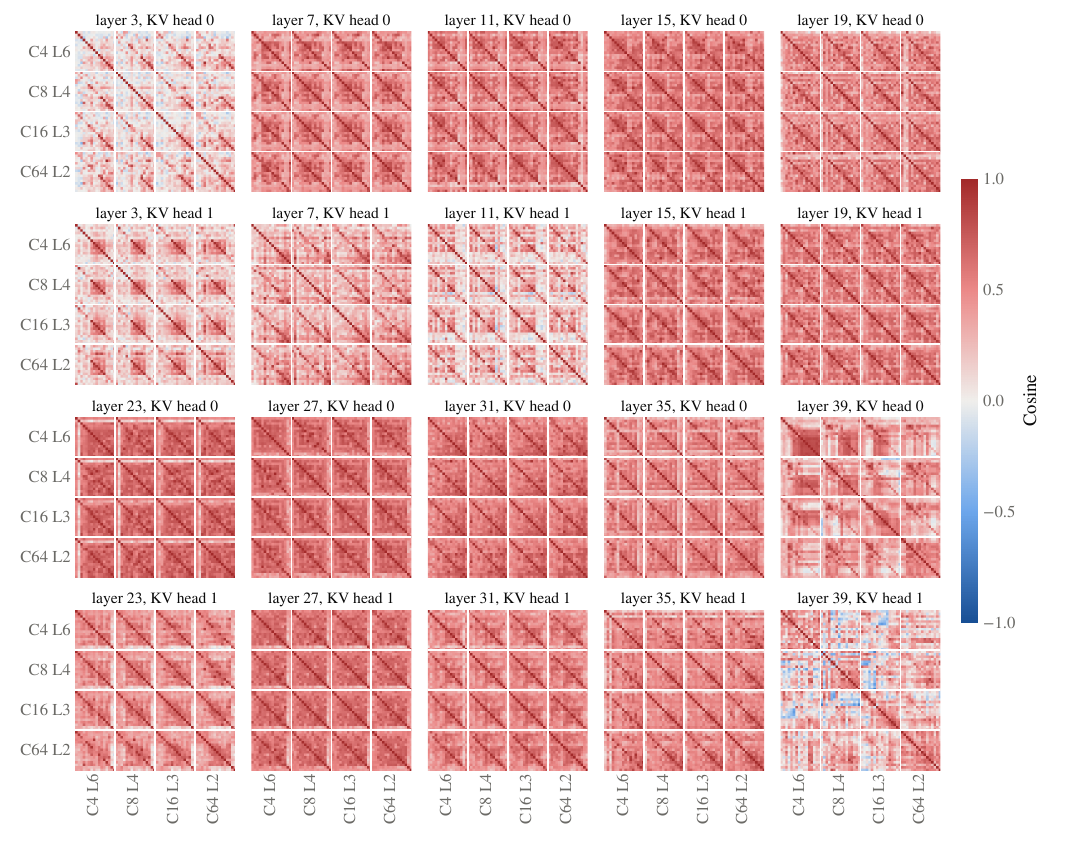}
\caption{Reference-query cosine similarities across the four Qwen3.6-35B-A3B ZHA geometries. The C16~L3 references are ordered by average-linkage clustering with cosine distance; references in each other geometry are Hungarian-matched to them by cosine similarity. Matched references therefore occupy corresponding positions. Diagonal structure within the off-diagonal blocks indicates similar reference directions across geometries.}
    \label{fig:qref-cos-r16}
\end{figure}

\begin{figure}[H]
    \centering
    \includegraphics[width=\linewidth,height=0.73\textheight,keepaspectratio]{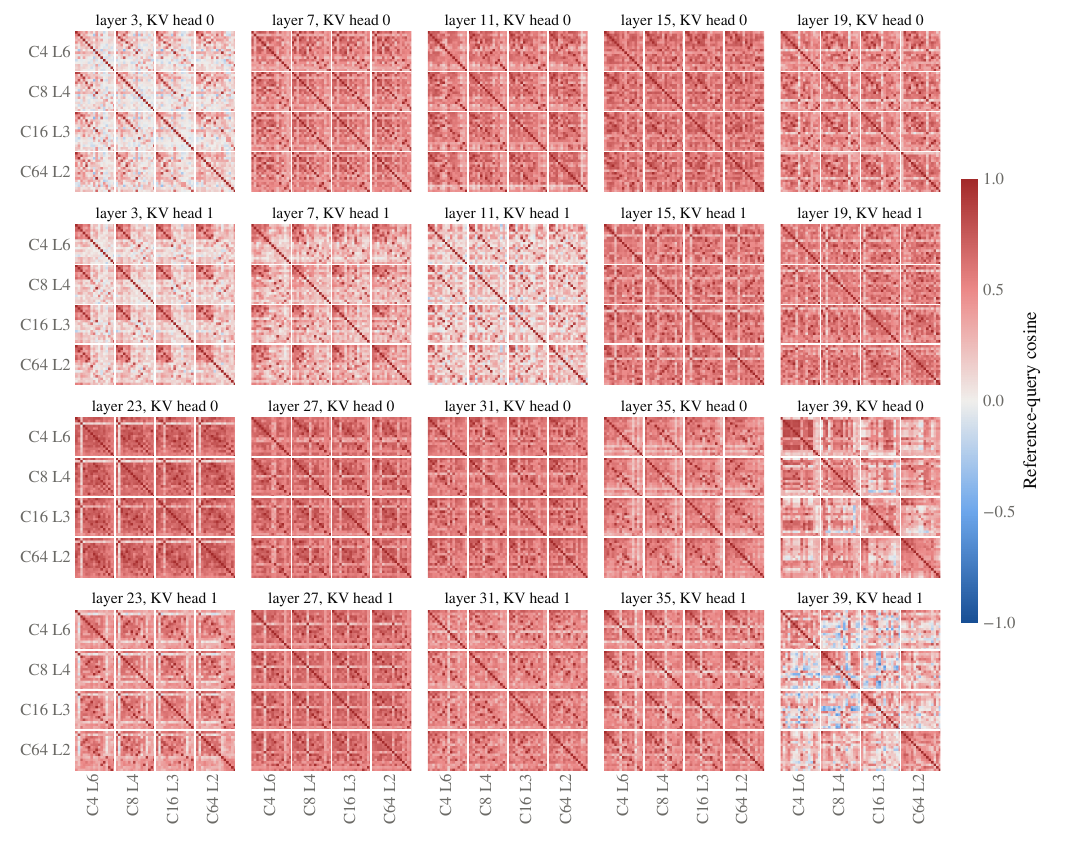}
\caption{Reference-query cosine similarities with references independently ordered by top-level selection frequency within each geometry and KV group, most to least frequent. Frequencies were measured during routed prefill on 16 RULER prompts at 64K tokens and $p=0.9$. No cross-geometry matching is applied; diagonal structure therefore indicates correspondence in both direction and selection rank.}
    \label{fig:qref-cos-r16-byusage}
\end{figure}

\begin{figure}[H]
    \centering
    \includegraphics[width=\linewidth,height=0.73\textheight,keepaspectratio]{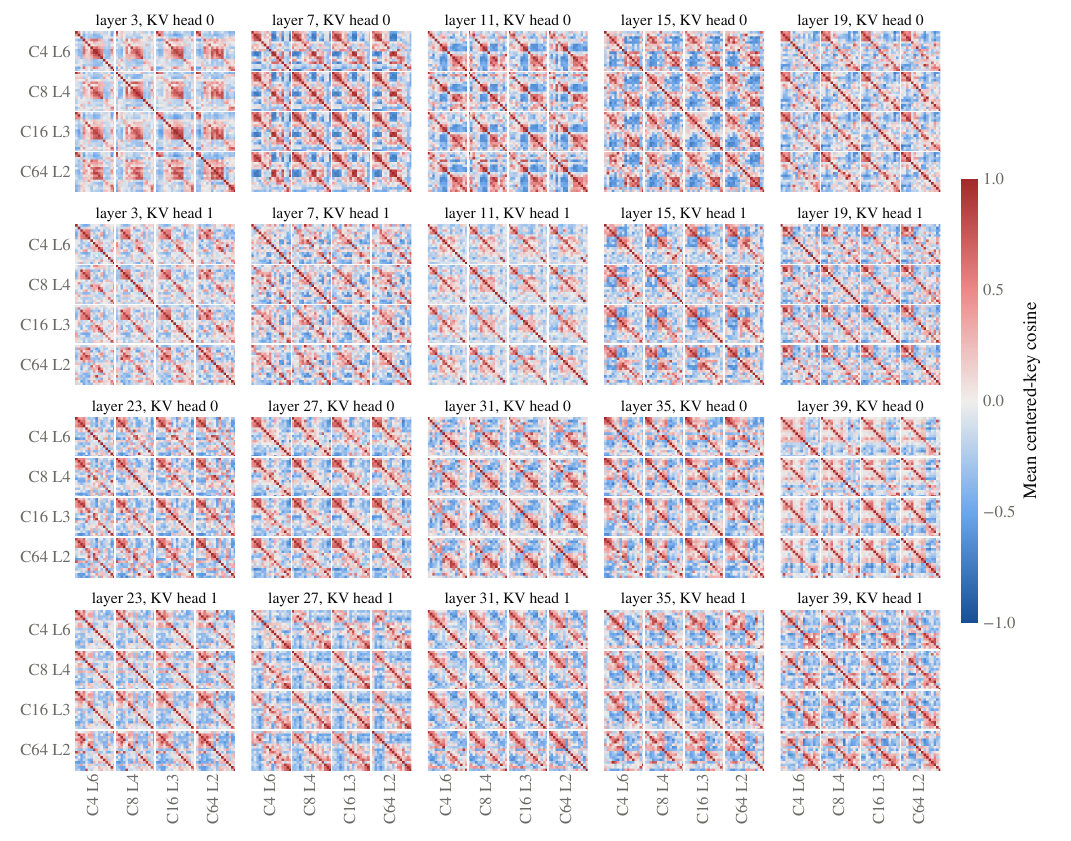}
\caption{Mean cosine similarities of centered reference keys across geometries, evaluated on the same 60 top-level nodes from four RULER prompts at 64K tokens. Each node covers 4096 keys. Within each node and KV group, the mean over the 16 reference keys is subtracted before computing cosine similarities, which are then averaged over nodes. C16~L3 references are clustered using reference-key cosine similarity; other geometries are Hungarian-matched using the same measure. The cross-geometry correspondence extends to the pooled key representations.}
    \label{fig:refkey-cos-r16-n4096}
\end{figure}

ZHA learns $R$ reference queries per KV group without explicitly regularizing their structure. Different geometries impose different chunk sizes, depths, and compression levels, raising the question of whether a common reference structure nonetheless emerges.
\Crefrange{fig:qref-cos-r16}{fig:refkey-cos-r16-n4096} compare the 16 references across four geometries, examining both the query vectors and the reference keys they produce.
Across most layers, matched references are similar across geometries, while reference keys remain differentiated within each geometry. This suggests that training finds a largely geometry-independent, specialized basis. All plots use the four independently trained R16 geometries C4~L6, C8~L4, C16~L3, and C64~L2; a top-level node covers exactly 4096 keys in each geometry.

\clearpage

\clearpage

\end{document}